\documentclass{article}

\PassOptionsToPackage{numbers,sort&compress}{natbib}
\usepackage[utf8]{inputenc} 
\usepackage[T1]{fontenc}    

\usepackage[preprint]{neurips_2025}

\usepackage{xcolor}         
\usepackage{hyperref}       
\usepackage{url}            
\usepackage{booktabs}       
\usepackage{amsfonts}       
\usepackage{nicefrac}       
\usepackage{microtype}      
\usepackage{graphicx}       
\usepackage{amsmath}        
\usepackage{booktabs}
\usepackage{threeparttable}
\usepackage{mathtools} 
\usepackage{cleveref}

\title{Causal-LLaVA: Causal Disentanglement for Mitigating Hallucination in Multimodal Large Language Models}

\author{%
	\begin{tabular}{c@{\hspace{2em}}c@{\hspace{2em}}c@{\hspace{2em}}c}
		\multicolumn{4}{c}{%
			\begin{tabular}{c}
				Xinmiao Hu\textsuperscript{1} \quad 
				Chun Wang\textsuperscript{1} \quad 
				Ruihe An\textsuperscript{1} \quad 
				ChenYu Shao\textsuperscript{1} \\
				Xiaojun Ye\textsuperscript{1} \quad 
				Sheng Zhou\textsuperscript{1} \quad 
				Liangcheng Li\textsuperscript{1} 
			\end{tabular}
		} \\
		\multicolumn{4}{c}{\textsuperscript{1}Zhejiang University} \\
		\multicolumn{4}{c}{\texttt{\{xinmiao\_hu, zjuheadmaster, 3220106416, 3230105139,}} \\
		\multicolumn{4}{c}{\texttt{yexiaojun,  zhousheng\_zju, liangcheng\_li\}@zju.edu.cn}}
	\end{tabular}
}

\begin{document}

\maketitle

\begin{abstract} Multimodal Large Language Models (MLLMs) perform well on visual understanding tasks, but they suffer from object hallucination and generating inconsistent or nonexistent object descriptions due to the dataset biases. 
This paper explores how dataset biases affect object hallucinations in MLLMs on feature space, revealing that biased object co-occurrences lead to entangled semantic representations across modalities. 
Thus, the model mistakenly activates the representation of an object that often appears with the input object but is absent, leading to hallucinations. 
To address this, we propose a causality-based disentanglement framework using causal intervention to deconfound the biases introduced by frequent object co-occurrences in the data. 
Specifically, we introduce a Causal-Driven Projector in the visual pathway and a Causal Intervention Module in the final LLM transformer layer, which work together to mitigate spurious correlations. 
Our experiments show a notable reduction in hallucinations while preserving strong performance across several benchmarks, along with visualization analysis confirming improved representation separability.  
Our code is available at \url{https://github.com/IgniSavium/Causal-LLaVA}.\end{abstract}

\section{Introduction}

\begin{figure}[h]
  \centering
  \includegraphics[width=0.32\linewidth]{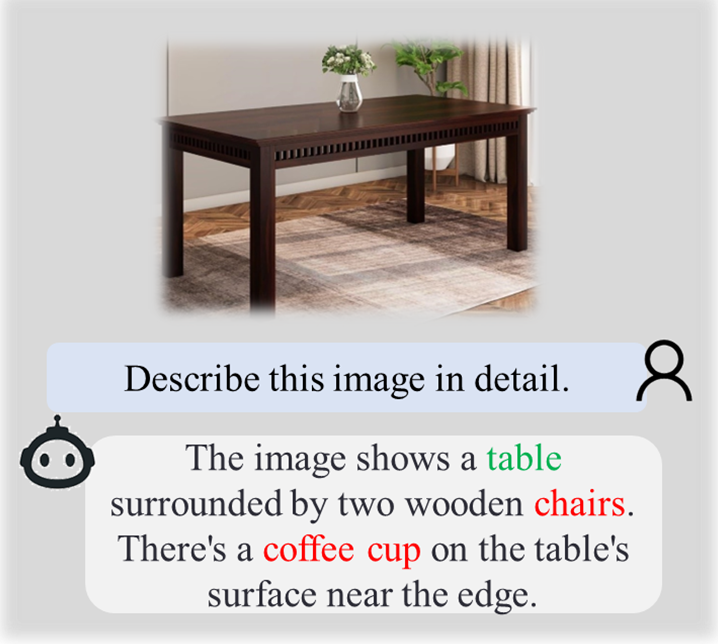}
  \hfill
  \includegraphics[width=0.32\linewidth]{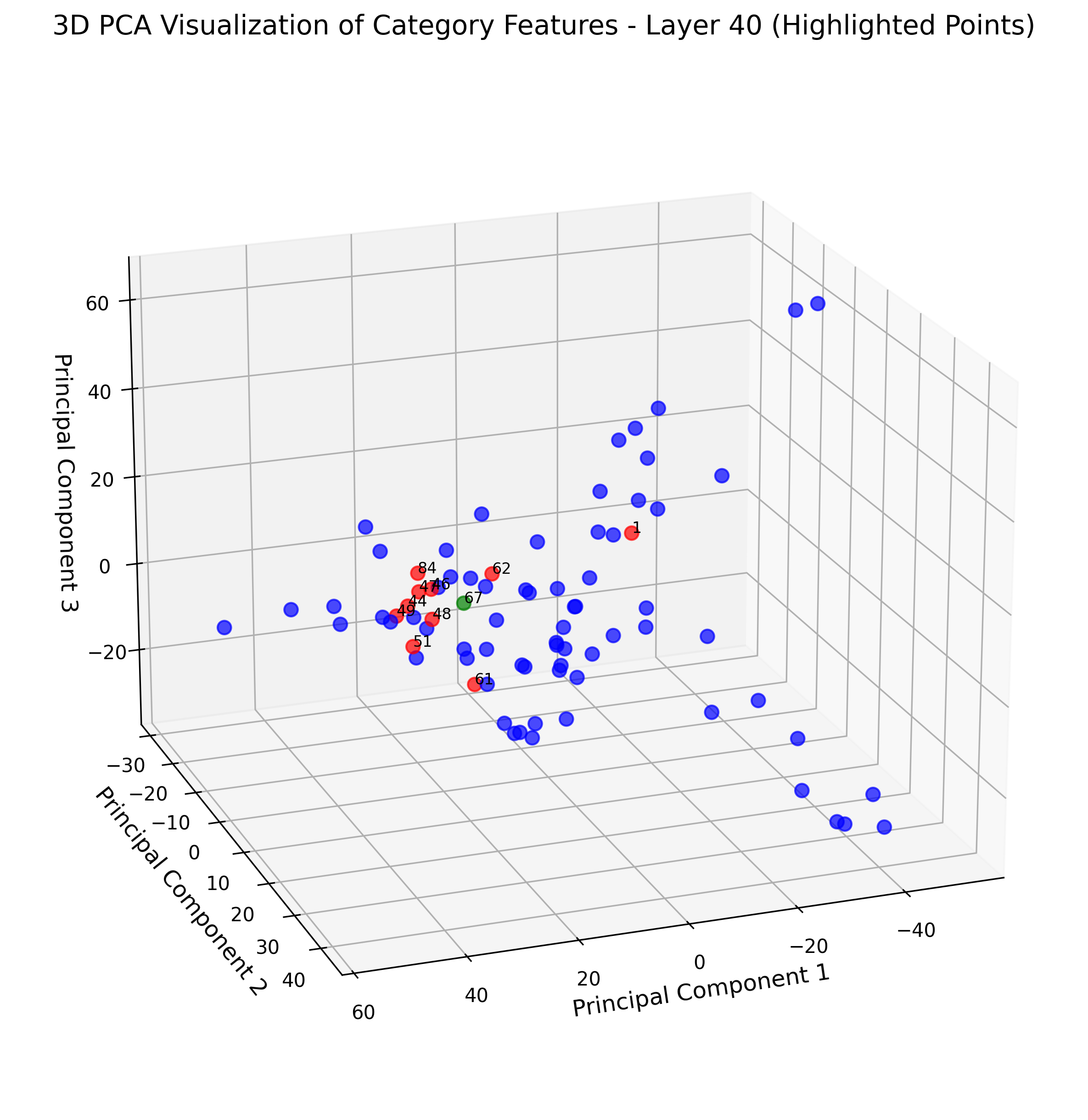}
  \hfill
  \includegraphics[width=0.32\linewidth]{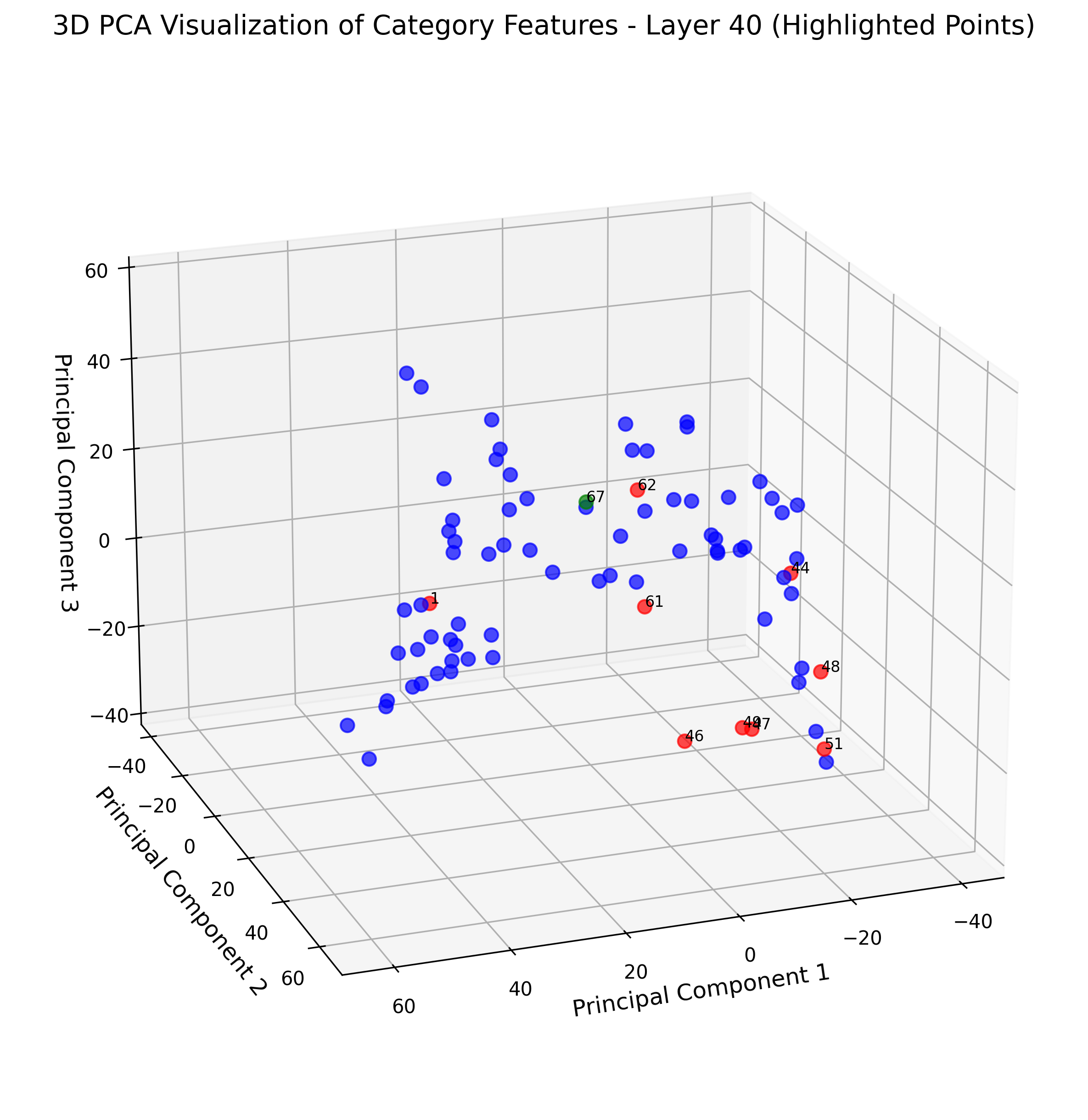}
  \caption{ \textbf{left:} A hallucination example where an MLLM incorrectly describes non-existent objects (chairs and a coffee cup) based on co-occurrence patterns. 
        \textbf{middle:} Representations of original LLaVA's final LLM Transformer layer. The dining table (green dot, id=67) is surrounded by its top-10 most frequently co-occurring objects (red dots).
        \textbf{right:} Disentangled representations in our framework, with previously clustered objects now drifting significantly apart.}
  \label{fig:figure-1}
\end{figure}

Multimodal Large Language Models (MLLMs) like LLaVA \citep{Liu2023VisualIT,Liu2023ImprovedBW}, MiniGPT-4 \citep{Zhu2023MiniGPT4EV} and InstructBLIP \citep{Dai2023InstructBLIPTG} have demonstrated their remarkable ability in visual understanding tasks, such as image captioning \citep{karpathy2015deep}, visual question answering (VQA) \citep{antol2015vqa}, and complex multimodal reasoning \citep{hudson2019gqa}. 
Despite these achievements, MLLMs still suffer from hallucination, where models generate content inconsistent with visual inputs, such as describing nonexistent objects, incorrect attributes, or implausible relationships \citep{Liu2024ASO,Bai2024HallucinationOM}. 
Among these issues, object hallucination, where models describe inconsistent or nonexistent objects, represents a significant challenge that has drawn substantial attention in the field.
Because of the co-occurrence biases on datasets, models will mistakenly infer one object’s presence due to its frequent co-occurrence with another object\citep{Li2023EvaluatingOH}. For example, an MLLM might hallucinate chairs around a table when processing visual inputs containing only the table itself, due to learned biased co-occurrence patterns that chairs are always around a table statistically from datasets (see Figure~\ref{fig:figure-1} left). 
Such hallucinations fundamentally undermine MLLMs' reliability in downstream reasoning, leading to erroneous semantic interpretations, breaking multi-step inference chains built on visual context \citep{Hudson2019GQAAN}, and impairing grounding for sequential vision-based agents in dynamic environments \citep{Das2018NeuralMC}. These failures not only erode user trust but also pose serious risks in safety-critical applications requiring accurate perception.

While previous work \citep{Li2023EvaluatingOH} has only shown that co-occurrence frequencies are statistically linked to hallucination rates, without exploring the deeper reasons behind this, we investigate the dynamic formation of these biases throughout the representation learning process. This helps us gain a more nuanced understanding of how statistical patterns in the training data gradually shape the model's internal representations, allowing us to identify critical stages where biases are most likely to be amplified or mitigated.  
We first examine the distribution of textual and visual representations within the semantic space of MLLMs.
In Figure \ref{fig:figure-1} (middle and right), dining table and its top-10 most frequently co-occurring objects on MSCOCO ~\cite{Lin2014MicrosoftCC} are visualized via average visual object representations in the original LLaVA’s final LLM Transformer layer using principal component analysis (PCA). 
We observe that after the \textbf{trained projector} and in \textbf{the lower half and final layers of the Transformer}, the entanglement of hidden states shows strong correlation with the co-occurrence frequency of objects in the instruction dataset (see Figure~\ref{fig:figure-1} middle).

Based on these findings, we conclude that such spurious correlations arise directly from the biased representation learning process. 
This representational entanglement causes the model to align the semantic representations of frequently co-occurring objects, thus one object occurs will raise the probability of the other related object occurring with the risk of ignoring the real visual information.

Statistically biased co-occurrence and associations embedded in the semantic space of MLLMs are a primary contributor to object hallucinations. We identify these entangled representations as a key underlying cause. They stem from two main sources: (1) object co-occurrence patterns present in the training data, and (2) the model’s inclination to associate objects based on superficial correlations rather than genuine causal relationships\citep{Kim2018LearningNT,Joachims2016UnbiasedLW,hong2021unbiased}. While many existing approaches attempt to address the first issue through data reweighting\citep{Liu2023MitigatingHI,Kim2023ExposingAM,Wang2023VIGCVI}, they often overlook the second and more fundamental challenge of understanding how the model internally learns and represents semantic relations. To enable MLLMs to comprehend essential semantic and logical structures, it is crucial to establish unbiased representations. Causal inference, which uncovers causal relationships from observational data, offers a principled way to bridge empirical patterns and underlying causality \citep{pearl2018book}. Building on this insight, we propose a causality-driven framework with two parts: a Causal-Driven Projector that decouples co-occurring object features, and Causal Intervention Modules that purify hidden states in top Transformer layers. This two-part design breaks the spread of bias during vision-language learning (see Figure~\ref{fig:figure-1} right), helping the model tell objects apart and reducing hallucinations from data bias."

Empirical evidence demonstrates that our method effectively suppresses hallucinations without compromising performance across various comprehensive benchmarks, with the integration of our Disentanglement Modules into LLaVA yielding a 22.6\% increase in the MME-Perception score \citep{Fu2023MMEAC}.

In conclusion, this paper makes the following contributions:

\begin{enumerate}
    \item We reveal that object hallucinations in MLLMs are rooted in entangled semantic representations formed during training, driven by co-occurrence biases in data. Our analysis focuses on how these representations evolve within the model, offering a new perspective at the representational level.
    \item We design a causality-based disentanglement architecture that separates object representations to reduce hallucinations from commonly confused combinations. This module can be easily adapted to different multimodal large language model architectures.
   \item We empirically validate that our approach not only effectively reduces hallucinations but also preserves strong performance across multiple benchmarks. The code and framework will be released publicly for further research.
\end{enumerate}
\section{Related Work}
\subsection{Hallucination in MLLMs}
Hallucination in Multimodal Large Language Models (MLLMs) refers to generating content inconsistent with input images \citep{Bai2024HallucinationOM}. 
The causes of hallucinations in MLLMs are multifaceted, including over-reliance on parametric or linguistic knowledge \citep{Zhai2023HallEControlCO,Wu2024NoiseBoostAH}, biases in training data (distribution imbalances and annotation irrelevance) \citep{Hu2023CIEMCI,You2023FerretRA}, insufficient attention to contextual information in large language models (LLMs) \citep{Lee2023VolcanoMM,Huang2023OPERAAH}, and resolution-constrained vision encoders that struggle to capture fine-grained semantics \citep{Li2023MonkeyIR,Shi2024EagleET}, all of which collectively exacerbate hallucination phenomena.

Prior efforts to mitigate hallucinations have explored diverse strategies. Wang et al. \citep{Wang2024AdvancingFV} and Jain et al. \citep{Jain2023VCoderVV} refine cross-modal interactions through auxiliary visual controls like segmentation maps to improve modality alignment. Kim et al. \citep{Kim2023ExposingAM} generate decorrelated synthetic datasets to reduce spurious correlations; Liu et al. \citep{Liu2023MitigatingHI} and Wang et al. \citep{Wang2023VIGCVI} employ negative samples or iterative correction during large-scale instruction dataset creation. Contrastive decoding strategies by Wang et al. \citep{Wang2024MitigatingHI} and Leng et al. \citep{Leng2023MitigatingOH} suppress hallucinations by contrasting outputs from original and perturbed inputs (e.g., altered instructions or distorted images), adjusting reliance on language priors. Yin et al. \citep{Yin2023WoodpeckerHC} and Zhou et al. \citep{Zhou2023AnalyzingAM} validate outputs using external expert models without modifying training processes. Reinforcement learning approaches like RLHF \citep{Sun2023AligningLM} align outputs with human preferences through iterative feedback.

Existing methods face two key limitations. Most rely on external tools like GPT-4 for data generation or expert models for validation, causing error propagation and inefficiency. Others use contrastive decoding or indirect alignment (RLHF), which fail to tackle the deep cause: the distorted semantic representations caused by biased data distributions. In contrast, we propose a novel architectural solution based on feature disentanglement learning, which requires no synthetic data, external models, or post-processing, enabling lightweight, end-to-end mitigation of hallucinations.

\subsection{Causal Inference}
Causal inference has been widely adopted across AI domains to tackle bias and spurious correlations. In visual question answering (VQA), Niu et al. \citep{Niu2020CounterfactualVA} explicitly model language bias as a causal effect, subtracting its direct influence from predictions to improve robustness. Liu et al. \citep{Liu2022ShowDA} extend this paradigm to image captioning by employing backdoor adjustment with interventional object detectors and transformer decoders to disentangle visual and linguistic confounders. In image-text matching, Li et al. \citep{Li2023TowardsDI} further advance the framework by decomposing intra- and inter-modal confounders through backdoor adjustments, enabling more reliable alignment. Beyond these structured applications, causal inference has also been explored in image recognition\citep{LopezPaz2016DiscoveringCS,Yang2020DeconfoundedIC,Wang2020VisualCR,Yang2021CausalAF}, dialog generation\citep{Qi2019TwoCP, Zhu2020CounterfactualOT}, video analysis\citep{Li2022FromRT}, and scene graph generation\citep{Tang2020UnbiasedSG,Sun2023UnbiasedSG}. Despite these advancements, existing approaches typically rely on task-specific architectures and smaller-scale models, restricting scalability. Moreover, few investigate causal integration within high-capacity multimodal systems like multimodal large language models (MLLMs). This gap highlights the urgent need for methodologies that can systematically mitigate confounding factors in large-scale, generalizable architectures while maintaining strong performance across diverse tasks.
\section{Influence of Instruction Bias on Hidden States}

Multimodal Large Language Models (MLLMs) typically use instruction-tuning datasets derived from traditional vision datasets like MSCOCO\citep{Lin2014MicrosoftCC} and Visual Genome\citep{Krishna2016VisualGC}. However, these datasets contain inherent biases from long-tailed object distributions, geographic skews, and annotator preferences for common scenes. These manifest as imbalanced object frequencies (e.g., "person" vs. "hairdryer") and constrained co-occurrence patterns (e.g., "table+chair" vs. "table+snowboard"). When MLLMs generate instruction data through automated template/rule-based methods\citep{Liu2023VisualIT,Liu2023MitigatingHI}, these biases become amplified, creating highly skewed object distributions.

We analyze LLaVA\citep{Liu2023VisualIT,Liu2023ImprovedBW}—trained on MSCOCO-derived data—to study how instruction dataset biases, particularly object pair co-occurrences (e.g., "dining table" centrality), affect latent representations in MLLMs. While we focus on visual modality in the main text, Appendix provides textual analysis and additional examples.

\begin{figure}[t]
  \centering
  \includegraphics[width=0.32\linewidth]{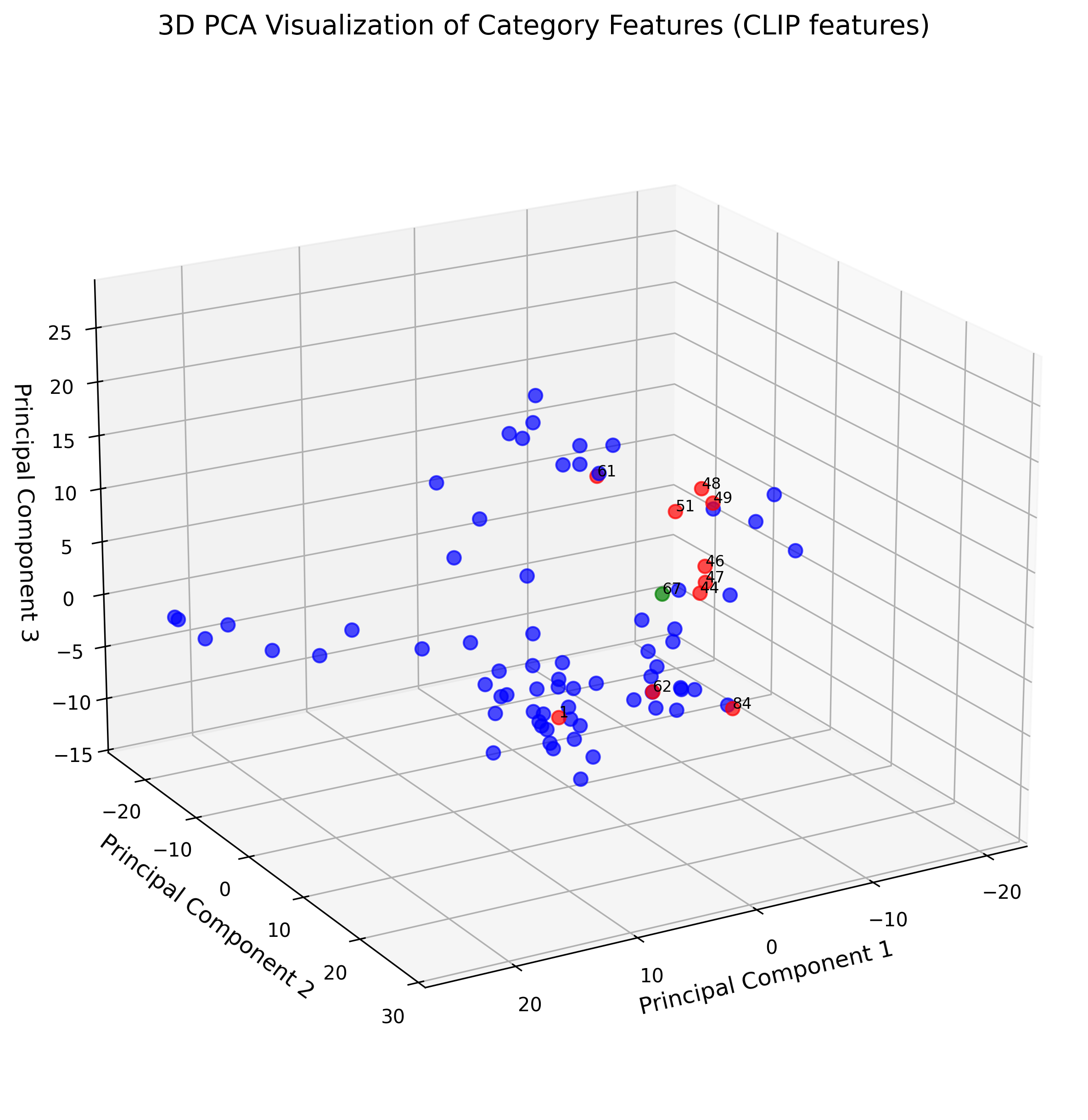}
  \hfill
  \includegraphics[width=0.32\linewidth]{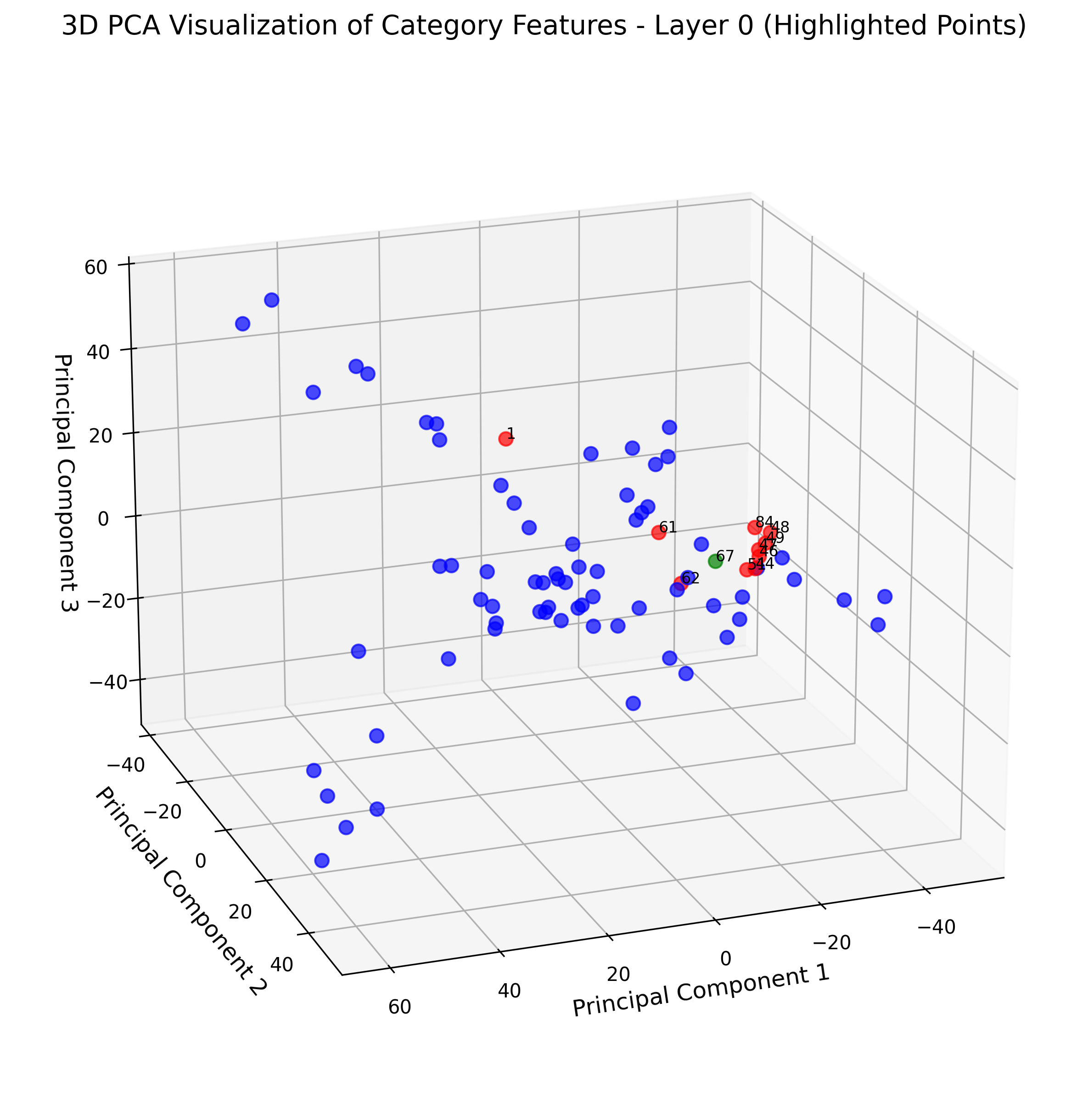}
  \hfill
  \includegraphics[width=0.32\linewidth]{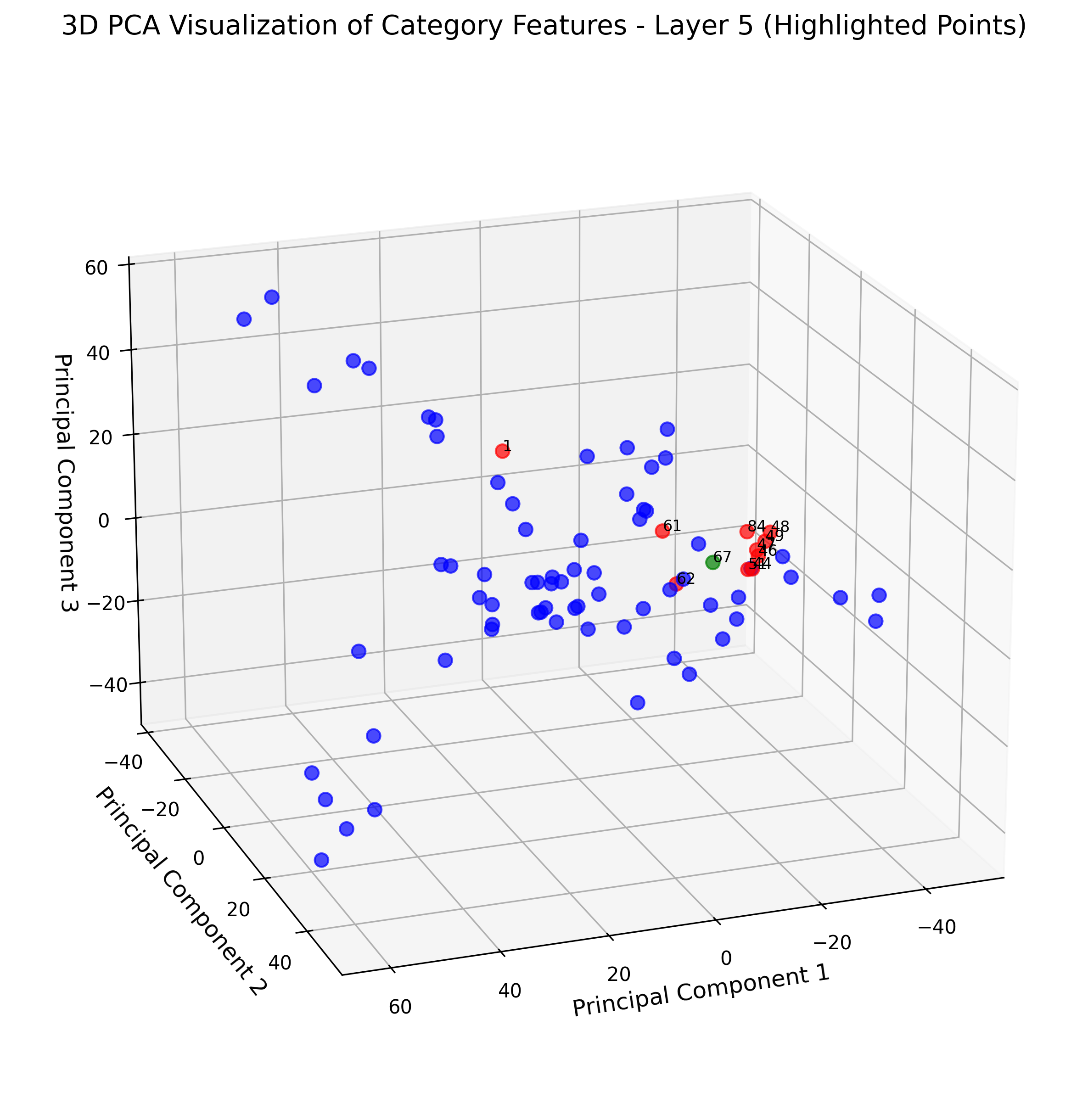}
  \caption{Original LLaVA average object visual representations PCA after Vision Encoder (left), Projector (middle), and LLM layer 5 (right). Green dots: "dining table"; Red dots: top-10 co-occurring objects.}
  \label{fig:visualization-1}
\end{figure}

\begin{figure}[t]
  \centering
  \includegraphics[width=0.32\linewidth]{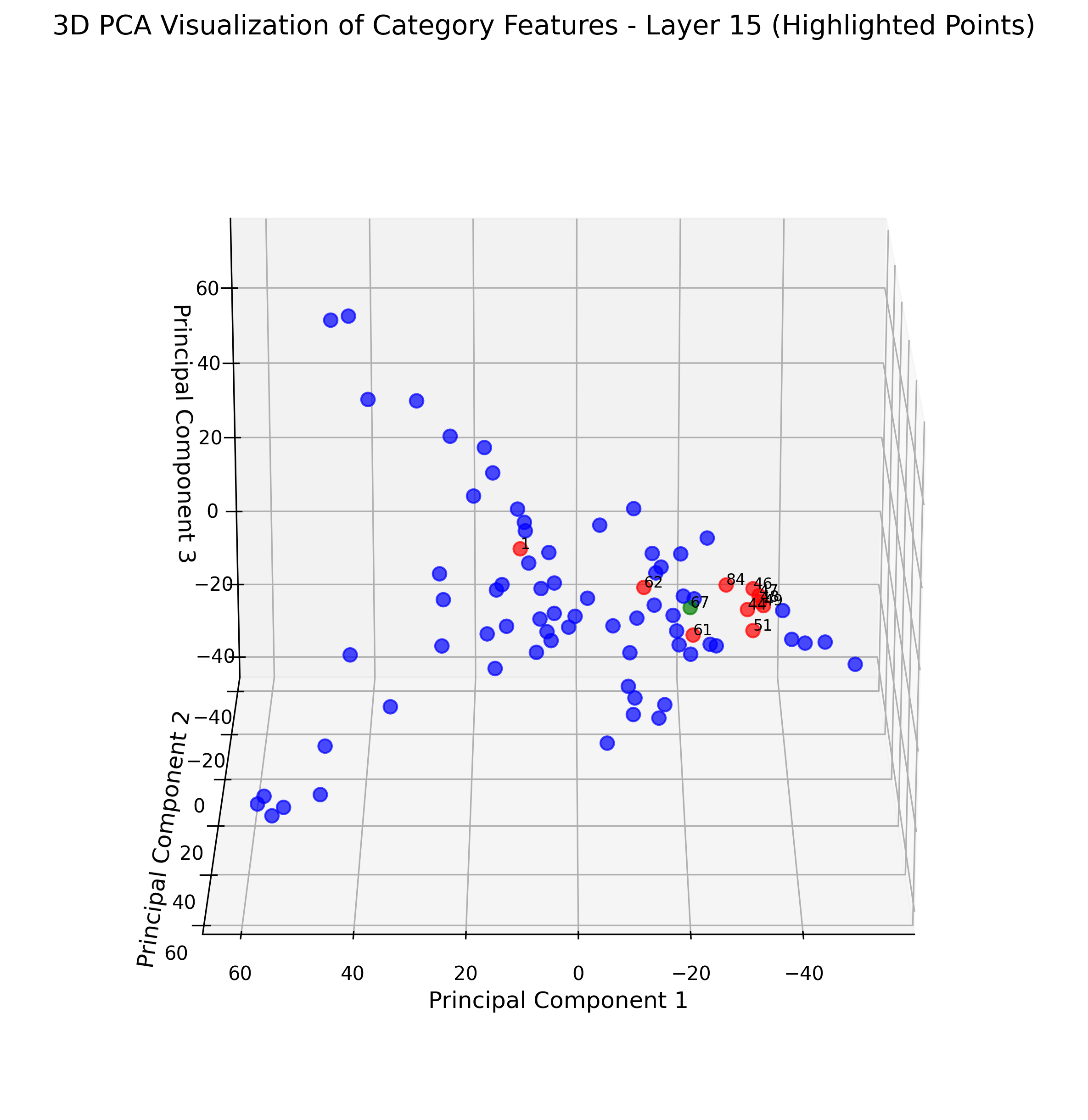}
  \hfill
  \includegraphics[width=0.32\linewidth]{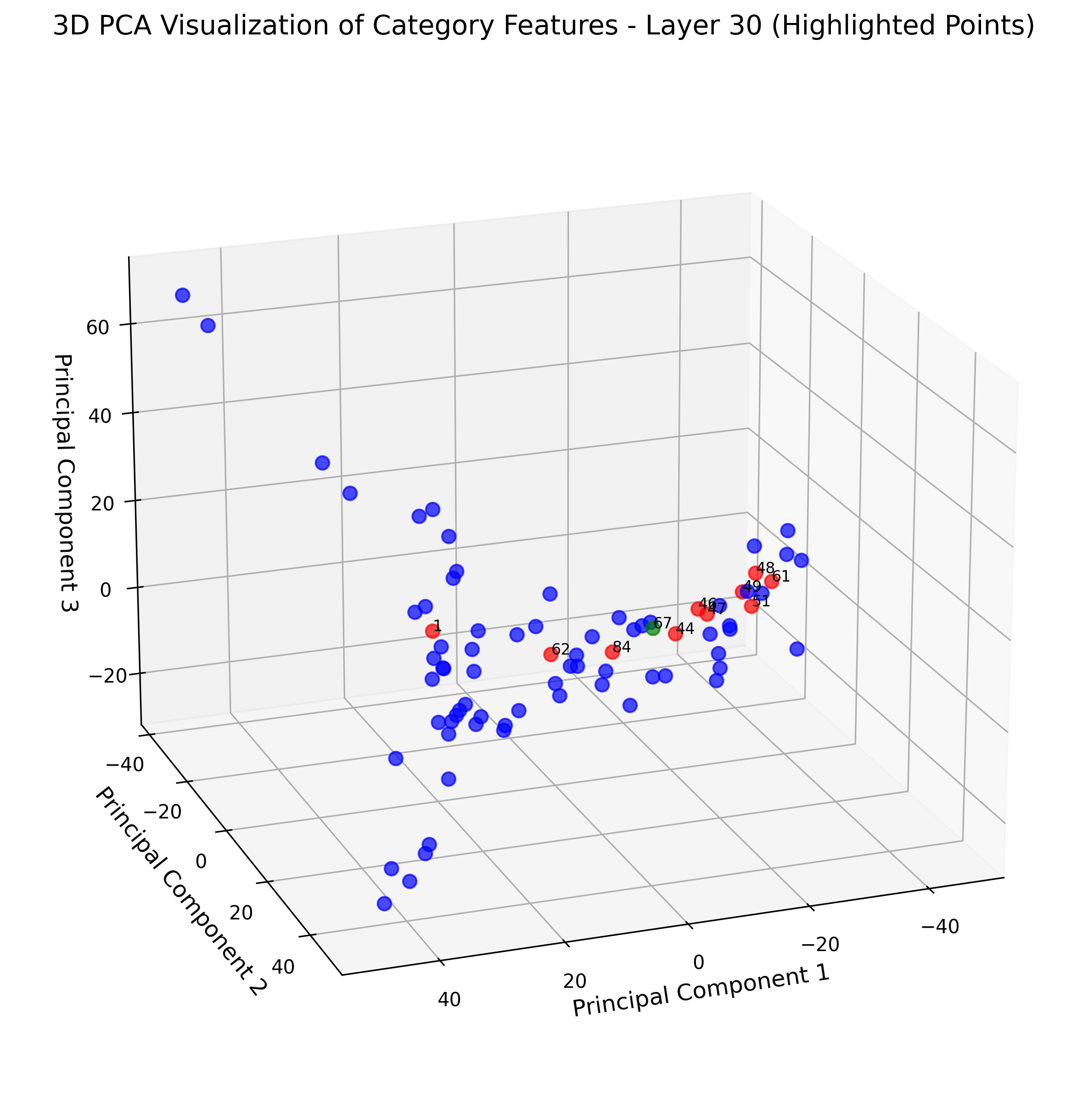}
  \hfill
  \includegraphics[width=0.32\linewidth]{./Paper_Draft_assets/3d_PCA_layer40_elev20_azim70.png}
  \caption{Original LLaVA average object visual hidden state distributions PCA in LLM layers 15 (left), 30 (middle), and 40 (right). Green dots: "dining table"; Red dots: top-10 co-occurring objects.}
  \label{fig:visualization-2}
\end{figure}

As shown in Figure~\ref{fig:visualization-1}, principal components of average visual representations (post-Vision Encoder/Projector) reveal distinct patterns. The green dot ("dining table") and red dots (top-10 co-occurring objects) show minimal clustering after the frozen CLIP Vision Encoder. However, post-Projector processing results in tight clustering, indicating that co-occurrence frequency drives representational entanglement—a key contributor to model hallucinations.

Figure~\ref{fig:visualization-1}(right) illustrates layer 5 of the LLM (representative of layers 1-5). Here, hidden state distributions directly and strongly mirror post-Projector visual tokens: "dining table" remains prominently surrounded by co-occurring objects, confirming that Projector-induced entanglement directly shapes early LLM semantic understanding.

Figure~\ref{fig:visualization-2}(left) extends this to layer 15 (representative of layers 6-15), where persistent clustering confirms a strong correlation between hidden state entanglement and training data co-occurrence frequencies. These results demonstrate that instruction biases propagate through multiple model layers, exerting a lasting and consistent influence on the MLLM's visual semantic understanding process.

At higher LLM layers (Figure~\ref{fig:visualization-2}(middle): layer 30), co-occurrence-driven clustering weakens as category representations disperse. This reflects a shift from an "understanding" phase (analyzing visual input semantics) to a "prediction" phase (next-token probability estimation), where object representations undergo distinct spatial transformations. These transformations map categories to more distinguishable semantic units, altering the relationship between co-occurrence frequency and representational entanglement compared to earlier stages.

Finally, Figure~\ref{fig:visualization-2}(right) shows that although there is a representation distribution shift during late prediction phases, layer 40 (final hidden states for the next token prediction) still retains strong entanglement between "dining table" and its co-occurring objects. This confirms that instruction bias-induced entanglement not only persists through visual semantic understanding but also fundamentally defines the structure of the final token prediction space.
\section{Methodology}

\subsection{Backdoor Adjustment}

Causal inference addresses dataset biases caused by confounders in AI applications \citep{Yao2020ASO,Jiao2024CausalIM,Yang2020DeconfoundedIC}. A confounder is a common cause of two variables, inducing spurious associations. In machine learning, they manifest as contextual biases (e.g., visual features influenced by real-world contexts in object detection or image captioning).

The observational probability $ P(Y \mid X) $ captures this distortion through:
\begin{equation}
    P(Y \mid X) = \sum_z P(Y \mid X, Z=z)P(Z=z \mid X),
    \tag{1}
\end{equation}
where confounder $ Z $ biases estimation via $ P(Z=z \mid X) $. Consider image captioning: if $ P(Z=\text{chair} \mid X=\text{table}) \gg P(Z=\text{toaster} \mid X=\text{table}) $, then $ P(Y=\text{Tokens}_\text{table} \mid X=\text{table}, Z=\text{chair}) $ dominates estimation over $ P(Y=\text{Tokens}_\text{table} \mid X=\text{table}, Z=\text{toaster}) $ during parameter training. This creates a bias where features of "chair" are incorrectly linked to "table", as the model prioritizes co-occurrence patterns over intrinsic characteristics.

\begin{figure}[t]
  \centering
  \begin{minipage}{0.12\linewidth}
    \centering
    \includegraphics[width=\linewidth]{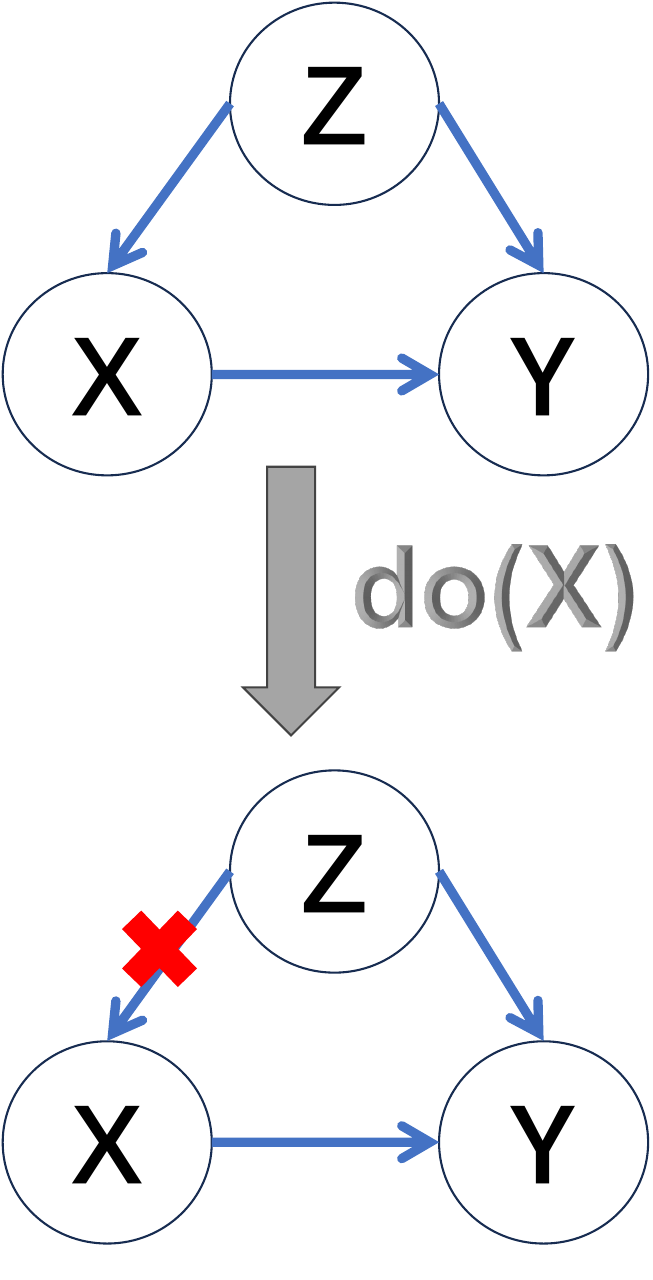}
    \label{fig:backdoor_adjustment}
  \end{minipage}
  \hfill
  \begin{minipage}{0.4\linewidth}
    \centering
    \includegraphics[width=\linewidth]{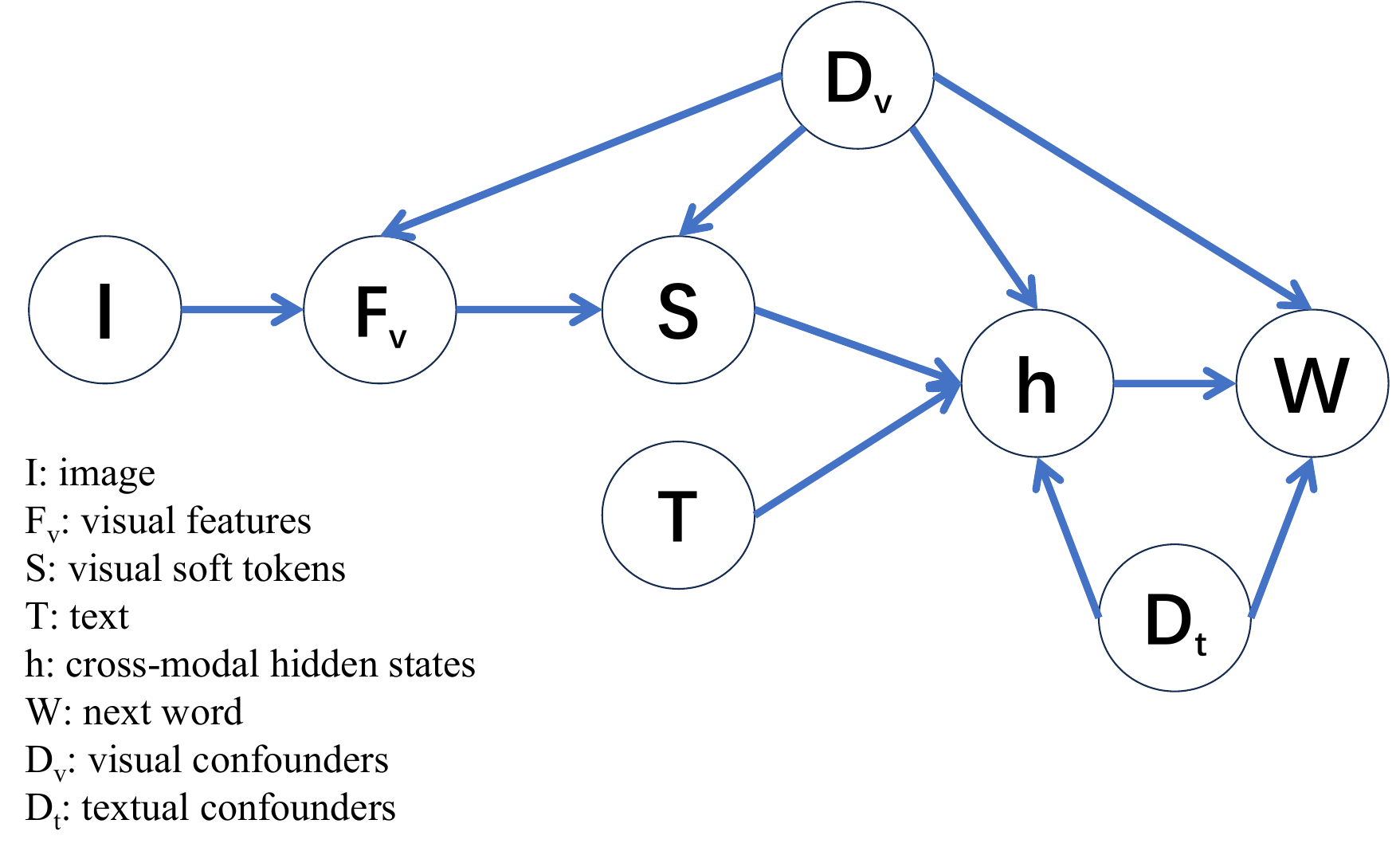}
    \label{fig:causal_paths}
  \end{minipage}
  \hfill
  \begin{minipage}{0.4\linewidth}
    \centering
    \includegraphics[width=\linewidth]{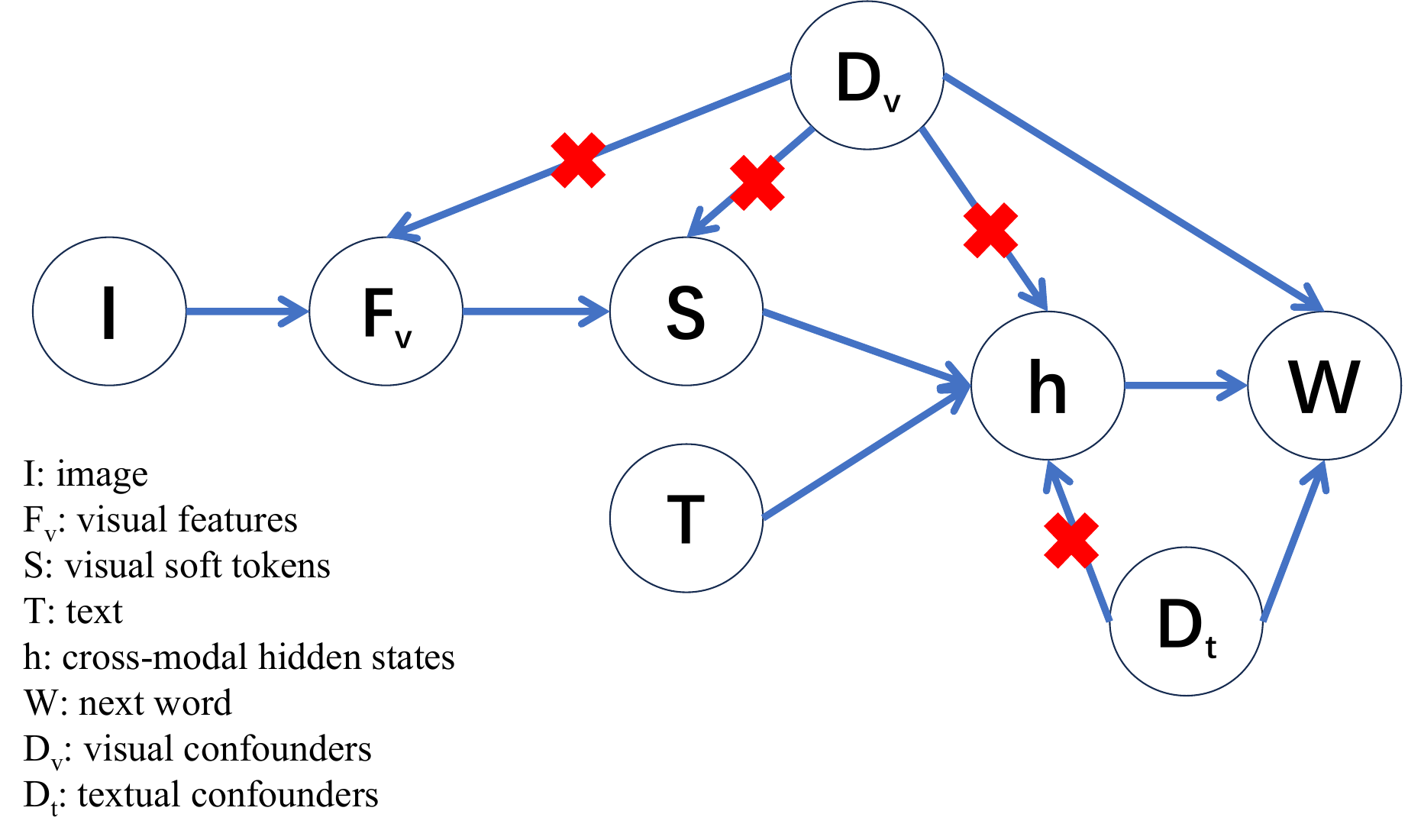}
    \label{fig:deconfounded_result}
  \end{minipage}
  \vspace{2ex}
  
  \caption{Unified illustration of causal mechanisms in MLLMs. (a) Backdoor adjustment blocks confounding paths. (b) Causal relationships involving visual/textual confounders in MLLM processing. (c) Final deconfounded results.}
  \label{fig:combined_causal}
\end{figure}

To address this, causal intervention via $ P(Y \mid \text{do}(X)) $ has been proposed \citep{Yue2020InterventionalFL,Wang2020VisualCR}. The $\text{do}$-operator severs the causal link between confounder $ Z $ and input $ X $, blocking the backdoor path $ X \gets Z \to Y $. As illustrated in Figure~\ref{fig:combined_causal}(a), this enables learning of the true causal effect of $ X $ on $ Y $. The backdoor adjustment formalizes this as:
\begin{equation}
    P(Y \mid \text{do}(X=x)) = \sum_{z} P(Y \mid X=x, Z=z) P(Z=z),
    \tag{2}
\end{equation}
with $ Z $ as the confounder.

This formula ensures fair integration of all $ Z $ values under a uniform prior $ P(Z=z) $, decoupling their correlation with $ X $. For example, in image captioning, treating frequent concepts as confounders in do-intervention deconfounds the relationship between image features and captions. By maximizing $ P(\text{Tokens}_\text{X} \mid \text{do}(X)) $, the model learns direct visual representations from images, avoiding spurious dataset-induced co-occurrences.

However, applying Equation (2) in deep learning is computationally intensive due to extensive sampling requirements. To address this, the Normalized Weighted Geometric Mean (NWGM) approximation has been proposed \citep{Wang2020VisualCR,Xu2015ShowAA}. For linear probability models $ f $, NWGM simplifies the causal intervention as (see Appendix):
\begin{equation}
    P(Y \mid \text{do}(X)) \overset{NWGM}{\approx} \text{Softmax}\left(f\left(x, \mathbb{E}_z[z]\right)\right),
    \tag{3}
\end{equation}
Here, $ Z = [z_1,\dots,z_n] $ is a predefined confounder dictionary with $ z_i $ approximated by category-averaged hidden features. This reduces computation by replacing marginalization over $ Z $ with expectation estimation. Our specific implementation will be mentioned later.

\subsection{Causal Relationship Analysis}
\label{sec:relationship_graph}
We model the causal relationships in LVLMs using a Structural Causal Model (SCM) with key variables: input image $ I $, visual features $ F_v $, visual soft tokens $ S $, textual tokens $ T $, LLM hidden states $ h $, predicted word $ W $, visual confounders $ D_v $, and textual confounders $ D_t $ \citep{Chalupka2017CausalFL,LopezPaz2016DiscoveringCS}. 

As shown in Figure~\ref{fig:combined_causal}(b), the causal paths $ h \to W $, $ S \to h $, and $ T \to h $ indicate that LLM hidden states determine word predictions through multimodal fusion. The chain $ I \to F_v \to S $ reflects sequential encoding via the visual encoder and projector.

The paths $ D_v \to F_v $, $ D_v \to S $, $ D_v \to h $, and $ D_t \to h $ reveal visual biases during training. For instance, object representations may skew toward frequent co-occurring contexts in the dataset. The effects $ D_v \to W $ and $ D_t \to W $ imply direct confounding on lexical distributions, particularly object-related tokens.

Based on the causal analysis, using the observational likelihood $ P(W \mid h) $ as the training target introduces biases through confounders $ D_v $ and $ D_t $. Its decomposition is:
\begin{equation}
    P(W \mid h) = \sum_{d_v} P(d_v \mid h) \cdot \sum_{d_t} P(d_t \mid h) P(W \mid h, d_v, d_t).
    \tag{4}
\end{equation}

To block the backdoor paths $ h \leftarrow D_v \to W $ and $ h \leftarrow D_t \to W $, we apply backdoor adjustment \citep{Yue2020InterventionalFL,Wang2020VisualCR} with the causal intervention(see Appendix):
\begin{equation}
    P(W \mid \text{do}(h)) \overset{NWGM}{\approx} P\big(W \mid h, \mathbb{E}_{d_v}[d_v], \mathbb{E}_{d_t}[d_t]\big),
    \tag{5}
\end{equation}

The causal path $ S \to h $ is further affected by visual confounders $ D_v $. While prior work approximates Eq. (5) via NWGM (Eq. 3) \citep{Wang2020VisualCR}, this approach limits causal learning to the final Softmax layer, leaving earlier biased representations unaddressed. Since neural networks inherently encode conditional probabilities through pattern recognition \citep{Nie2018TheDR}, we shift causal intervention from output logits to learned features, hypothesizing that unbiased features yield unbiased predictions.

To block the backdoor path $ S \leftarrow D_v \to h $, we propose a causally intervened projector module that generates disentangled visual tokens $ S $. Let $ g $ denote the intervention-driven projector function:
\begin{equation}
    \mathbf{CausalDrivenProjector}(F_v, Z) = \mathbb{E}_z [g(f_v, z)] \overset{\text{NWGM}}{\approx} g_f(f_v) + g_z(\mathbb{E}_z[z]),
    \tag{6}
\end{equation}
where $ Z $ is a confounder dictionary.

Finally, the path $ F_v \leftarrow D_v \to S $ is approximately blocked in practice, owing to two factors: (1) CLIP vision encoders are trained on diverse datasets, reducing potential bias in $ F_v $; and (2) CLIP is typically frozen during MLLM training (e.g., LLaVA), thereby decoupling $ D_v $ from $ F_v $.

\subsection{Disentangled Model Architecture}
\label{sec:disentangled}

\begin{figure}[t]
  \centering
  \includegraphics[width=0.65\linewidth]{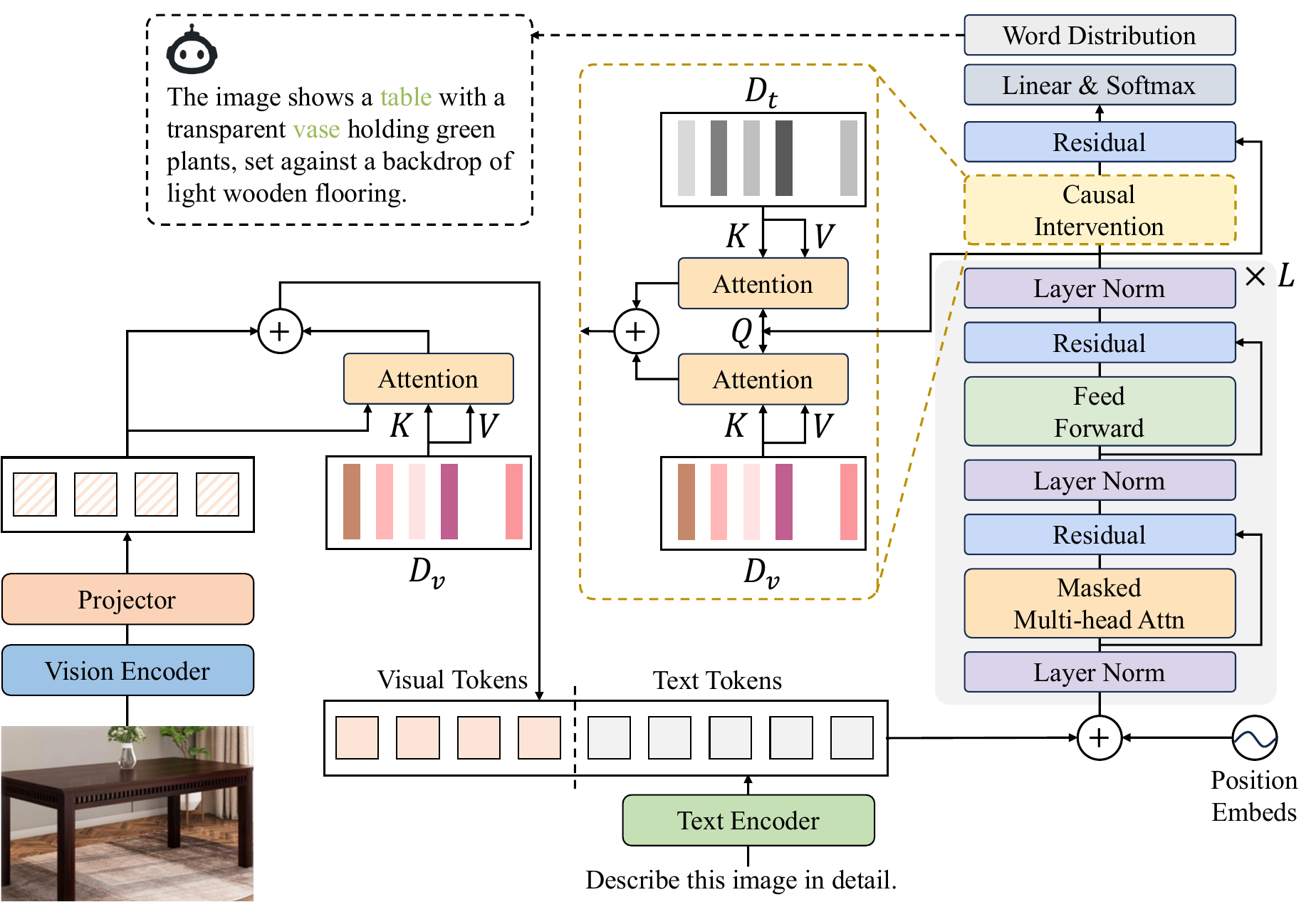}
  \caption{Illustration of our innovative disentangled MLLM model architecture.}
  \label{fig:disentangled_architecture}
\end{figure}

We enhance the MLLM architecture by integrating causal intervention modules into the modality-alignment projector and transformer decoder layers to mitigate spurious correlations (Fig.~\ref{fig:disentangled_architecture}).

\paragraph{Disentangled Visual Projector}  
\label{sec:projector}
The projector follows Eq. (6), where $ g_f $ inherits the original MLP structure (e.g., LLaVA's projector). To approximate $ \mathbb{E}_z[z] $, we build a visual confounder dictionary $ D \in \mathbb{R}^{K \times \sigma} $, with $ K $ as the number of object classes and $ \sigma $ as the hidden size. Each entry in $ D $ is the average post-projector visual representation of an object category, aggregated from 5,000 instances sampled from the LLaVA-Instruct dataset.

The expectation $ \mathbb{E}_z[z] $ is approximated via cross-attention:
\begin{equation}
    \mathbb{E}_{Z}[Z] \overset{\text{modeled as}}{\approx} \mathrm{CrossAttn}(X, D, D) 
    = \mathrm{softmax}\left( \frac{X(D W_k)^\top}{\sqrt{\sigma}} \right) (D W_v),
    \tag{7}
\end{equation}
where $ X $ is visual encoding features, and $ W_k, W_v \in \mathbb{R}^{\sigma \times \sigma} $. This formulation allows fixed confounder vectors $ d \in D $ to be contextually modulated by $ X $, enabling input-specific confounding estimation while avoiding static representations \citep{Yang2020DeconfoundedIC,Liu2022ShowDA}.

The final Causal-Driven Projector combines the original structure with causal intervention:
\begin{equation}
    \mathbf{C.D.Projector}(F_v, Z) 
    = \underbrace{\mathrm{OriginalProjector}(f_v)}_{g_f(f_v)} 
    + \underbrace{\mathrm{softmax}\left( \frac{f_v (D W_k)^\top}{\sqrt{\sigma}} \right) (D W_v) \cdot W_o}_{g_z(\mathbb{E}_z[z])},
    \tag{8}
\end{equation}
where $ W_o \in \mathbb{R}^{\sigma \times \sigma} $ as $ g_z $ in Eq. (6) enhances representation capacity.

\paragraph{Disentangled LLM Transformer}
\label{sec:transformer}
Using NWGM (Eq. 3, 5), we approximate $ P(W \mid \text{do}(h)) \approx P(W \mid h, \mathbb{E}_{d_v}[d_v], \mathbb{E}_{d_t}[d_t]) $. The expectations $ \mathbb{E}_{d_v}[d_v] $ and $ \mathbb{E}_{d_t}[d_t] $ are computed via cross-attention with visual/textual confounder dictionaries $ D_v, D_t \in \mathbb{R}^{K \times \sigma} $ (following the Disentangled Projector's procedure in Eq. 7). These dictionaries are built by aggregating object-category-specific features from 5,000 LLaVA-Instruct samples, similar as \cref{sec:projector}.

The causal intervention module for LLM layers is defined as:
\begin{equation}
    \mathbf{CausalIntervention}(h) 
    = \underbrace{\mathrm{CrossAttn}(h, D_v,D_v)}_{\text{visual}} 
    + \underbrace{\mathrm{CrossAttn}(h, D_t,D_t)}_{\text{textual}},
    \tag{9}
\end{equation}
where each cross-attention component expands to:
\begin{equation}
\mathrm{CrossAttn}(h, D) = \mathrm{softmax}\left( \frac{h(D W_k)^\top}{\sqrt{\sigma}} \right) (D W_v) \cdot W_o,\tag{10}
\end{equation}
with $ W_k, W_v, W_o \in \mathbb{R}^{\sigma \times \sigma} $. This formulation decouples visual and textual confounders while preserving input-specific modulation through dynamic attention weights.
\section{Experiments}

\subsection{Experiment Setup}

\textbf{Baselines} We adopt LLaVA \citep{Liu2023VisualIT,Liu2023ImprovedBW} as our primary baseline framework for causal learning evaluation. This choice is motivated by three key advantages: (1) its synthetic instruction data generation pipeline from MSCOCO with quantifiable instruction biases, (2) its fully tunable architecture enabling multi-level model intervention capabilities, and (3) its reproducible, modular open-source implementation that serves as a representative Multimodal Large Language Model (MLLM) prototype.

For comparative analysis, we include results from established frameworks: BLIP-2\citep{Li2023BLIP2BL}, InstructBLIP\citep{Dai2023InstructBLIPTG}, MiniGPT-4\citep{Zhu2023MiniGPT4EV}, Shikra\citep{Chen2023ShikraUM}, IDEFICS\citep{IDEFICS2023}, and Qwen-VL\citep{Bai2023QwenVLAV}.

\textbf{Causal-LLaVA Implementation Details}   
Our causal learning framework builds upon LLaVA~\citep{Liu2023VisualIT,Liu2023ImprovedBW}, integrating the Causal Disentanglement Module into the projection layer and final transformer block (\cref{sec:disentangled}). We maintain LLaVA's official training configuration with two critical modifications: For Causal-LLaVA-v1 pretraining, we implement (1). doubled batch size (256) and (2). halved learning rate ($1 \times 10^{-3}$) to accommodate additional projector parameters while ensuring training stability. All other phases retain default hyperparameters. The confounder dictionary estimation utilizes an intermediate checkpoint from a non-causal model trained for 0.1 epoch. Experiments were conducted on 8×NVIDIA H20 GPUs, with confounder estimation adding approximately 1 hour to total training duration.

\textbf{Evaluation Benchmarks} We evaluate our model across 10 mainstream benchmarks, including CHAIR\citep{Rohrbach2018ObjectHI}\citep{Zhai2023HallEControlCO}, POPE\citep{Li2023EvaluatingOH}, MME\citep{Fu2023MMEAC} for hallucination evaluation and VQA-v2\citep{Goyal2016MakingTV}, GQA\citep{Hudson2019GQAAN}, VizWiz\citep{Gurari2018VizWizGC}, Text-QA\citep{Singh2019TowardsVM}, ScienceQA-IMG\citep{Lu2022LearnTE}, MMBench\citep{Liu2023MMBenchIY} and MM-Vet\citep{Yu2023MMVetEL} for general comprehension evaluation.

\subsection{Quantitative Results}

\textbf{Effectiveness of Causal-LLaVA on Mitigating Hallucination}  
As shown in Table~\ref{tab:hallucination}, quantitative results demonstrate the effectiveness of causal intervention in reducing hallucinations. On POPE benchmarks, Causal-LLaVA consistently surpasses LLaVA baselines. For instance, on POPE$_\text{rnd}$, LLaMA-2-7B and LLaMA-2-13B versions of Causal-LLaVA yield improvements of \textbf{+1.42\%}(71.70~$\rightarrow$~72.72) and \textbf{+1.20\%} (78.60~$\rightarrow$~79.54), respectively. While gains on other POPE subsets are modest, they reflect stable improvements. CHAIR metrics show meaningful reductions in hallucination: CHAIR$_s$ drops by \textbf{6.36\%} (33.0~$\rightarrow$~30.9) and CHAIR$_i$ by \textbf{3.16\%} (9.5~$\rightarrow$~9.2) for LLaVA. Importantly, MME$^P$ improves by \textbf{6.0\%} (714.29~$\rightarrow$~757.16) for LLaMA-2-7B and \textbf{22.6\%} (711.22~$\rightarrow$~872.09) for LLaMA-2-13B, confirming that disentangled representations significantly suppress perceptual hallucination while maintaining overall performance stability.

\begin{table*}[t]
\centering
\footnotesize
\caption{Hallucination Evaluation Results on POPE, MME$^P$, and CHAIR. CHAIR metrics are computed on COCO val2014. Abbreviations: POPE$_{rnd}$ (Random), POPE$_{pop}$ (Popular), POPE$_{adv}$ (Adversarial), MME$^P$ (Perception).}
\label{tab:hallucination}
\setlength{\tabcolsep}{2pt}
\begin{tabular}{@{}l|c|cccccccc@{}}
\toprule
\textbf{Model} & \textbf{LLM} & 
\textbf{POPE$_{rnd}$} $\mathclap{\uparrow}$ & 
\textbf{POPE$_{pop}$} $\mathclap{\uparrow}$ & 
\textbf{POPE$_{adv}$} $\mathclap{\uparrow}$ & 
\textbf{MME$^P$} $\mathclap{\uparrow}$ & 
\textbf{CHAIR$_{s}$} $\mathclap{\downarrow}$ & 
\textbf{CHAIR$_{i}$} $\mathclap{\downarrow}$ & 
\textbf{Recall} $\mathclap{\uparrow}$ \\
\midrule
MiniGPT-4 & LLaMA-7B & 43.35 & 50.74 & 48.07 & 581.67 & - & - & - \\
mPLUG-Owl & LLaMA-7B & 68.39 & 66.94 & 66.82 & 967.34 & - & - & - \\
InstructBLIP & Vicuna-7B & 89.27 & 84.66 & 77.32 & 1084.0 & - & - & - \\
Shikra & Vicuna-7B & 86.19 & 83.16 & 82.49 & - & - & - & - \\
\midrule
LLaVA & LLaMA-2-7B & 71.70 & 67.15 & 67.21 & 714.29 & 33.0 & 9.5 & 65.6 \\
Causal-LLaVA & LLaMA-2-7B & 72.72 & 67.92 & 67.48 & 757.16 & 30.9 & 9.2 & 65.9 \\
LLaVA & LLaMA-2-13B  & 78.60 & 74.09 & 70.07 & 711.22 & 30.3 & 8.7 & 65.9 \\
Causal-LLaVA & LLaMA-2-13B & 79.54 & 75.21 & 70.81 & 872.09 & 28.2 & 8.5 & 66.1 \\
LLaVA1.5 & Vicuna-v1.5-7B & 87.34 & 86.13 & 84.21 & 1508.51 & 52.1 & 14.9 & 79.4 \\
Causal-LLaVA1.5 & Vicuna-v1.5-7B & 88.18 & 86.65 & 84.55 & 1522.10 & 51.4 & 14.8 & 79.9 \\
\bottomrule
\end{tabular}
\end{table*}

\textbf{Effectiveness of Causal-LLaVA on Visual Comprehension}  
As shown in Table~\ref{tab:comprehension}, Causal-LLaVA maintains or improves visual reasoning capabilities across multiple benchmarks. On Vicuna-v1.5-7B, it achieves \textbf{+2.0\%} on MMBench (64.6~$\rightarrow$~65.9) and \textbf{+4.8\%} on MM-Vet (31.2~$\rightarrow$~32.7). Additional gains are observed on GQA (\textbf{+2.7\%}, 37.3~$\rightarrow$~38.3) and VizWiz (\textbf{+8.4\%}, 44.2~$\rightarrow$~47.9) with LLaMA-2-13B. While improvements on TextVQA and ScienceQA are marginal, Causal-LLaVA consistently preserves or slightly enhances performance across tasks, indicating that causal disentanglement improves semantic grounding without compromising general comprehension.

\begin{table*}[t]
\centering
\footnotesize
\caption{General Visual Comprehension Evaluation Results on MMBench, MM-Vet, VQA$^{v2}$, GQA, VizWiz, TextVQA, and ScienceQA. Abbreviations: MMB (MMBench), MMV (MM-Vet).}
\label{tab:comprehension}
\setlength{\tabcolsep}{3pt}
\begin{tabular}{@{}l|c|ccccccc@{}}
\toprule
\textbf{Model} & \textbf{LLM} & 
\textbf{MMB} $\mathclap{\uparrow}$ & 
\textbf{MMV} $\mathclap{\uparrow}$ & 
\textbf{VQA$^{v2}$} $\mathclap{\uparrow}$ & 
\textbf{GQA} $\mathclap{\uparrow}$ & 
\textbf{VizWiz} $\mathclap{\uparrow}$ & 
\textbf{TextVQA} $\mathclap{\uparrow}$ & 
\textbf{SciQA} $\mathclap{\uparrow}$ \\
\midrule
MiniGPT-4 & Vicuna-7B & 23.0 & 22.1 & 65.2 & 30.08 & 30.2 & 52.3 & 58.4\\
BLIP-2 & Vicuna-7B & - & 22.4 & 65.0 & 41.0 & 19.6 & 42.5 & 61.0\\
InstructBLIP & Vicuna-7B & 36.0 & 26.2 & - & 49.2 & 34.5 & 50.1 & 60.5\\
IDEFICS & LLaMA-7B & 48.2 & - & 50.9 & 38.4 & 35.5 & - & - \\
Qwen-VL-Chat & Qwen-7B & 60.6 & - & 78.2 & 57.5 & 38.9 & 61.5 & 68.2\\
\midrule
LLaVA & LLaMA-2-7B & 39.0 & 26.2 & 67.6 & 37.3 & 43.8 & 52.0 & 61.5 \\
Causal-LLaVA & LLaMA-2-7B & 39.2 & 29.1 & 68.3 & 37.7 & 43.9 & 52.0 & 62.5\\
LLaVA & LLaMA-2-13B & 37.5 & 32.2 & 68.1 & 37.3 & 44.2 & 55.2 & 65.1\\
Causal-LLaVA & LLaMA-2-13B & 40.3 & 33.5 & 70.4 & 38.3 & 47.9 & 55.6 & 65.0\\
LLaVA1.5 & Vicuna-v1.5-7B & 64.6 & 31.2 & 75.1 & 62.5 & 50.1 & 61.0 & 70.2 \\
Causal-LLaVA1.5 & Vicuna-v1.5-7B & 65.9 & 32.7 & 75.3 & 63.2& 49.5 & 61.1 & 70.3\\
\bottomrule
\end{tabular}
\end{table*}

\subsection{Disentanglement Visualization} 

\begin{figure}[t]  
  \centering    
  \includegraphics[width=0.32\linewidth]{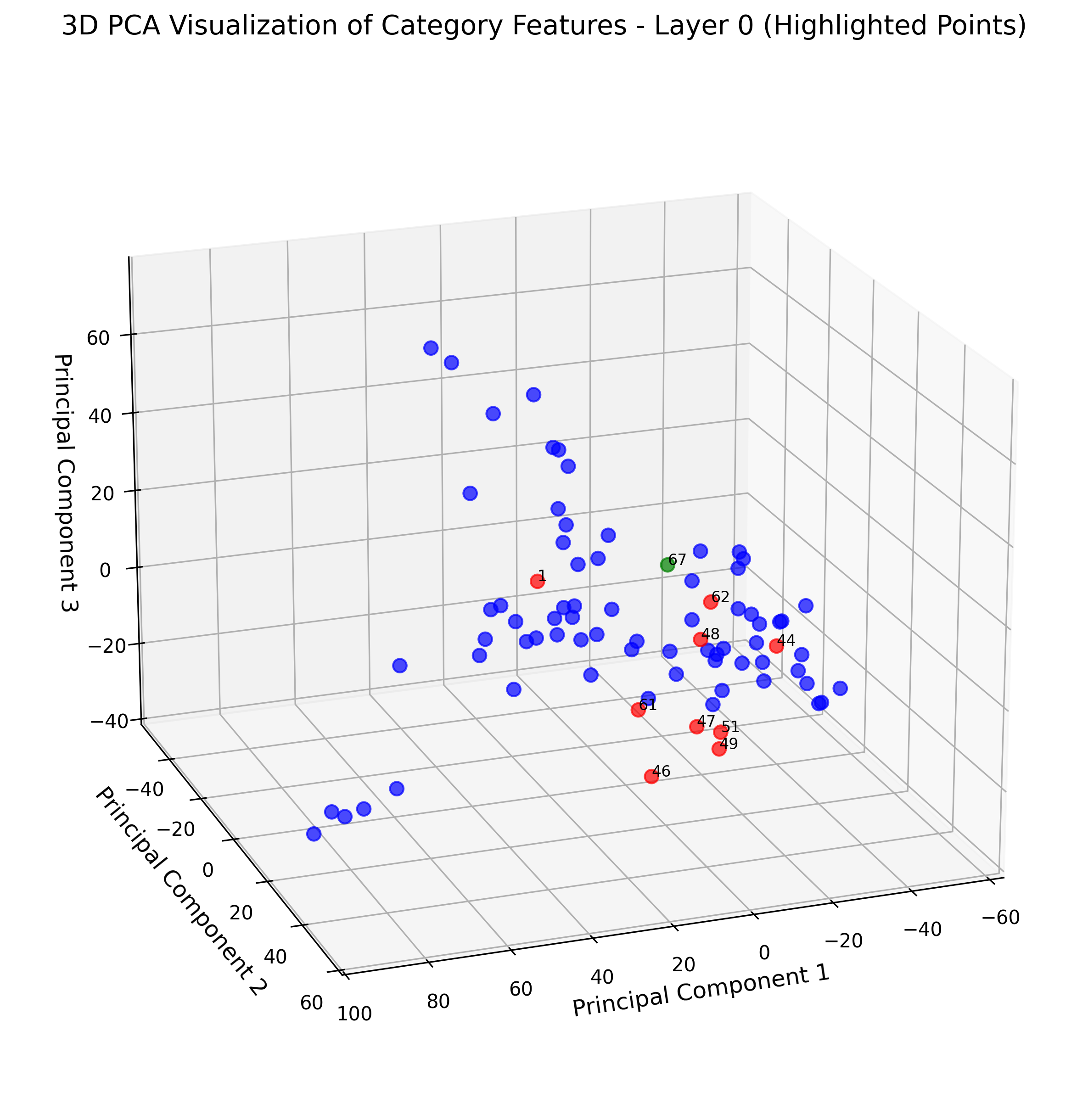}  
  \hfill  
  \includegraphics[width=0.32\linewidth]{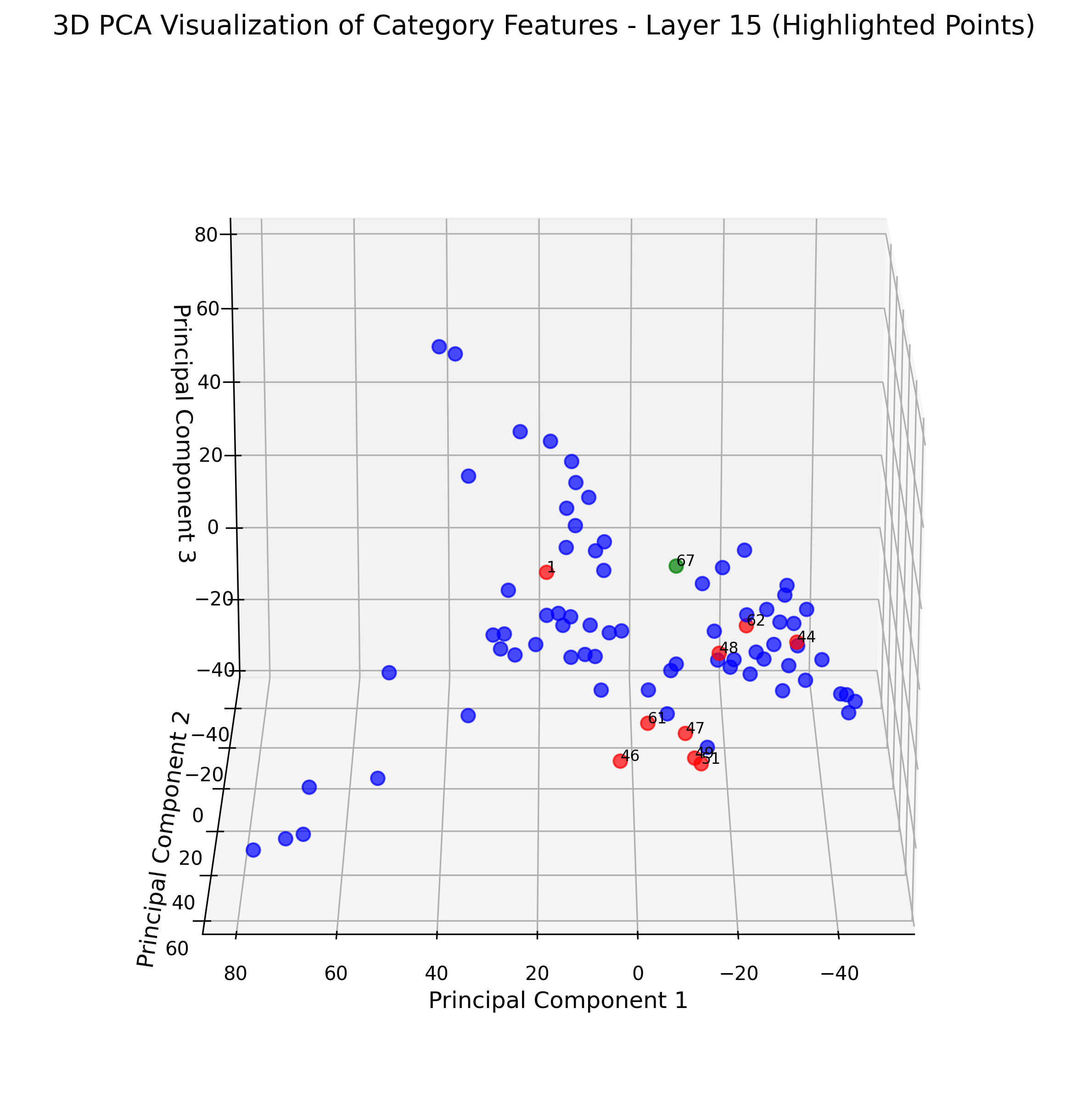}  
  \hfill  
  \includegraphics[width=0.32\linewidth]{./Paper_Draft_assets/disentangled/3d_PCA_layer40_elev20_azim70.png}  
  \vspace{-0.3cm}  
  \caption{Semantic distributions PCA after disentanglement at three stages: Projector output (left), LLM layer 15 (middle), and LLM layer 40 (right). Green dots: "dining table"; red dots: top-10 co-occurring objects.}  
  \label{fig:disentanglement}  
\end{figure} 

\textbf{Qualitative Analysis of Disentangled Representations}  
Building on the entangled patterns observed in earlier stages (Fig.~\ref{fig:visualization-1} and Fig.~\ref{fig:visualization-2}), Figure~\ref{fig:disentanglement} visualizes the progressive separation of object representations through our framework. For "dining table" (green dot) and its co-occurring objects (red dots), the Causal-Driven Projector begins to spread them apart. By layer 15, they are no longer close in space, and by layer 40, they are fully separated.  This progression confirms that dual-path causal intervention effectively disrupts bias propagation. 

\section{Conclusion}

In conclusion, this paper uncovers an essential cause of object hallucinations in MLLMs—entangled semantic representations driven by dataset object co-occurrence biases—and proposes a novel causality-based disentanglement framework to mitigate this issue. With a Causal-Driven Projector and a Causal Intervention Module in the final LLM layer, our approach reduces hallucinations while maintaining strong performance across general comprehensive benchmarks, offering both theoretical insights and a practical solution for enhancing MLLM reliability.

\section{Limitations}
The primary limitation is substantial computational resource requirements, specifically utilizing 8 NVIDIA H20 GPUs for model training, causing relatively high requirements for computing resources. Meanwhile, the confounders may be affected by noisy potential variables, which are related with different environments and different datasets.

\small
\bibliographystyle{plainnat}
\bibliography{references}
\normalsize


\clearpage




{\Huge \section*{Supplementary Material}} 
\addcontentsline{toc}{section}{Supplementary Material}

In this supplementary document, we present additional material about our causal disentanglement framework, including \textbf{A.} Formula Derivations of Causal Intervention with NWGM approximation, \textbf{B.} Ablation Experiments, and \textbf{C.} More Visualization Results.

\section*{A. Formula Derivations of Causal Intervention with NWGM approximation}

The methodology presented in our primary work incorporates the Normalized Weighted Geometric Mean (NWGM) approximation\citep{Wang2020VisualCR,Xu2015ShowAA} to evaluate Eq.~(2) from Section 4.1 and Eq.~(4) from Section 4.2. This section outlines the comprehensive derivations.

To demonstrate the application of NWGM in relocating the outer expectation within the Softmax operator shown in Eq.~(3) and Eq.~(5), we first define the Weighted Geometric Mean (WGM) of a function $y(x)$ as follows\citep{Xu2015ShowAA}:
\begin{equation*}
    \text{WGM}(y(x)) = \prod_x y(x)^{P(x)}, \tag{11}
\end{equation*}
where $P(x)$ denotes the probability distribution over $x$. When $y(x)$ takes an exponential form, specifically $y(x) = \exp[f(x)]$, Eq.~(12) transforms into:
\begin{align*}
    \text{WGM}(y(x)) &= \prod_x \exp[f(x)]^{P(x)} \\
                     &= \prod_x \exp[f(x)P(x)] \\
                     &= \exp\left[\sum_x f(x)P(x)\right] \\
                     &= \exp\{\mathbb{E}_x[f(x)]\}. \tag{12}
\end{align*}
This allows us to approximate the expectation of $y(x)$ through:
\begin{equation*}
    \mathbb{E}_x[y(x)] = \sum_x y(x)P(x) \approx \text{WGM}(y(x)) = \exp\{\mathbb{E}_x[f(x)]\}, \tag{13}
\end{equation*}
given $y(x) = \exp[f(x)]$.

Within our Disentangled LLM Transformer framework, $P(Y | \text{do}(X))$ serves as the predictive distribution for subsequent tokens, modeled using the final linear word embedding layer $g$ followed by a Softmax layer:
\begin{equation*}
    P(Y | X, Z) = \text{Softmax}[g(X, Z)] \propto \exp[g(X, Z)]. \tag{14}
\end{equation*}
Combining Eq.~(2), Eq.~(12), and Eq.~(14), the following derivation holds:
\begin{align*}
    P(Y | \text{do}(X = x)) &= \sum_Z P(Y | X = x, Z = z)P(Z = z) \\
                            &= \mathbb{E}_{Z}[P(Y | Z = z, X = x)] \\
                            &\approx \text{WGM}(P(Y | Z = z, X = x)) \\
                            &\propto \exp\{[g(\mathbb{E}_Z[Z], x)]\}. \tag{15}
\end{align*}
To ensure the resulting probabilities sum to 1, Eq.~(15) undergoes normalization through the NWGM approximation:
\begin{equation*}
P(Y | \text{do}(X = x)) \approx \frac{\exp\{g(\mathbb{E}_Z[Z], x)\}}{\sum_{y'} \exp\{g(\mathbb{E}_Z[Z], x)_{y'}\}} = \text{Softmax}[g(\mathbb{E}_Z[Z], x)].
\tag{16}    
\end{equation*}

Assuming $g$ exhibits linearity, it can be decomposed into:
\begin{equation*}
    P(Y | \text{do}(X = x)) \approx \text{Softmax}[g_x(x) + g_z(\mathbb{E}_Z[Z])]. \tag{17}
\end{equation*}

Extending to the Disentangled LLM Transformer architecture, consider a combined final hidden state $h$ associated with word $w$, alongside estimated visual confounder $D_v$ and textual confounder $D_t$. Analogous to Eq.~(17), $P(w|\text{do}(h))$ admits the approximation:
\begin{equation*}
    P(W | \text{do}(h)) \approx \text{Softmax}[g_h(h) + \mathbb{E}_{D_v}[g_v(D_v)] + \mathbb{E}_{D_t}[g_t(D_t)]]. \tag{18}
\end{equation*}
Here, $g_h$ retains the structure of the standard LLM transformer. Critically, when computing $\mathbb{E}_{D}[g_d(D)]$ for both $D_v$ and $D_t$, we condition $d$ on $h$ via $\mathbb{E}_{[D|h]}[g_d(D)]$ (refer to \cref{sec:transformer}). This prevents the expectation from collapsing to a static vector, a strategy validated in~\cite{Kingma2014AdamAM, Maaten2008VisualizingDU, Liu2022ShowDA} to enhance model representational capacity.

\section*{B. Ablation Experiments}

\paragraph{Projection Matrix Selection in Causal Disentangle Cross-Attention Module}  
We first investigate the impact of different projection matrix configurations in the causal disentangled cross-attention module, using LLaVA-7B pretraining as a case study. We find inclusion of Wv/Wo is critical for regulating feature scaling and preventing significant loss fluctuations. As shown in Table~\ref{tab:projection_ablation}, comparative experiments under these constraints yield key insights:  

\begin{table}[h]  
\centering  
\caption{Ablation study on projection matrix configurations. CHAIR metrics are computed on COCO val2014.}  
\label{tab:projection_ablation}  
\begin{tabular}{c|ccc}  
\hline  
\textbf{Projection Type} & \textbf{CHAIR$_s$} $\downarrow$ & \textbf{CHAIR$_i$} $\downarrow$ & \textbf{Recall} $\uparrow$ \\  
\hline  
Original baseline & 29.5 & 8.6 & 50.4 \\  
Shared $W_k/W_v$ ($W_{kv}$) & 27.7 & \textbf{8.1} & \textbf{51.1} \\  
Independent $W_k/W_v$ & 28.1 & 8.3 & 50.8 \\  
Independent $W_q/W_o$ & \textbf{16.1} & 24.2 & 17.5 \\  
Independent $W_q/W_v$ & 17.3 & 24.8 & 16.9 \\  
\hline  
\end{tabular}  
\end{table} 
    
- Removing the causal disentanglement module entirely (original baseline) results in elevated hallucination scores (CHAIR$_s=29.5$, CHAIR$_i=8.6$) while maintaining moderate recall (50.4), demonstrating its essential role in mitigating spurious correlations .
    
- Shared $W_k/W_v$ parameterization (single $W_{kv}$) achieves improved performance (CHAIR$_s$=27.7, CHAIR$_i$=8.1, Recall=51.1) compared to independent $W_k/W_v$ variants. This validates our design choice to preserve causal disentanglement while enhancing training stability through parameter sharing, as evidenced by reduced CHAIR scores and improved recall.

- Independent $W_q/W_o$ and $W_q/W_v$ configurations exhibit significantly higher hallucination metrics (CHAIR$_i$: 24.2-24.8) coupled with degraded recall ($\leq $ 18.0). This indicates that improper confounder estimation disrupts semantic coherence, creating conflicting signals that degrade factual accuracy.

The shared $W_k/W_v$ configuration achieves an optimal trade-off between causal disentanglement and representational stability. By jointly mapping key/value pairs from the fixed confounder dictionary ($D$), this design prevents overfitting to spurious correlations while enabling efficient gradient propagation, particularly critical during limited pretraining stages.

\paragraph{Component-Level Causal Disentanglement Integration}  
We systematically evaluate the necessity of dual-path causal disentanglement integration across projector and transformer layers. As shown in Table~\ref{tab:component_ablation}, isolated implementation of the module in either component yields marginal improvements:  

\begin{table*}[t]
\centering
\footnotesize
\caption{Ablation study on causal module integration locations.}
\label{tab:component_ablation}
\setlength{\tabcolsep}{4pt}
\begin{tabular}{@{}c|cccccc@{}}
\toprule
\textbf{Component} & \textbf{CHAIR$_s$} $\mathclap{\downarrow}$ & \textbf{CHAIR$_i$} $\mathclap{\downarrow}$ & \textbf{POPE$_{rnd}$} $\mathclap{\uparrow}$ & \textbf{POPE$_{pop}$} $\mathclap{\uparrow}$ & \textbf{POPE$_{adv}$} $\mathclap{\uparrow}$ & \textbf{MME$^P$} $\mathclap{\uparrow}$ \\  
\midrule
Original baseline & 33.0 & 9.5 & 71.70 & 67.15 & 67.21 & 714.29 \\  
Only-transformer & 32.7 & 9.5 & 71.87 & 67.31 & 67.24 & 726.15 \\  
Only-projection & 31.9 & 9.3 & 72.24 & 67.51 & 67.33 & 748.70 \\  
Both-causal & \textbf{30.9} & \textbf{9.2} & \textbf{72.72} & \textbf{67.92} & \textbf{67.48} & \textbf{757.16} \\  
\bottomrule
\end{tabular}
\end{table*}

- \textit{Only-transformer} variant demonstrates minimal improvement over baseline (CHAIR$_s=32.7$ vs. 33.0), suggesting that late-stage intervention alone cannot rectify biased feature propagation originating from earlier stages.

- \textit{Only-projection} achieves significant MME$^P$ gains (748.70 vs. 714.29 baseline) but falls short of the full model (757.16). This highlights the complementary nature of dual-path intervention: while early disentanglement in the projector mitigates visual confounding, downstream transformer-level adjustments are essential to fully suppress bias amplification during contextual reasoning.

The full \textit{Both-causal} model achieves superior performance with the lowest hallucination rates, validating our dual-path causal intervention strategy. These results align with the structural causal model, where confounding effects originate from:  
1) Visual feature extraction ($S \gets D_v$): Early disentanglement in the projector blocks bias injection at the source  
2) Multimodal fusion ($h \gets D_v/D_t$): Transformer-level intervention eliminates residual correlations during contextualized reasoning (refer to \cref{sec:relationship_graph}).

\section*{C. More Visualization Results}

\subsection*{C.1. Table-Centric Original Visual Representation Visualizations}

We present extended PCA visualizations of LLaVA's original visual representations across different network depths. All visualizations follow the same color scheme as main text: green dots represent "dining table" tokens, while red dots indicate its top-10 co-occurring objects in instruction data.

\begin{figure}[t]
  \centering
  \includegraphics[width=0.32\linewidth]{./Paper_Draft_assets/3d_PCA_CLIP_elev20_azim60.png}
  \hfill
  \includegraphics[width=0.32\linewidth]{./Paper_Draft_assets/3d_PCA_layer0_elev20_azim70.png}
  \hfill
  \includegraphics[width=0.32\linewidth]{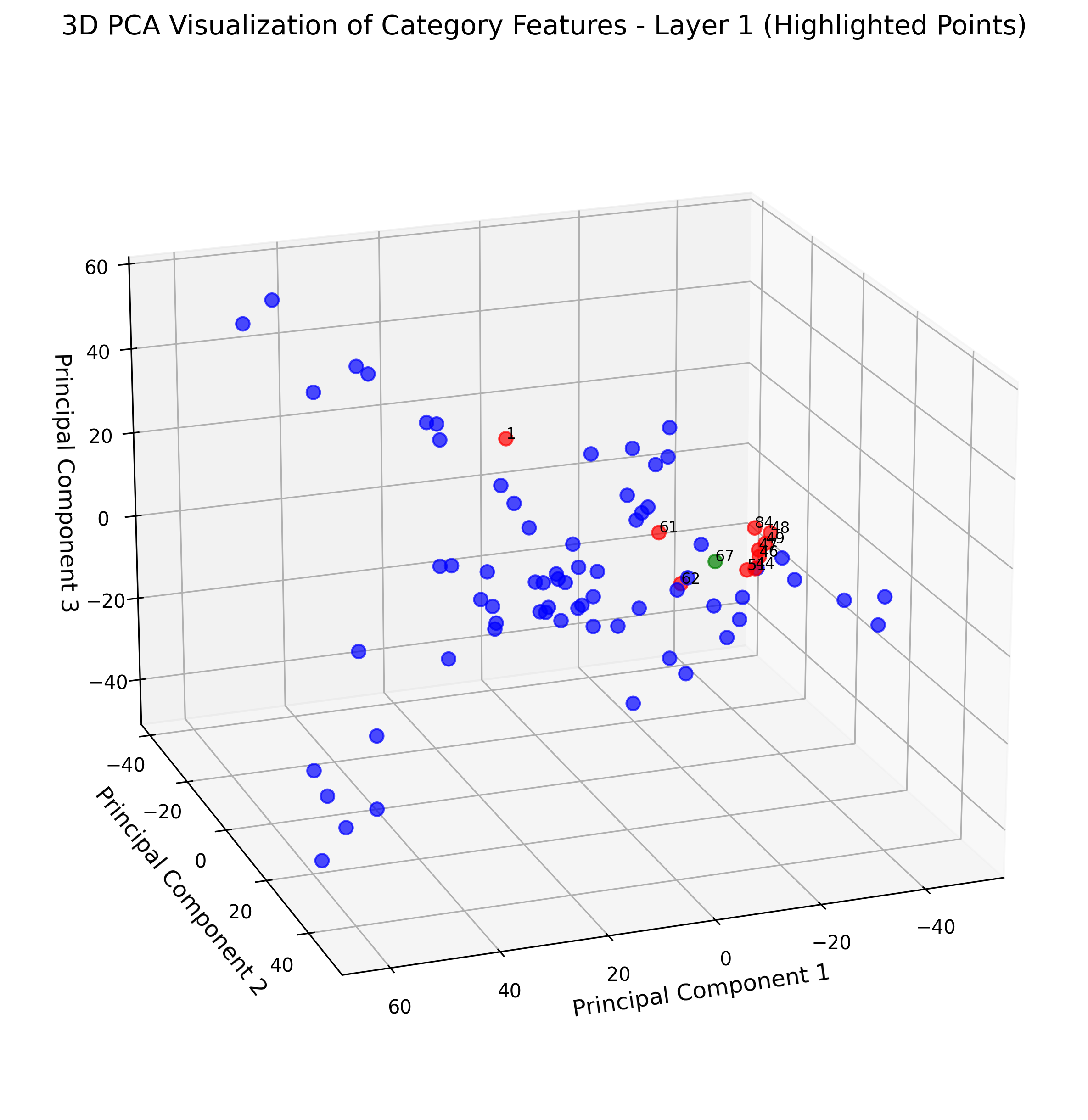}
  \caption{(Table-Centric:Visual) Layer-wise visual representations: CLIP Vision Encoder output (left), Projector Output (middle), and LLM layer 1 (right).}
  \label{fig:layer0_1}
\end{figure}

\begin{figure}[t]
  \centering
  \includegraphics[width=0.32\linewidth]{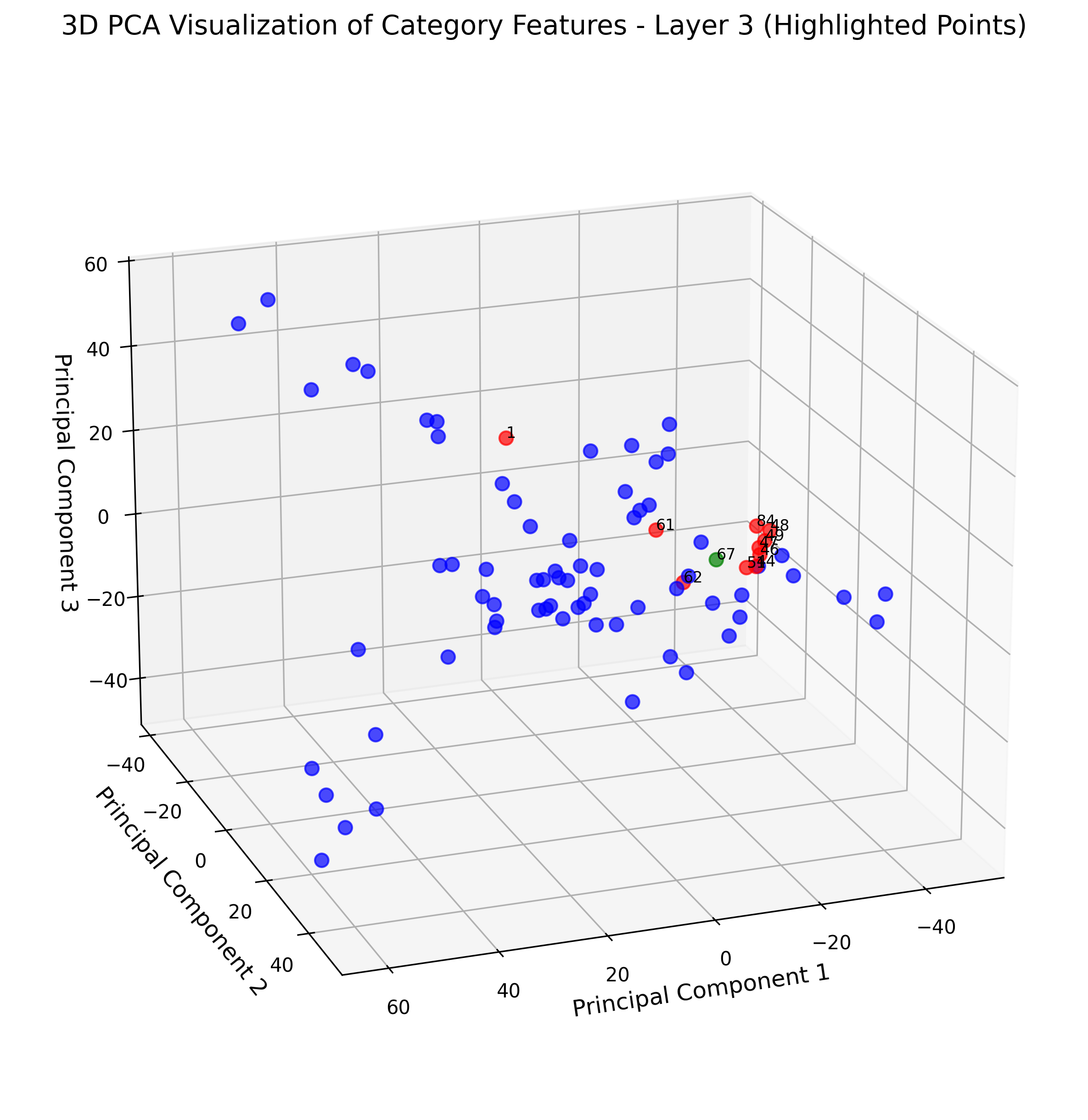}
  \hfill
  \includegraphics[width=0.32\linewidth]{./Paper_Draft_assets/3d_PCA_layer5_elev20_azim70.png}
  \hfill
  \includegraphics[width=0.32\linewidth]{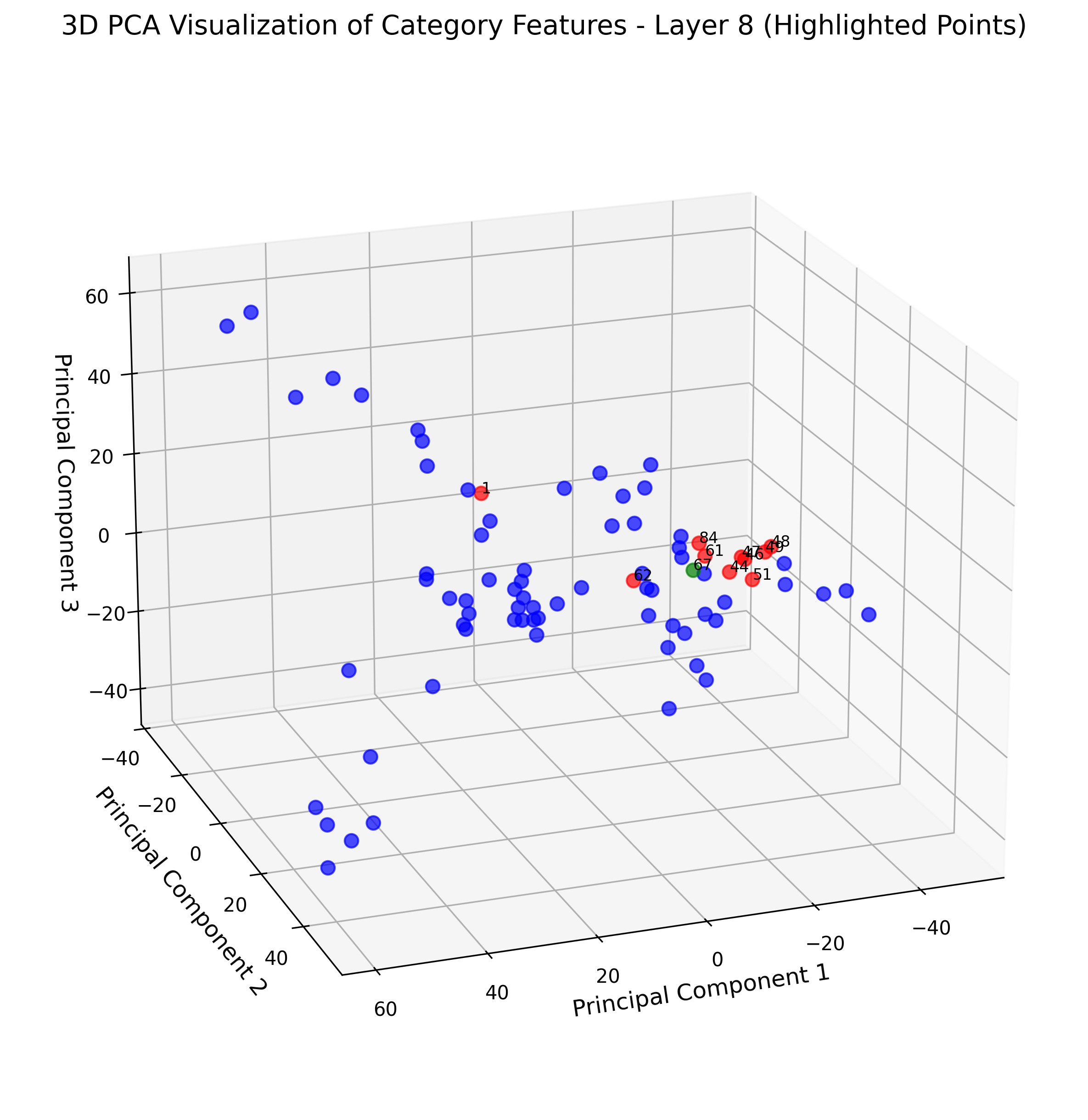}
  \caption{(Table-Centric:Visual) LLM layer 3 (left), layer 5 (middle), and layer 8 (right) visualizations showing progressive clustering formation.}
  \label{fig:layer3_5}
\end{figure}

\begin{figure}[t]
  \centering
  \includegraphics[width=0.32\linewidth]{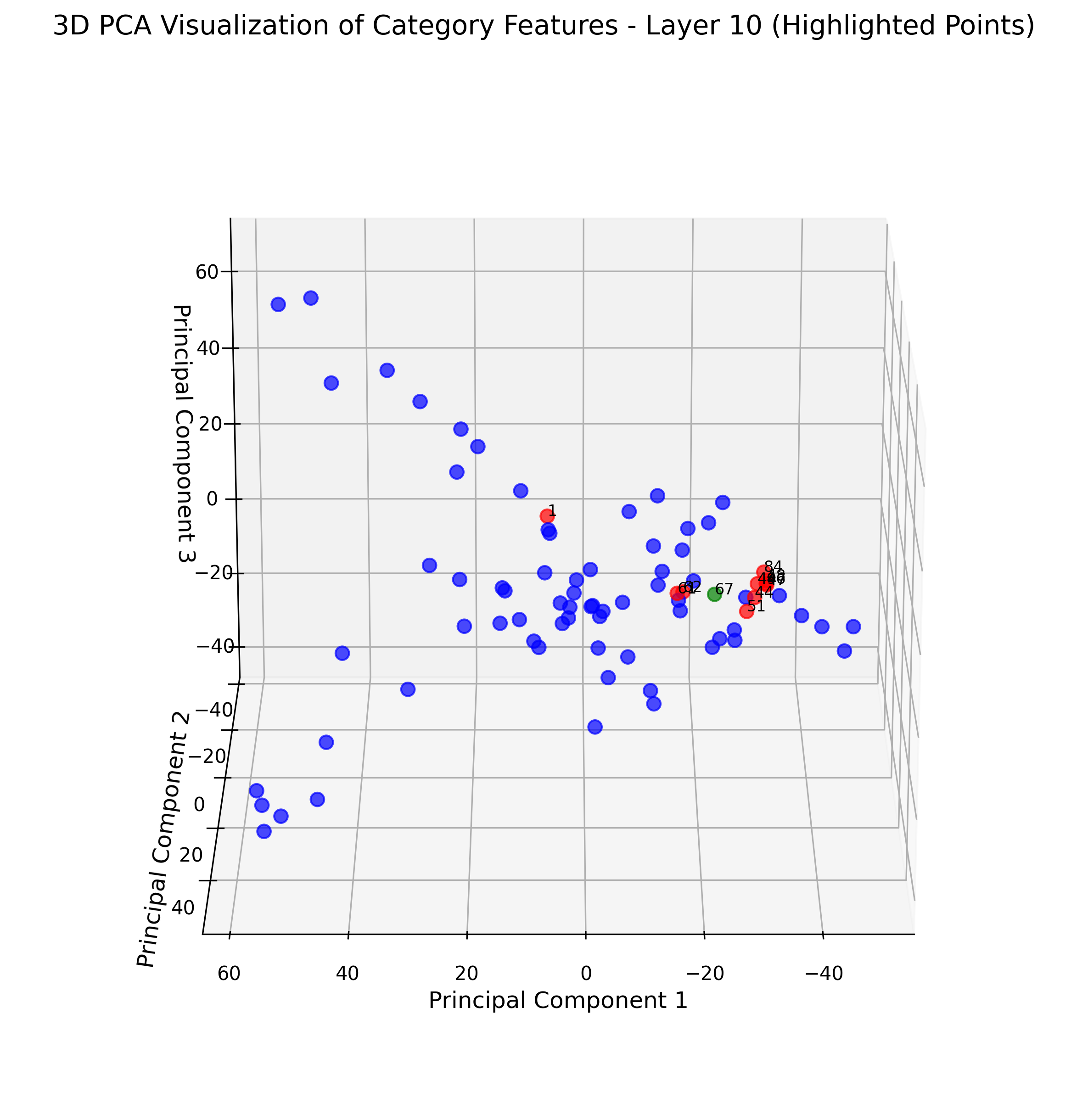}
  \hfill
  \includegraphics[width=0.32\linewidth]{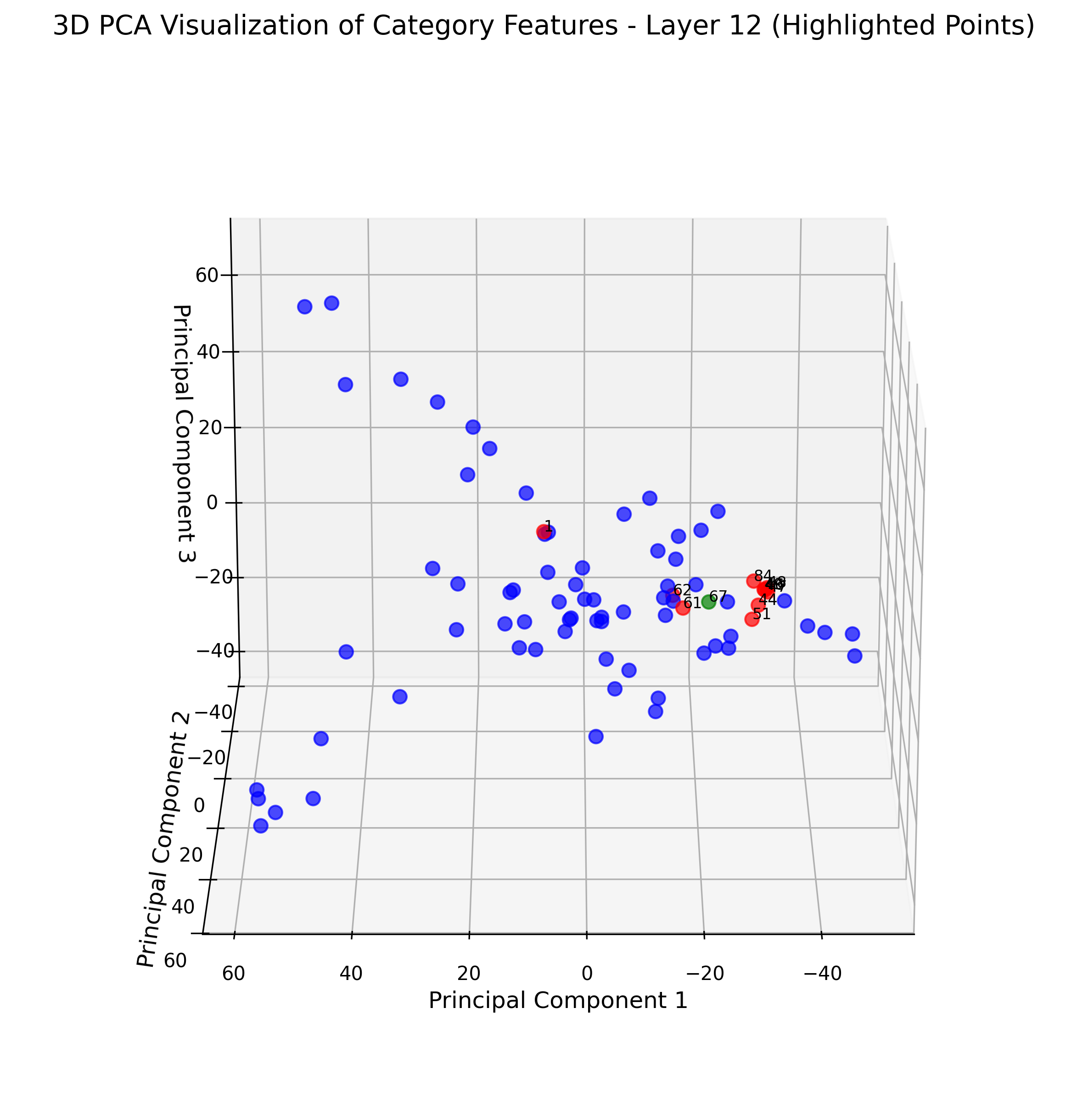}
  \hfill
  \includegraphics[width=0.32\linewidth]{./Paper_Draft_assets/3d_PCA_layer15_elev20_azim90.png}
  \caption{(Table-Centric:Visual) Mid-depth layers: layer 10 (left), layer 12 (middle), and layer 15 (right) maintaining strong co-occurrence patterns.}
  \label{fig:layer10_15}
\end{figure}

\begin{figure}[t]
  \centering
  \includegraphics[width=0.32\linewidth]{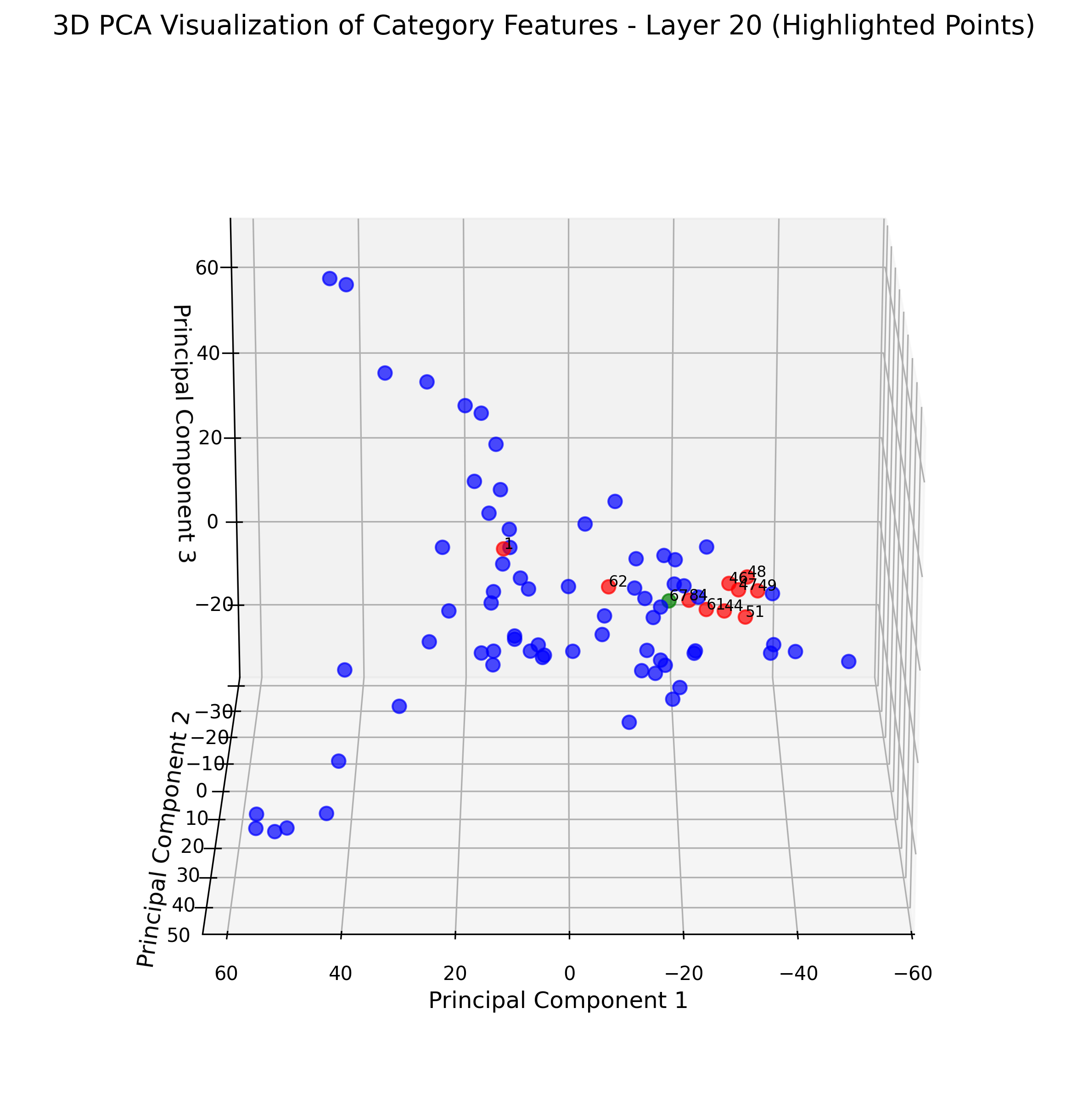}
  \hfill
  \includegraphics[width=0.32\linewidth]{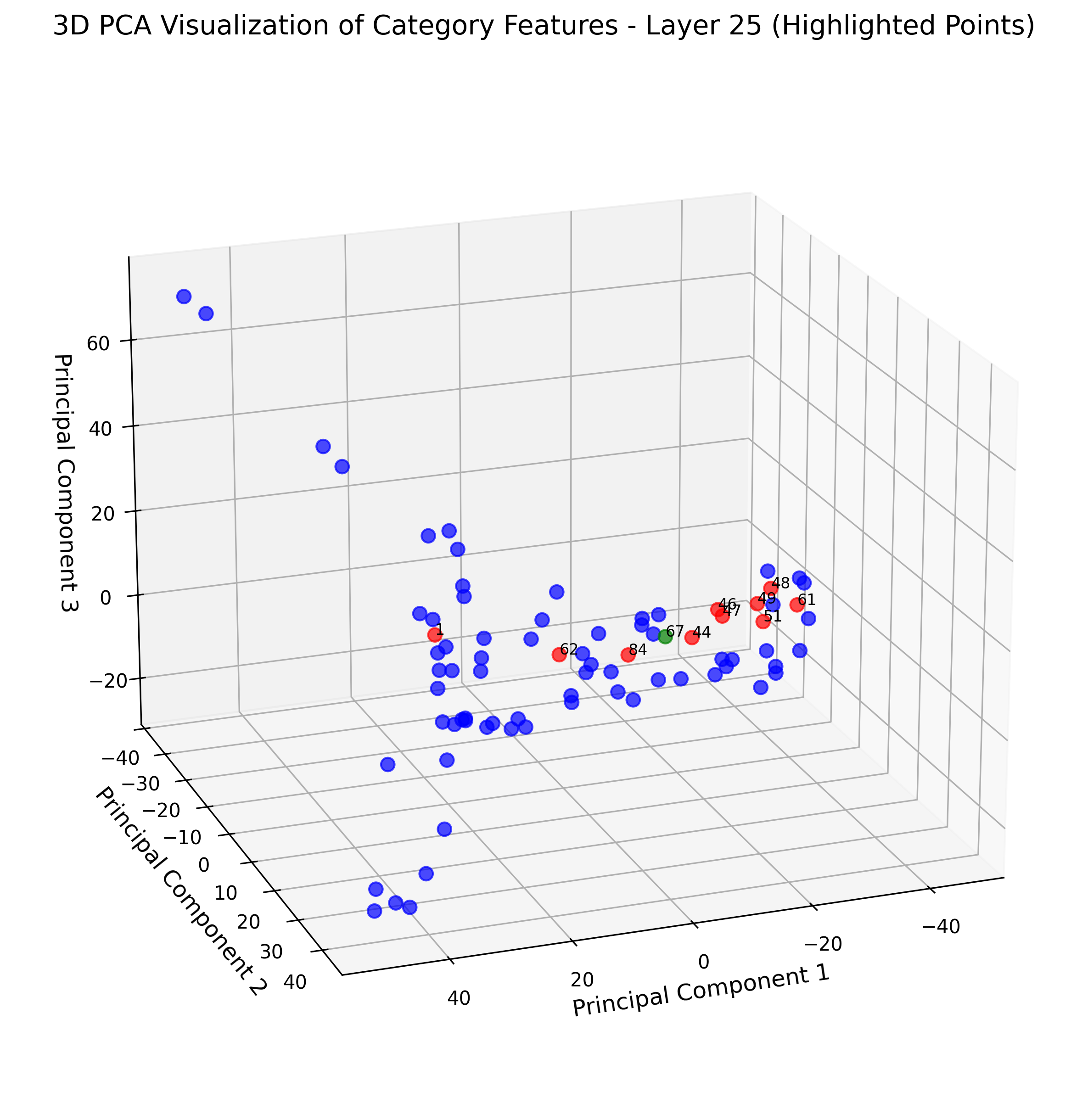}
  \hfill
  \includegraphics[width=0.32\linewidth]{./Paper_Draft_assets/3d_PCA_layer30_elev20_azim70.png}
  \caption{(Table-Centric:Visual) Transition phase layers: layer 20 (left), layer 25 (middle), and layer 30 (right) showing gradual dispersion of object clusters.}
  \label{fig:layer20_30}
\end{figure}

\begin{figure}[t]
  \centering
  \includegraphics[width=0.32\linewidth]{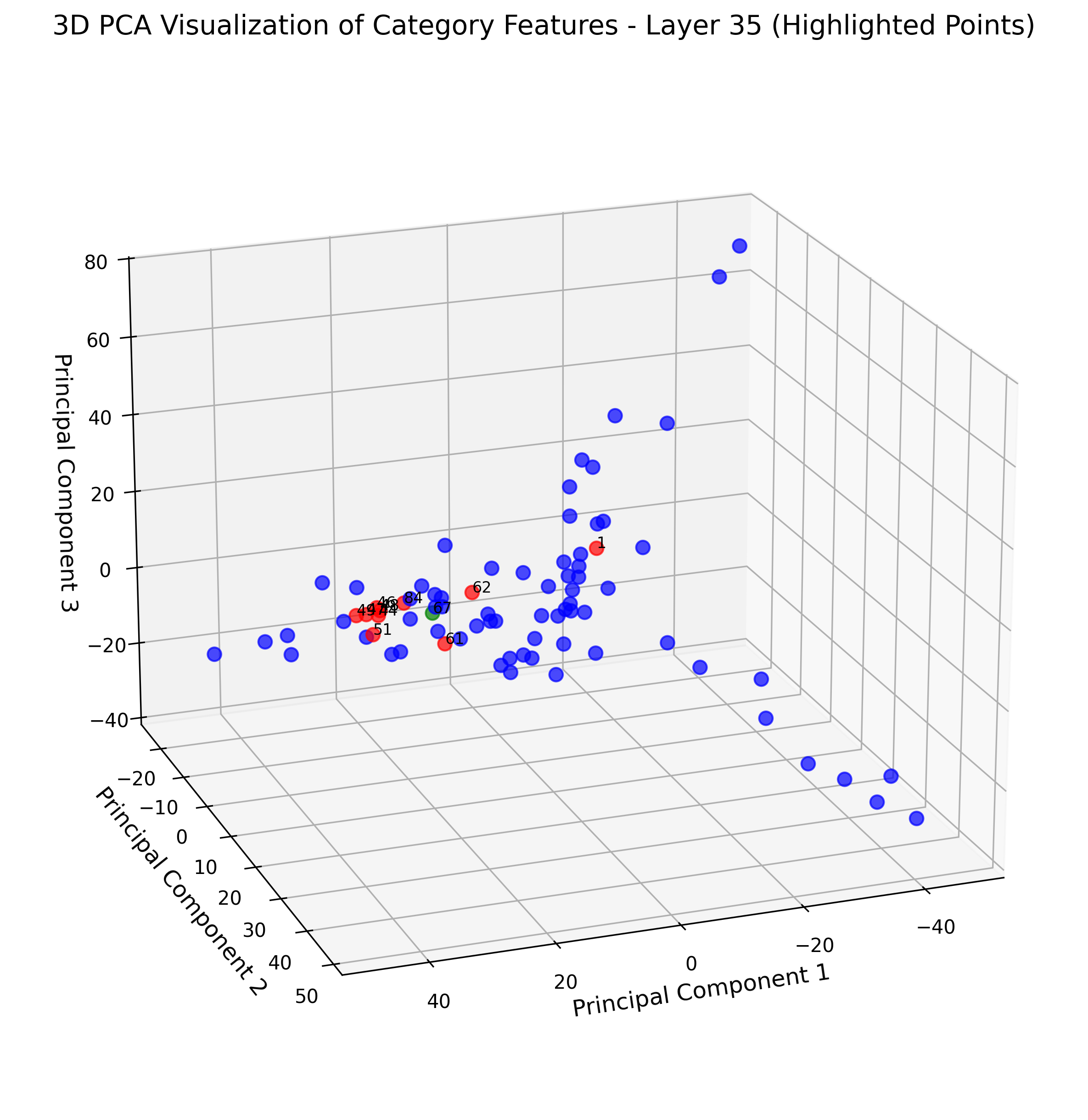}
  \hfill
  \includegraphics[width=0.32\linewidth]{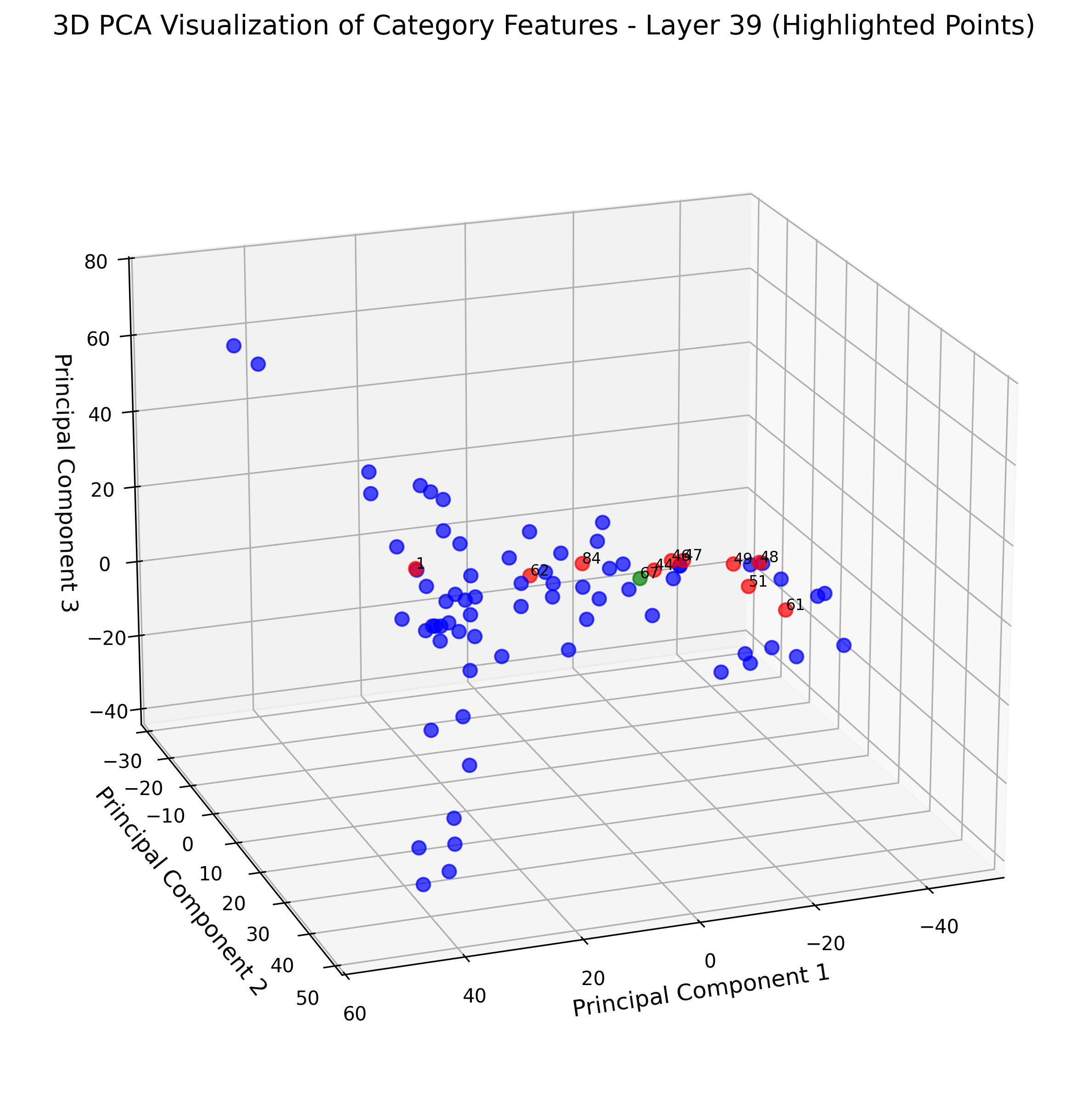}
  \hfill
  \includegraphics[width=0.32\linewidth]{./Paper_Draft_assets/3d_PCA_layer40_elev20_azim70.png}
  \caption{(Table-Centric:Visual) Final prediction layers: layer 35 (left), layer 39 (middle), and layer 40 (right) retaining core entanglement patterns.}
  \label{fig:layer35_40}
\end{figure}

Across all layers, we observe three key patterns:
\begin{enumerate}
  \item \textbf{Early-stage entanglement}: Layers 0-5 maintain tight clustering driven by projector-induced biases, consistent with main text findings (\cref{fig:layer0_1,fig:layer3_5}).
  
  \item \textbf{Mid-layer persistence}: Surprisingly, strong co-occurrence patterns persist until layer 25 despite increasing semantic abstraction (\cref{fig:layer10_15,fig:layer20_30}).
  
  \item \textbf{Final-layer retention}: Even at the final prediction stage (layer 40), dining tables retain significant entanglement with co-occurring objects (\cref{fig:layer35_40}), confirming the lasting impact of training data biases.
\end{enumerate}

These results reinforce our main conclusion that instruction data biases create persistent representational structures that survive through most stages of the vision-language understanding pipeline. The progressive visualization sequence provides empirical validation of how initial representational biases propagate through network depth, resisting the expected semantic disentanglement in later processing stages.

\subsection*{C.2. Table-Centric Original Textual Representation Visualizations}

\begin{figure}[t]
  \centering
  \includegraphics[width=0.32\linewidth]{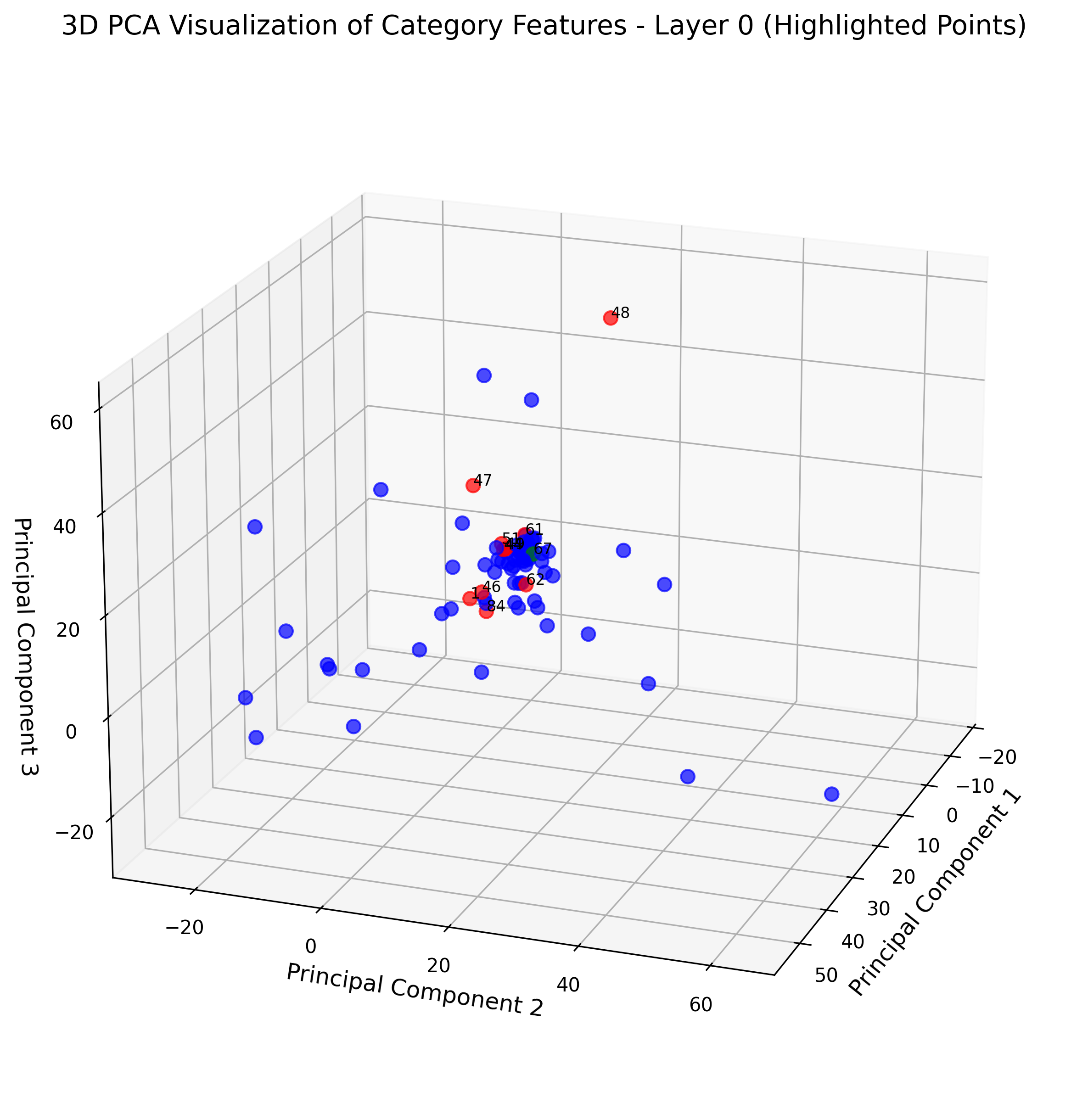}
  \hfill
  \includegraphics[width=0.32\linewidth]{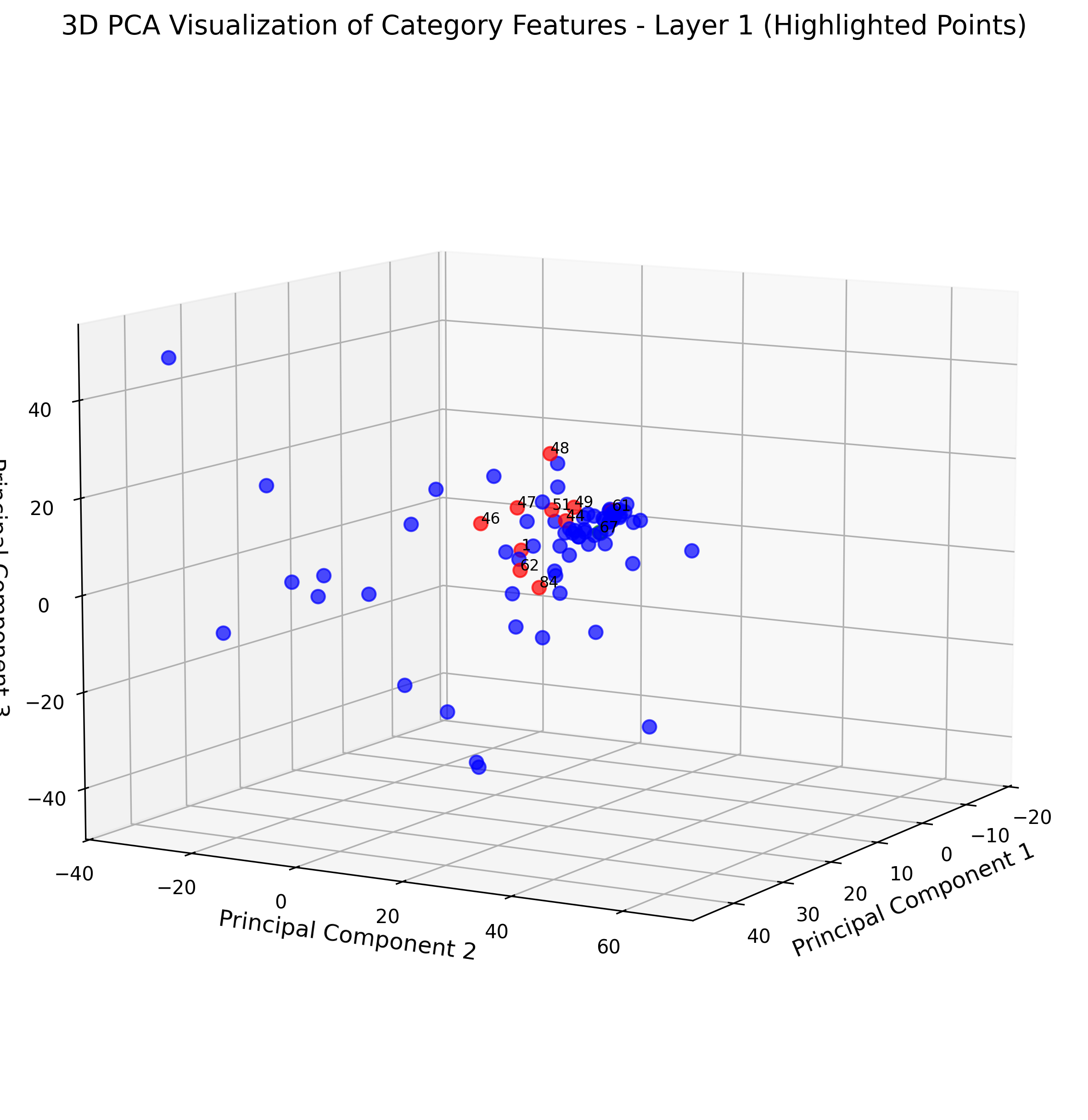}
  \hfill
  \includegraphics[width=0.32\linewidth]{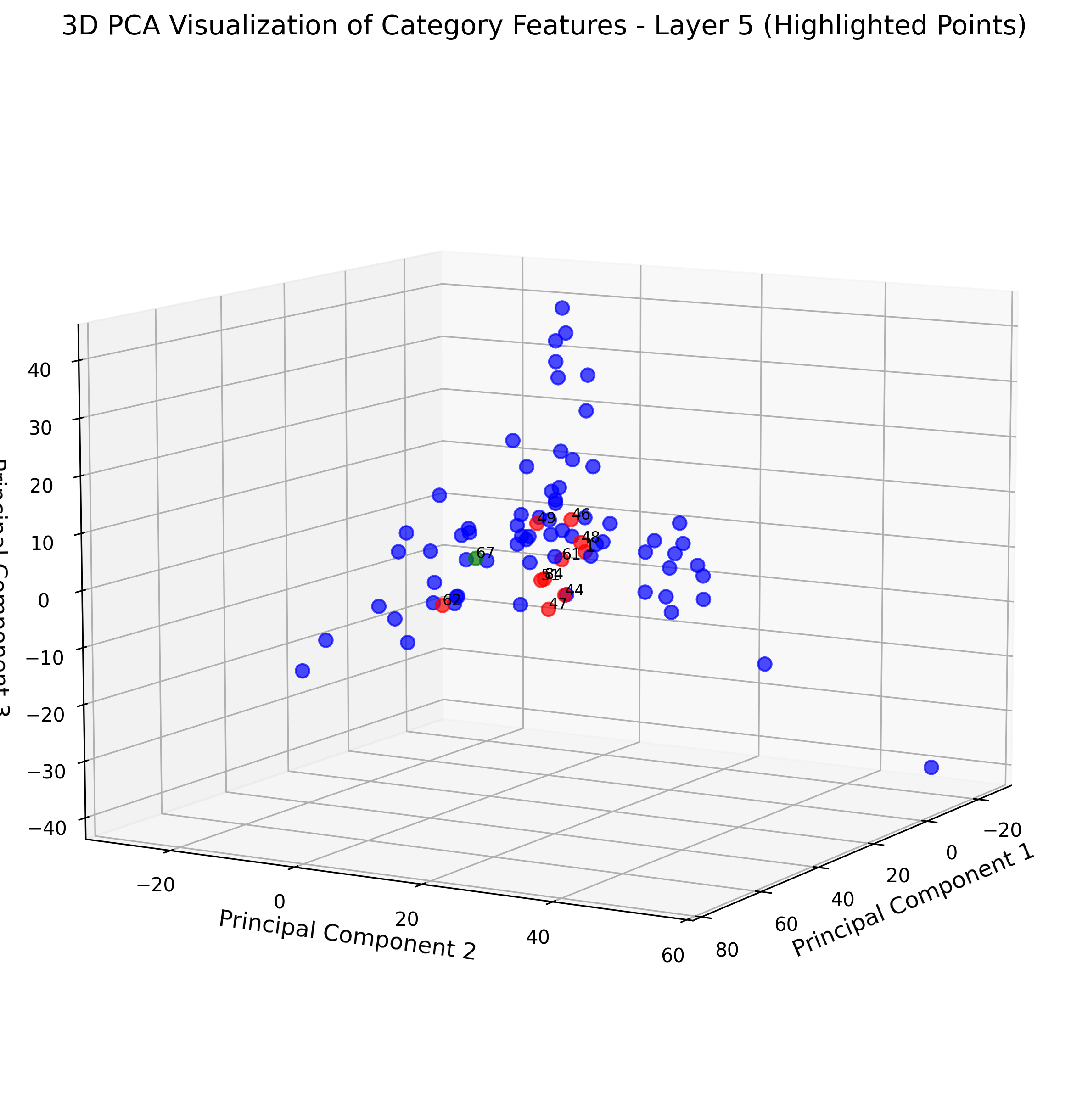}
  \caption{(Table-Centric:Textual) Textual representations after Projector (left), LLM layer 1 (middle), and layer 5 (right).}
  \label{fig:text_layer0_5}
\end{figure}

\begin{figure}[t]
  \centering
  \includegraphics[width=0.32\linewidth]{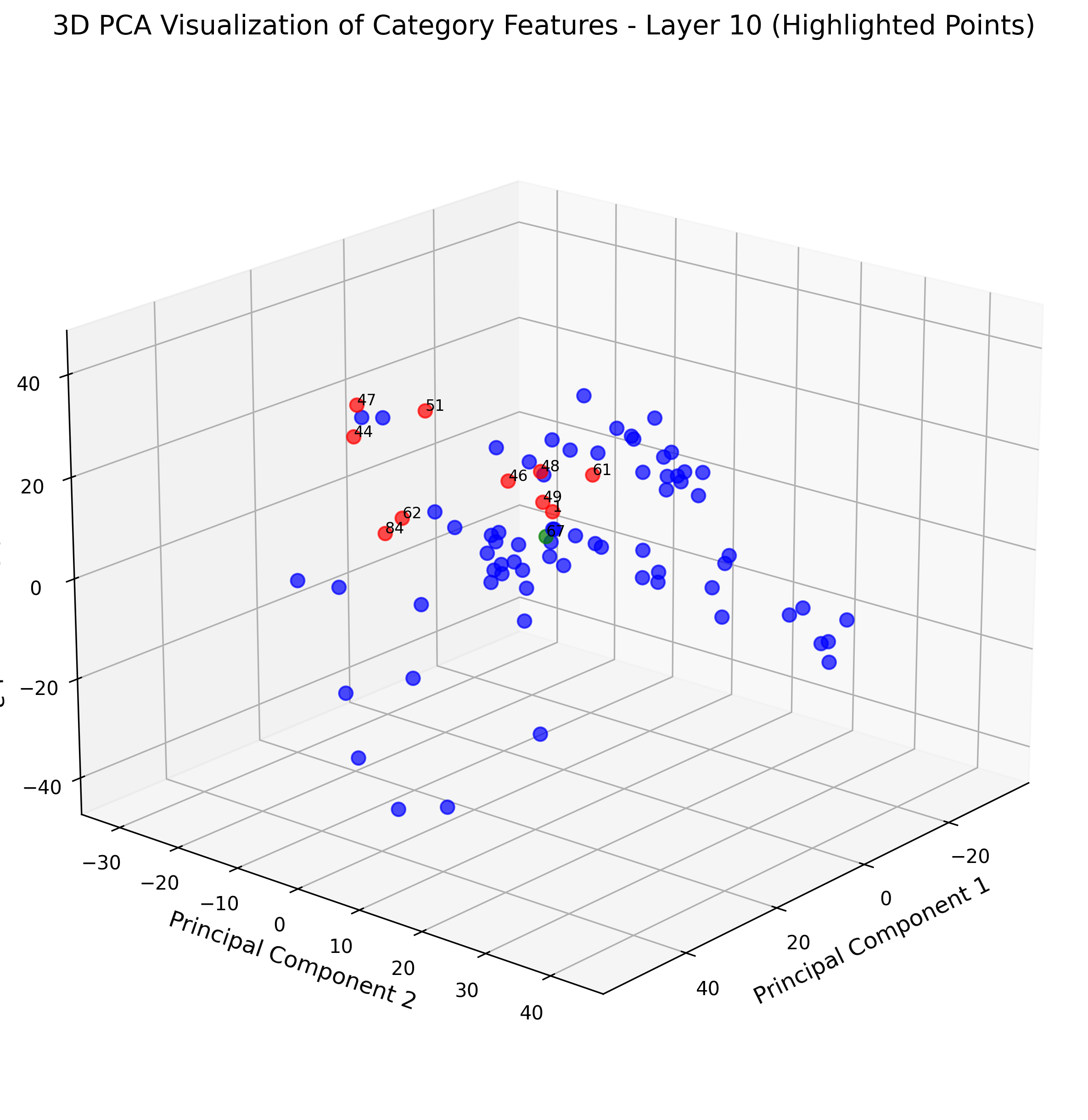}
  \hfill
  \includegraphics[width=0.32\linewidth]{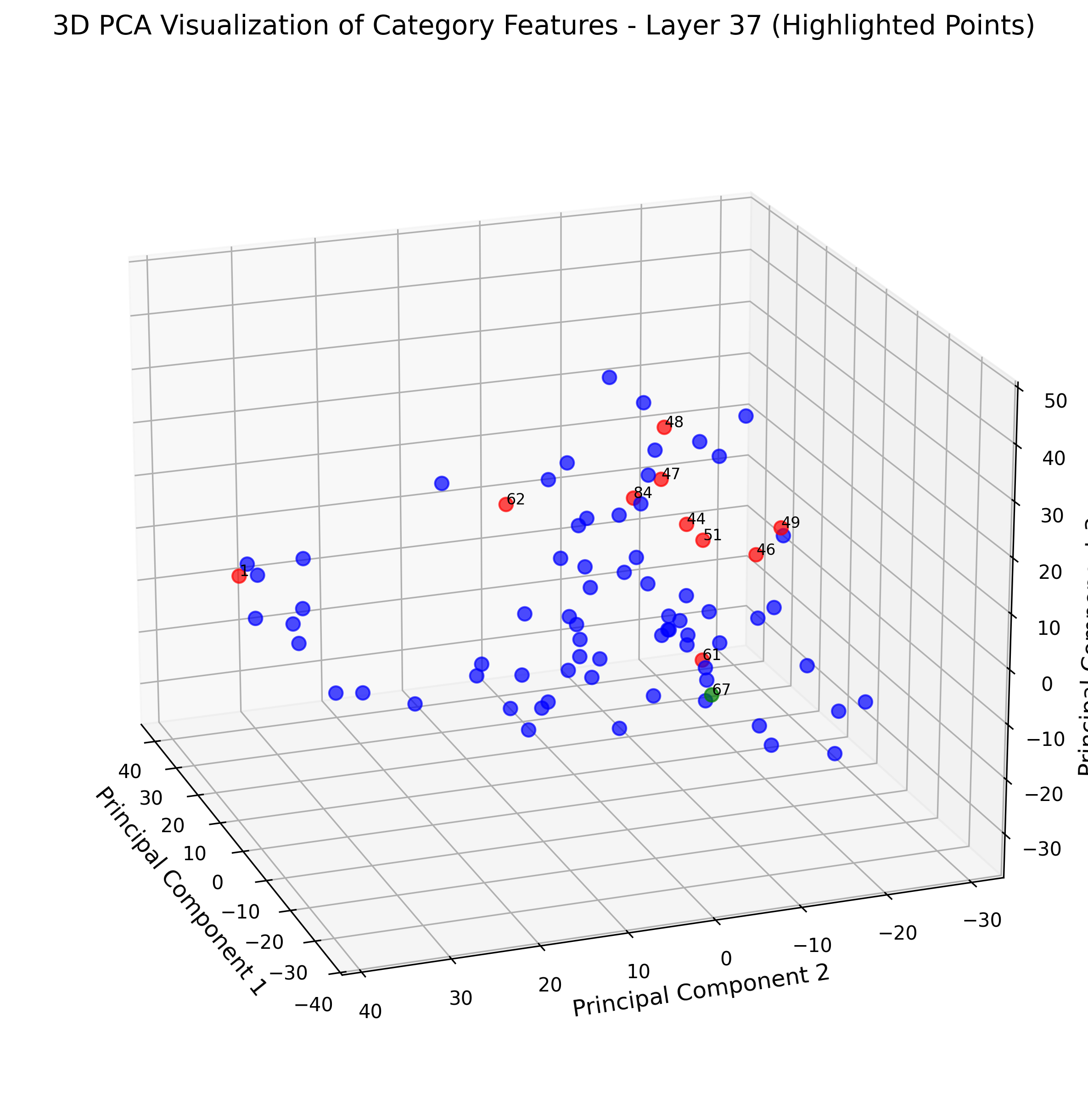}
  \hfill
  \includegraphics[width=0.32\linewidth]{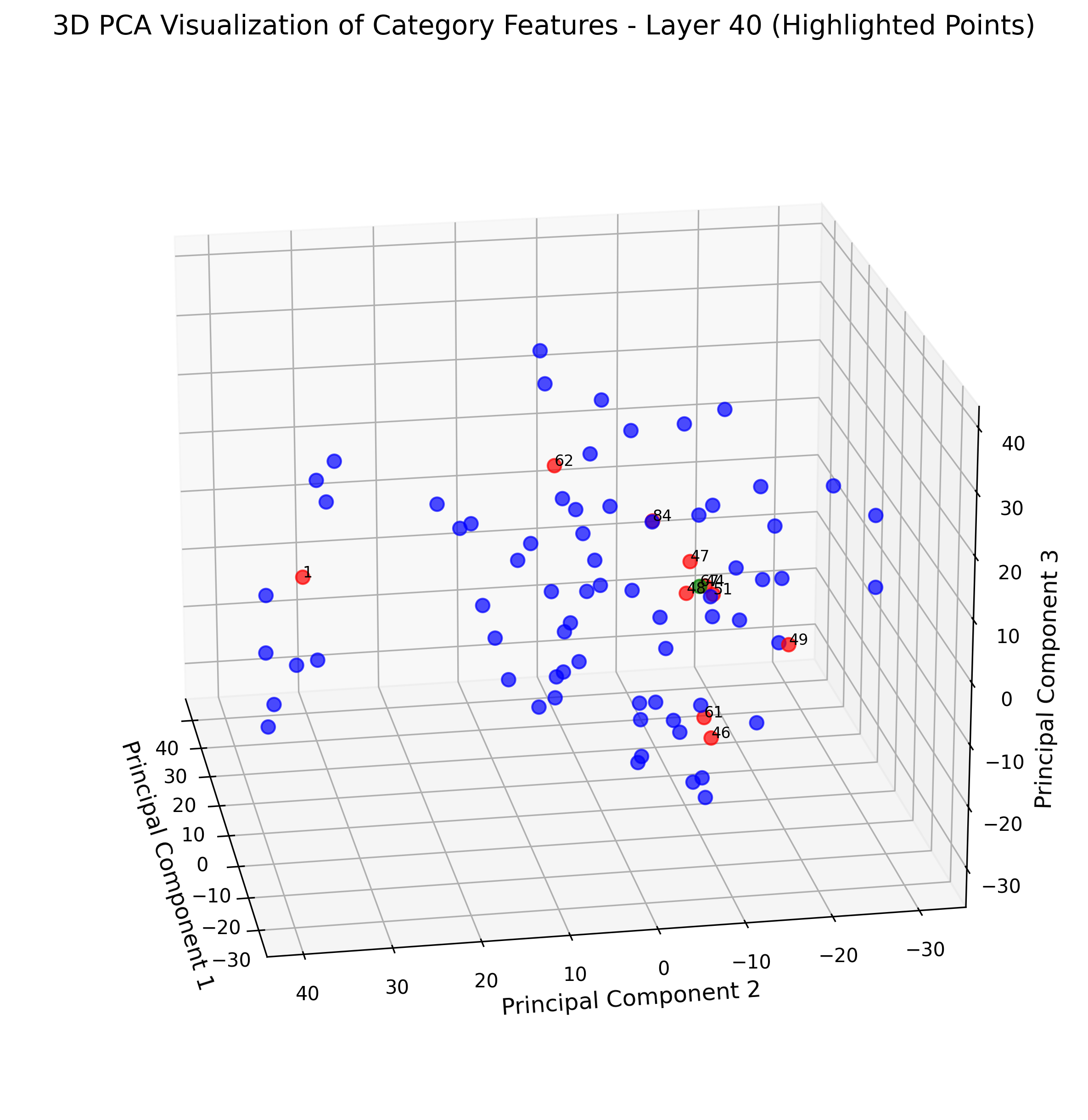}
  \caption{(Table-Centric:Textual) Layer 10 (left), layer 37 (middle), and layer 40 (right) showing persistent co-occurrence patterns in final prediction stages.}
  \label{fig:text_layer10_40}
\end{figure}

We visualize the original textual representations of "dining table" (green) and its top-10 co-occurring objects (red). Similar to the visual modality analysis, these textual representations exhibit co-occurrence-driven entanglement patterns: lower transformer layers (layer 0-10) show tight clustering between dining tables and frequently co-occurring objects (\cref{fig:text_layer0_5}), while the final prediction layers (layer 37-40) maintain entanglement despite semantic abstraction (\cref{fig:text_layer10_40}). This confirms that instruction bias affects both modalities in parallel, with frequent co-occurrence patterns shaping representational geometry across all stages of MLLM processing.

\subsection*{C.3. Car-Centric Original Visual Representation Visualizations}

We extend table-centric analysis methodology to "car" object visualizations, showing PCA projections of car (green) and its top-10 co-occurring objects (red) across network depths:

\begin{figure}[t]
  \centering
  \includegraphics[width=0.32\linewidth]{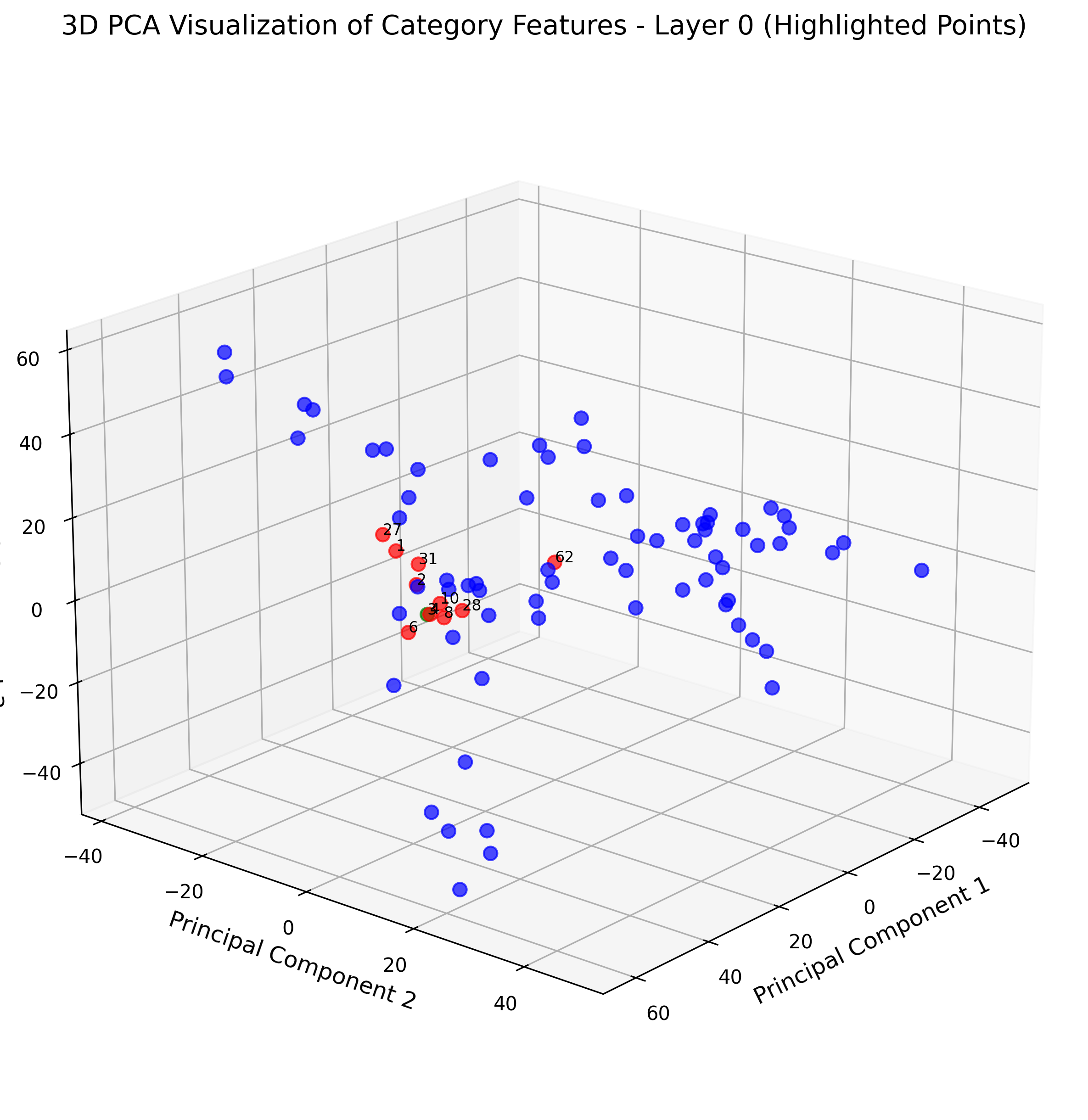}
  \hfill
  \includegraphics[width=0.32\linewidth]{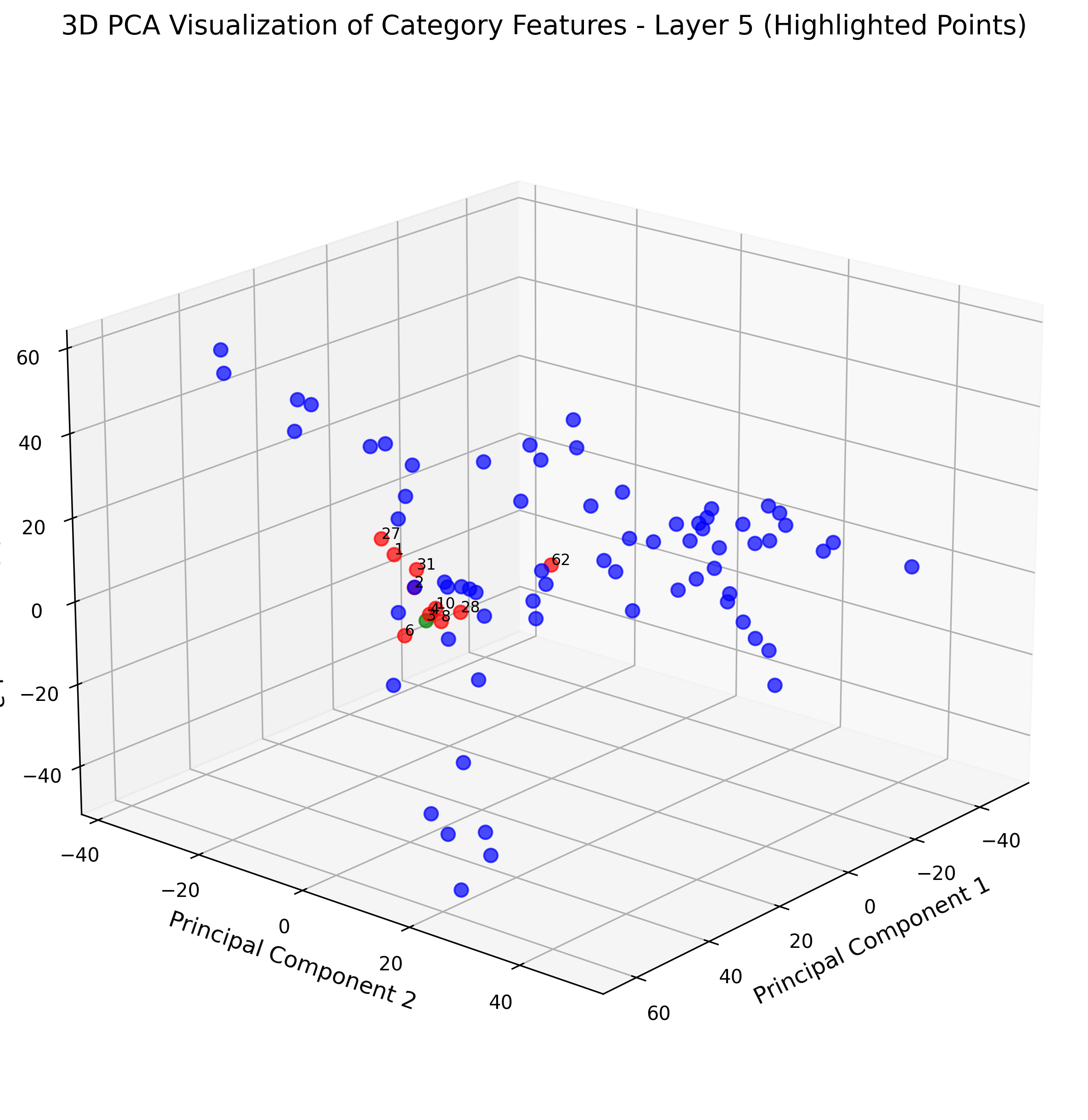}
  \hfill
  \includegraphics[width=0.32\linewidth]{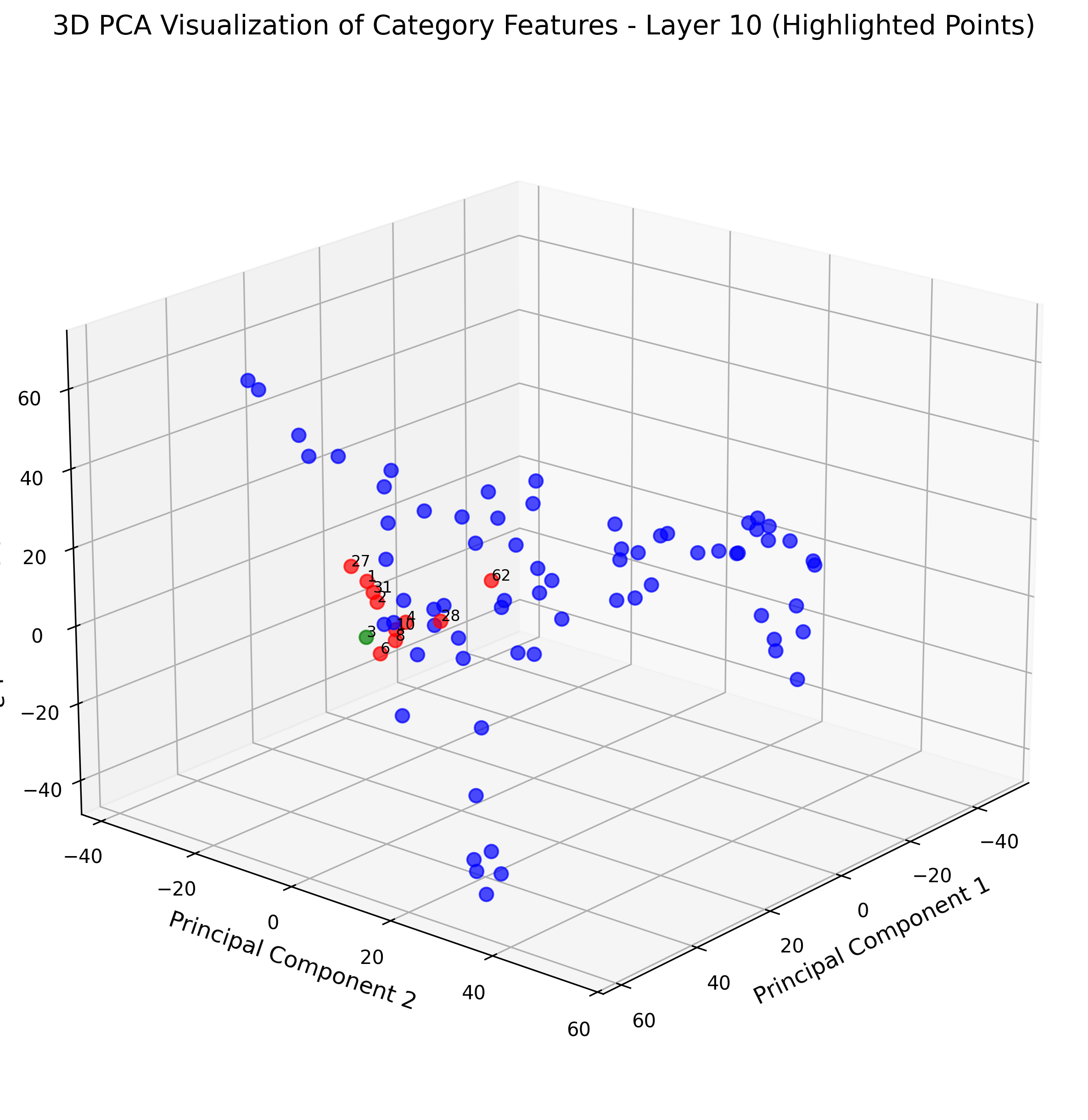}
  \caption{(Car-Centric:Visual) Early layers (0-10) showing initial entanglement patterns.}
  \label{fig:car_layer0_10}
\end{figure}

\begin{figure}[t]
  \centering
  \includegraphics[width=0.32\linewidth]{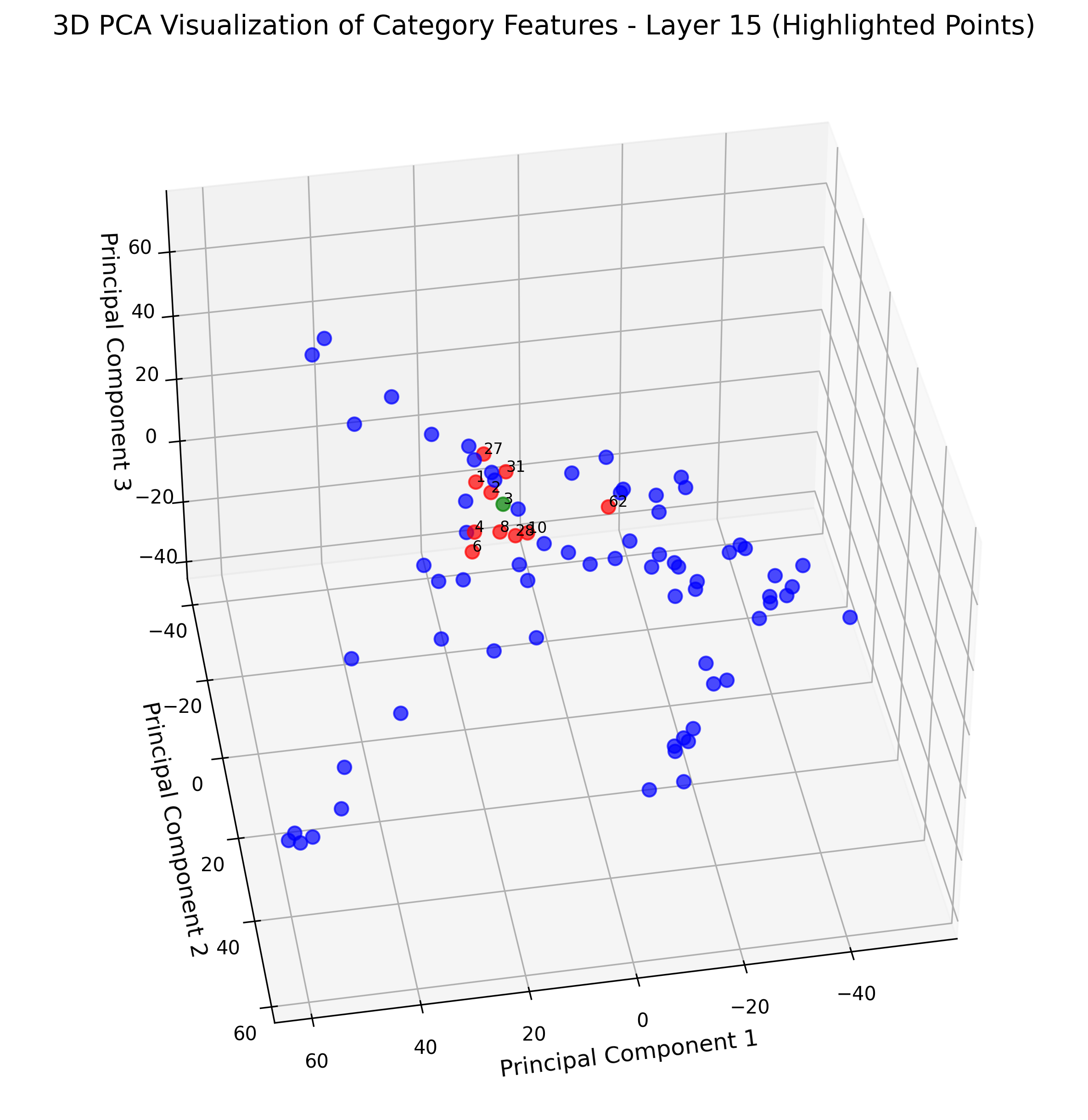}
  \hfill
  \includegraphics[width=0.32\linewidth]{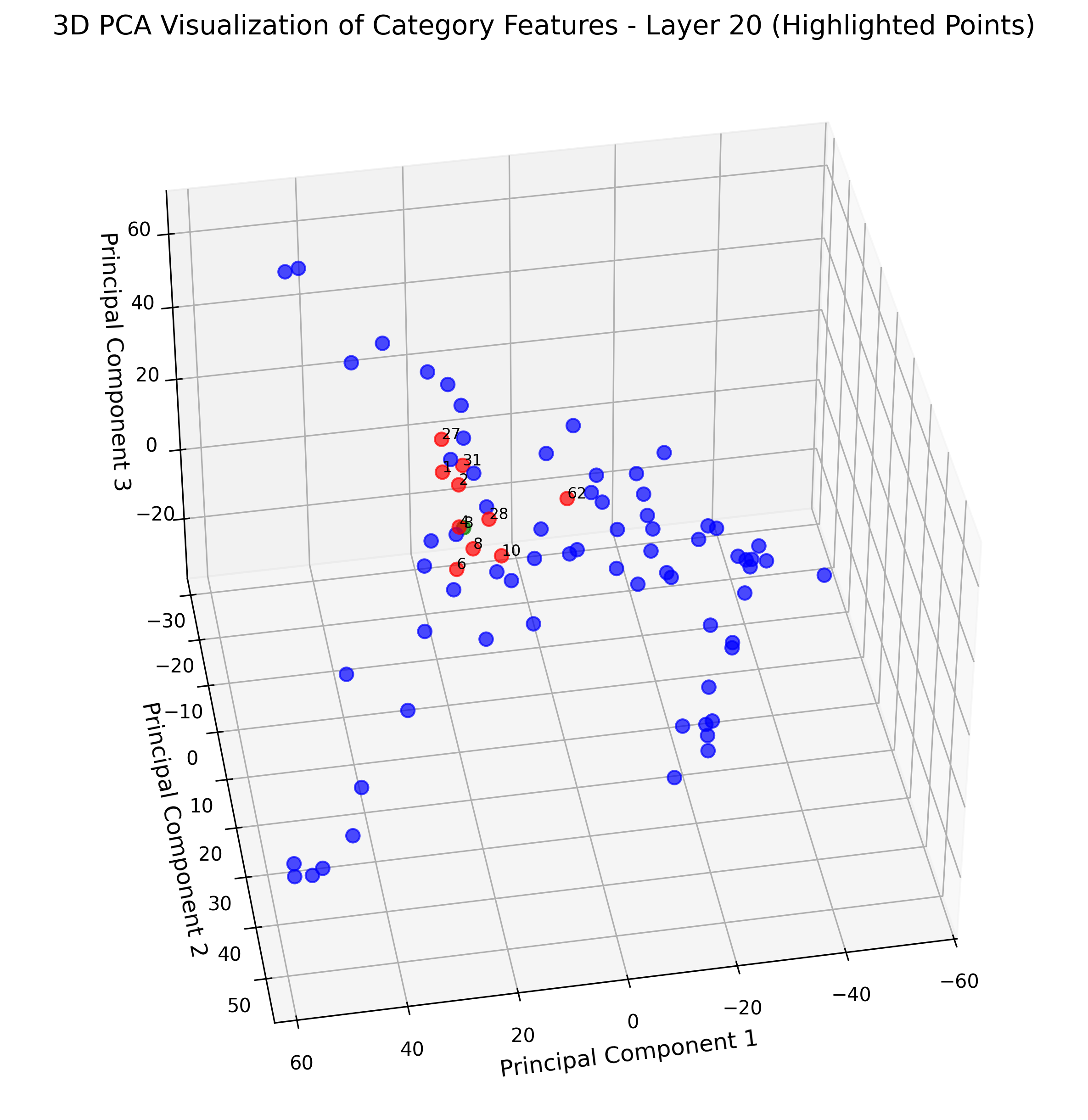}
  \hfill
  \includegraphics[width=0.32\linewidth]{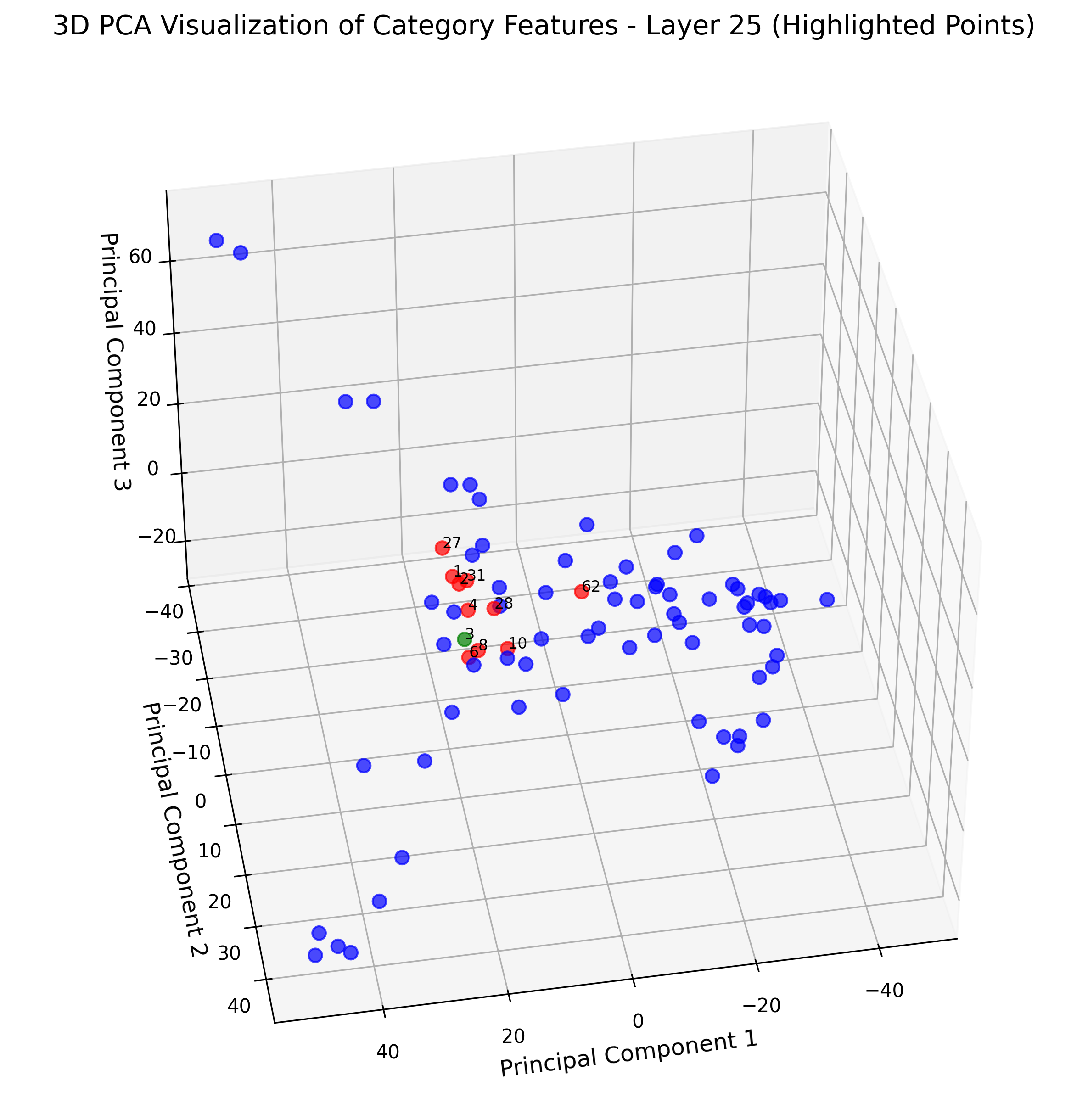}
  \caption{(Car-Centric:Visual) Mid-depth layers (15-25) maintaining co-occurrence structures.}
  \label{fig:car_layer15_25}
\end{figure}

\begin{figure}[t]
  \centering
  \includegraphics[width=0.32\linewidth]{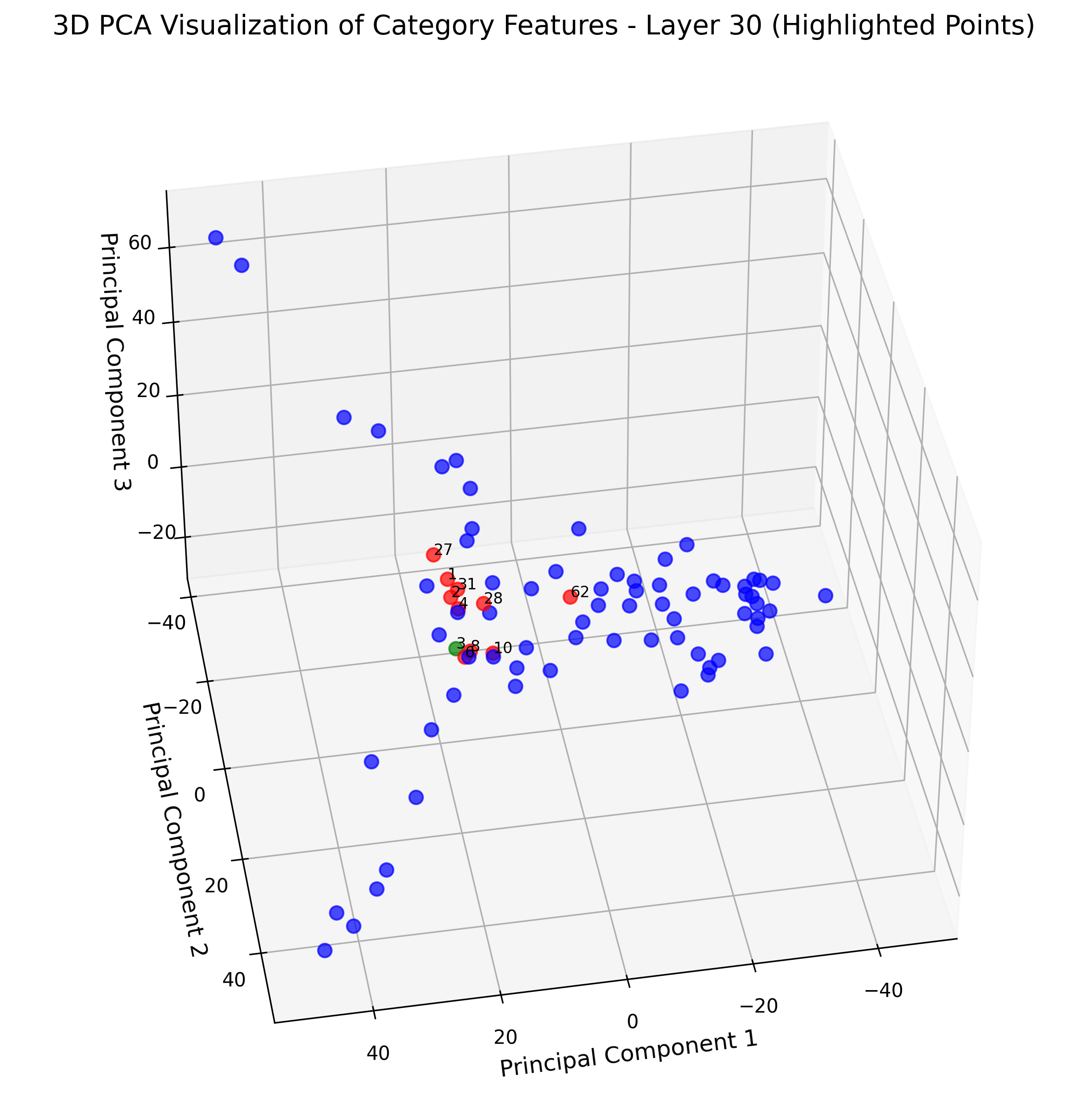}
  \hfill
  \includegraphics[width=0.32\linewidth]{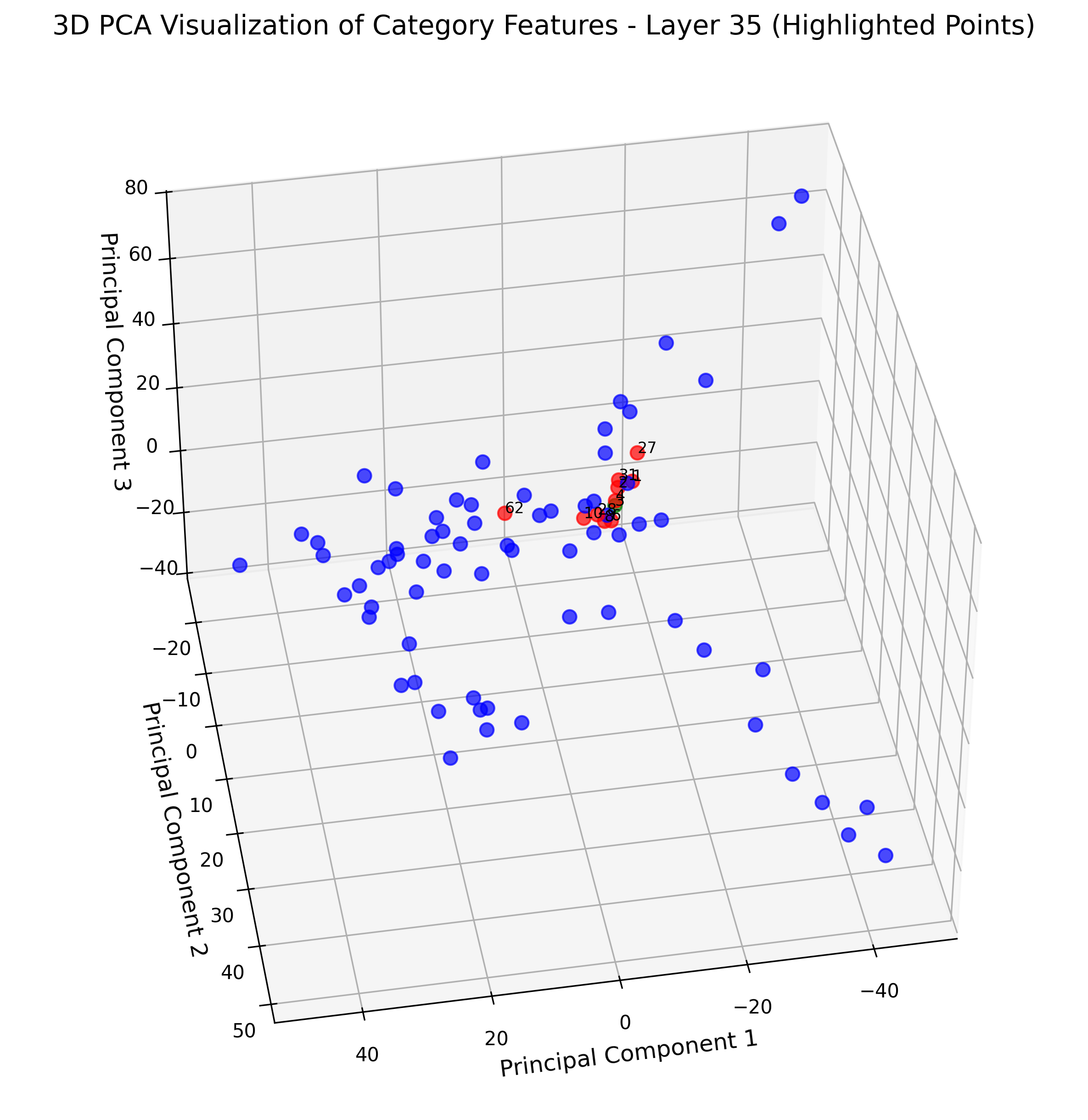}
  \hfill
  \includegraphics[width=0.32\linewidth]{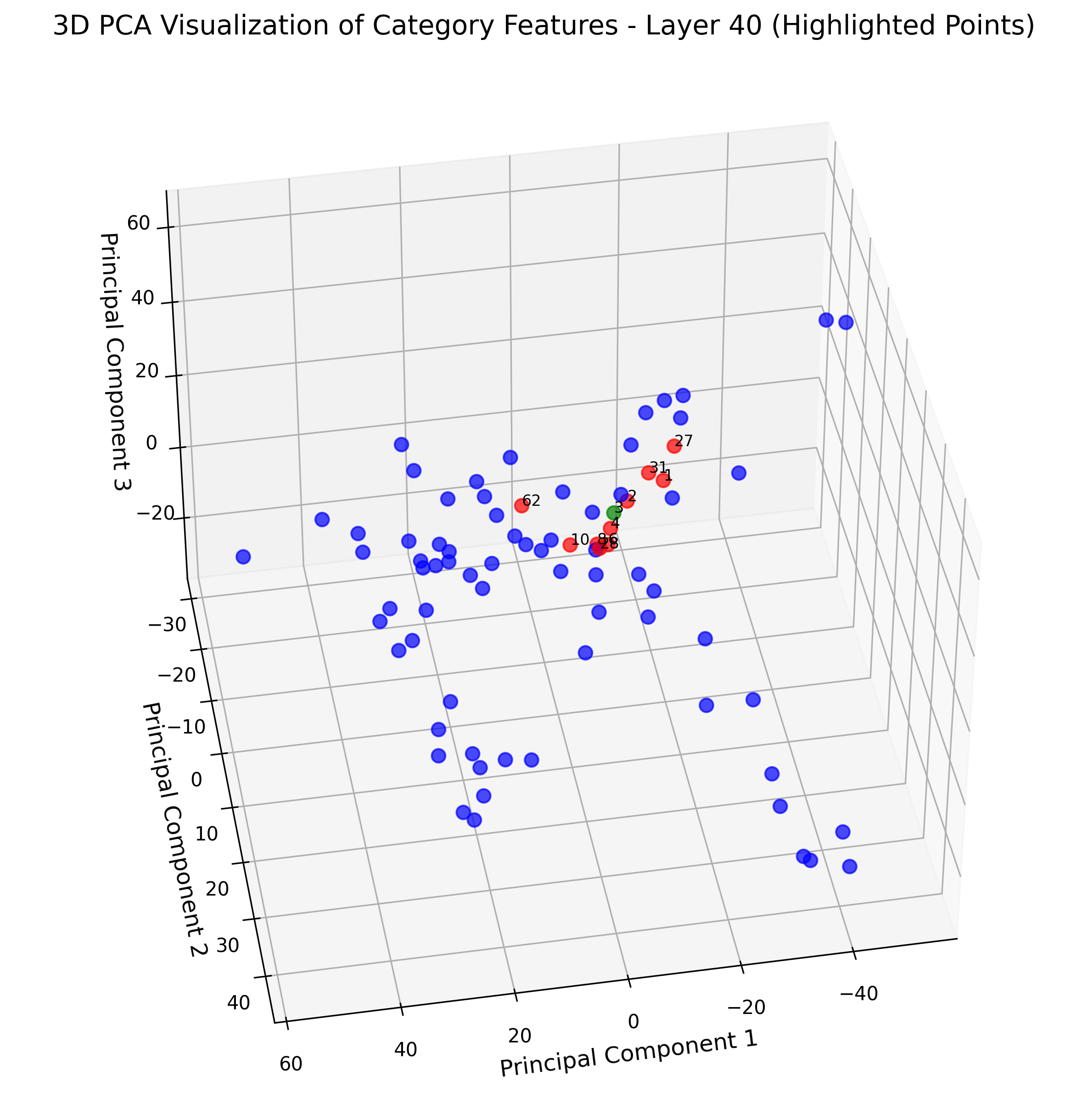}
  \caption{(Car-Centric:Visual) Final layers (30-40) retaining co-occurrence patterns.}
  \label{fig:car_layer30_40}
\end{figure}

Key findings mirror table-centric patterns:
\begin{itemize}
  \item Early layers (0-10) show training data bias clustering (\cref{fig:car_layer0_10})
  \item Mid-depth layers (15-25) maintain entanglement despite abstraction (\cref{fig:car_layer15_25})
  \item Final layers (30-40) retain significant co-occurrence patterns (\cref{fig:car_layer30_40})
\end{itemize}

This confirms instruction bias-induced representational entanglement generalizes across object categories, with co-occurrence clustering persisting through network depth in both car-centric and table-centric analyses.

\subsection*{C.4. Table-Centric Disentangled Visual Representation Visualizations}

We visualize disentangled dining table (green) and top-10 co-occurring objects (red) using PCA across LLM layers, demonstrating our framework's ability to separate entangled representations compared to original visual patterns:

\begin{figure}[t]
  \centering
  \includegraphics[width=0.32\linewidth]{./Paper_Draft_assets/disentangled/3d_PCA_layer0_elev20_azim70.png}
  \hfill
  \includegraphics[width=0.32\linewidth]{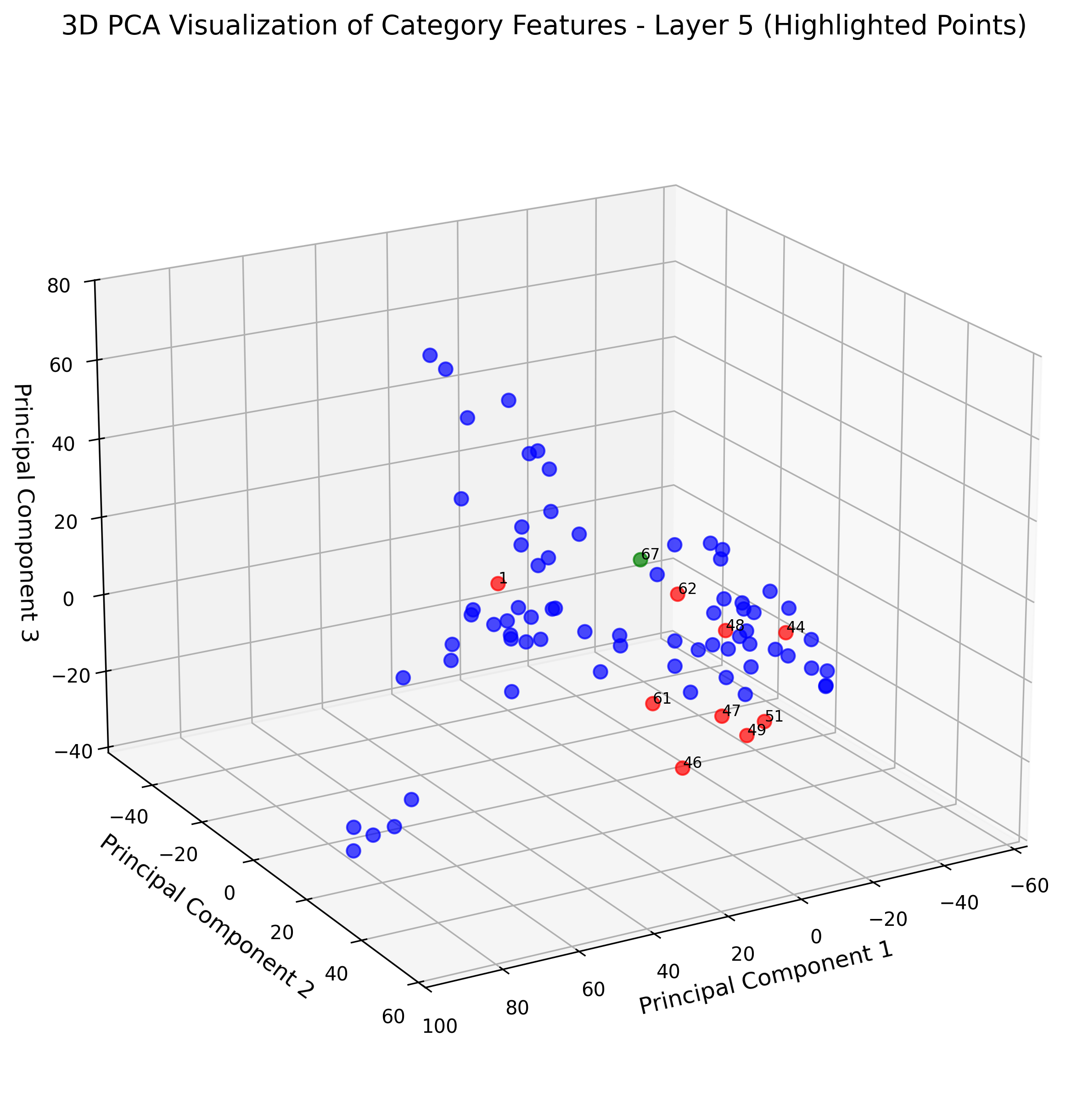}
  \hfill
  \includegraphics[width=0.32\linewidth]{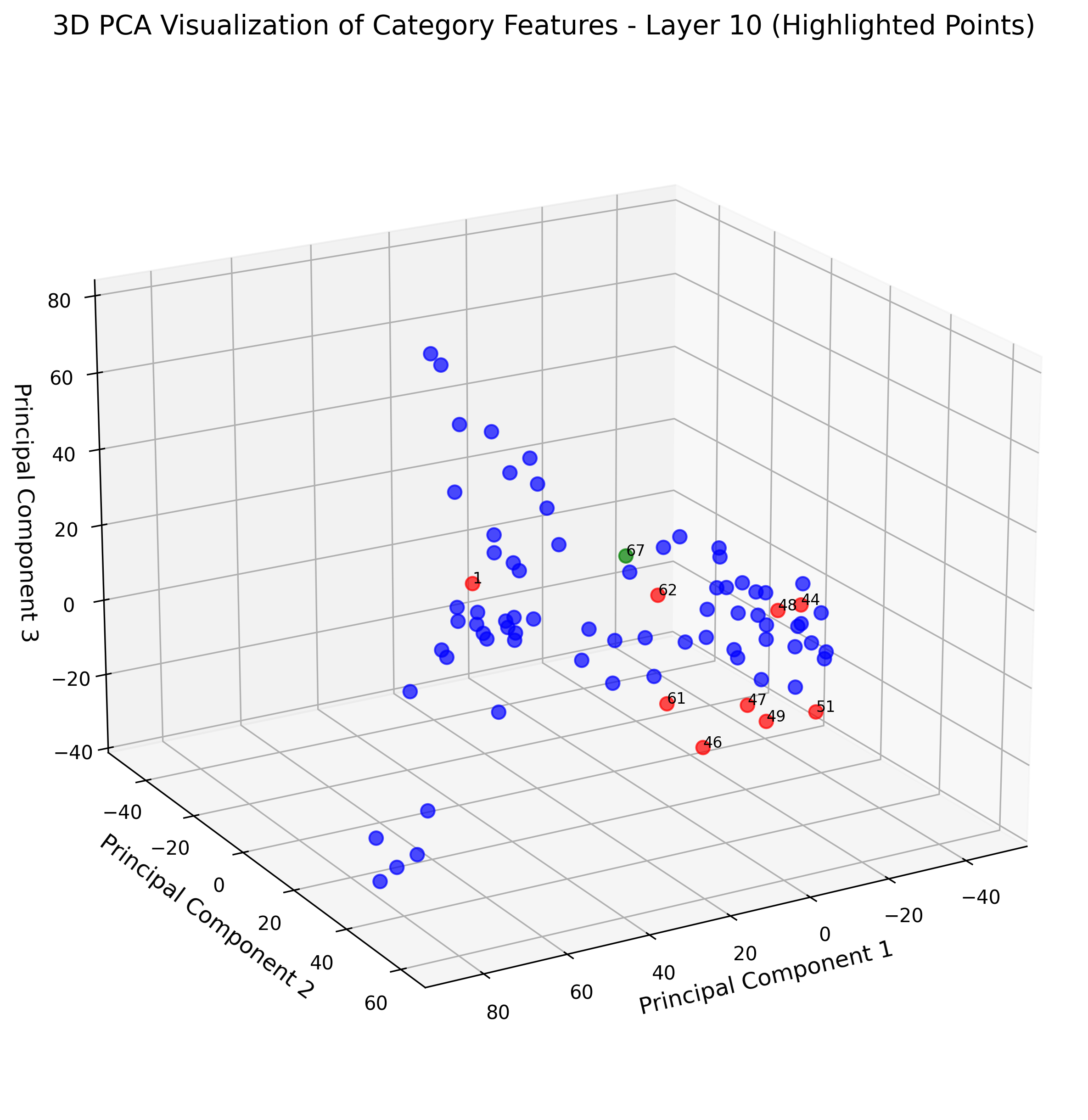}
  \caption{(Disentangled) Layer 0 (left) to layer 10 (right) showing progressive separation.}
  \label{fig:disentangle_layer0_10}
\end{figure}

\begin{figure}[t]
  \centering
  \includegraphics[width=0.32\linewidth]{./Paper_Draft_assets/disentangled/3d_PCA_layer15_elev20_azim90.png}
  \hfill
  \includegraphics[width=0.32\linewidth]{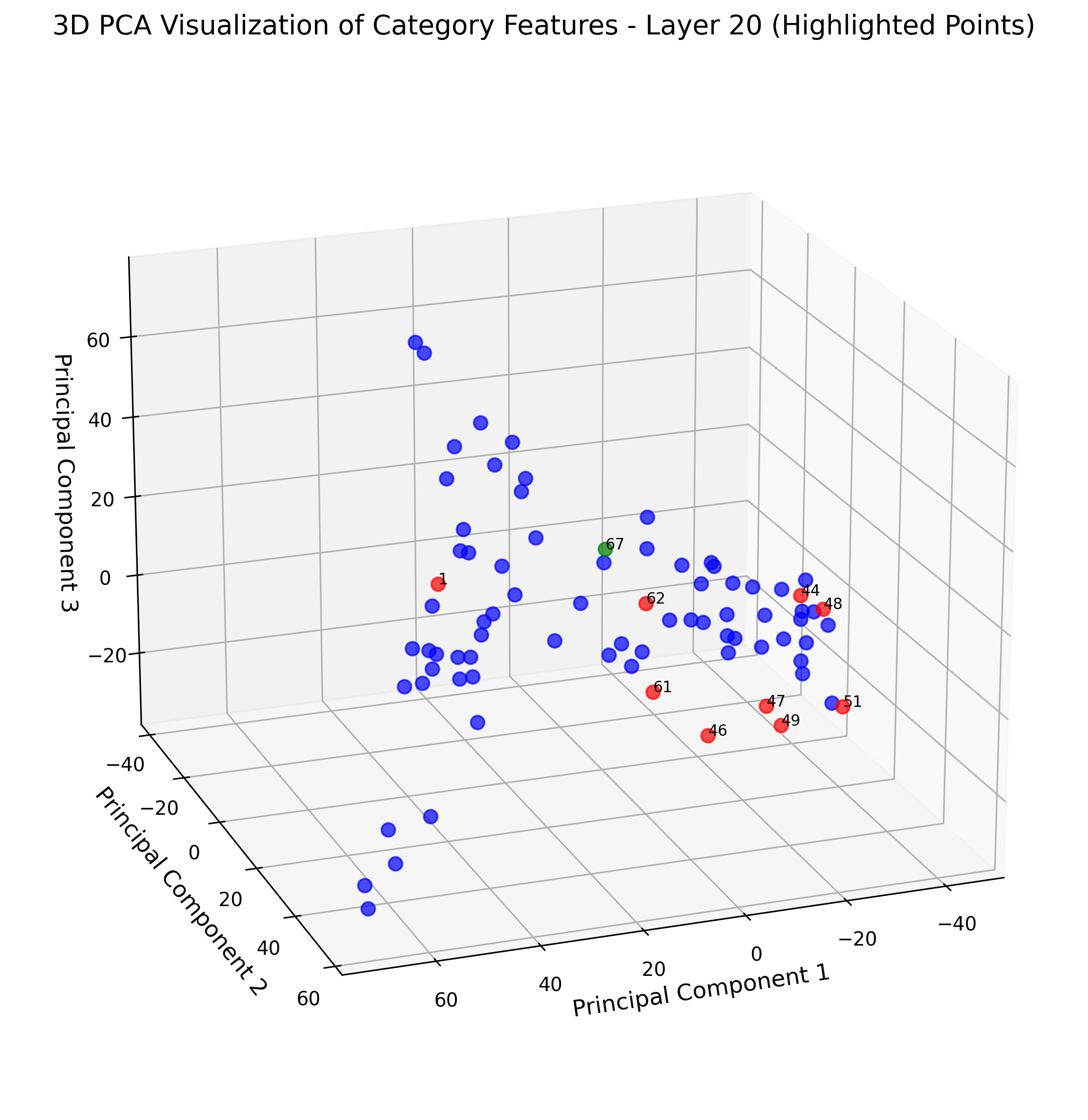}
  \hfill
  \includegraphics[width=0.32\linewidth]{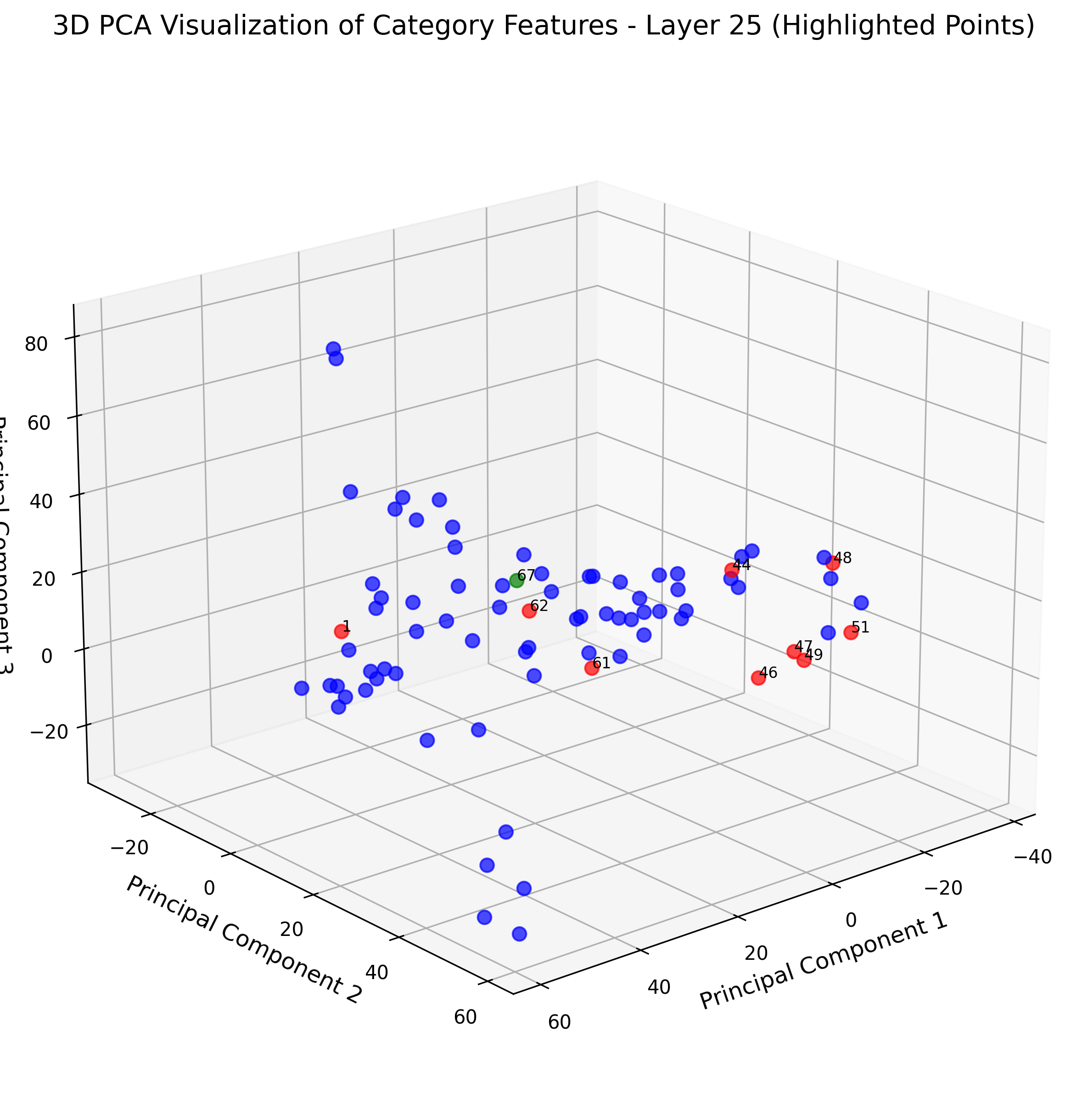}
  \caption{(Disentangled) Mid-depth layers (15-25) showing increasing spatial differentiation.}
  \label{fig:disentangle_layer15_25}
\end{figure}

\begin{figure}[t]
  \centering
  \includegraphics[width=0.32\linewidth]{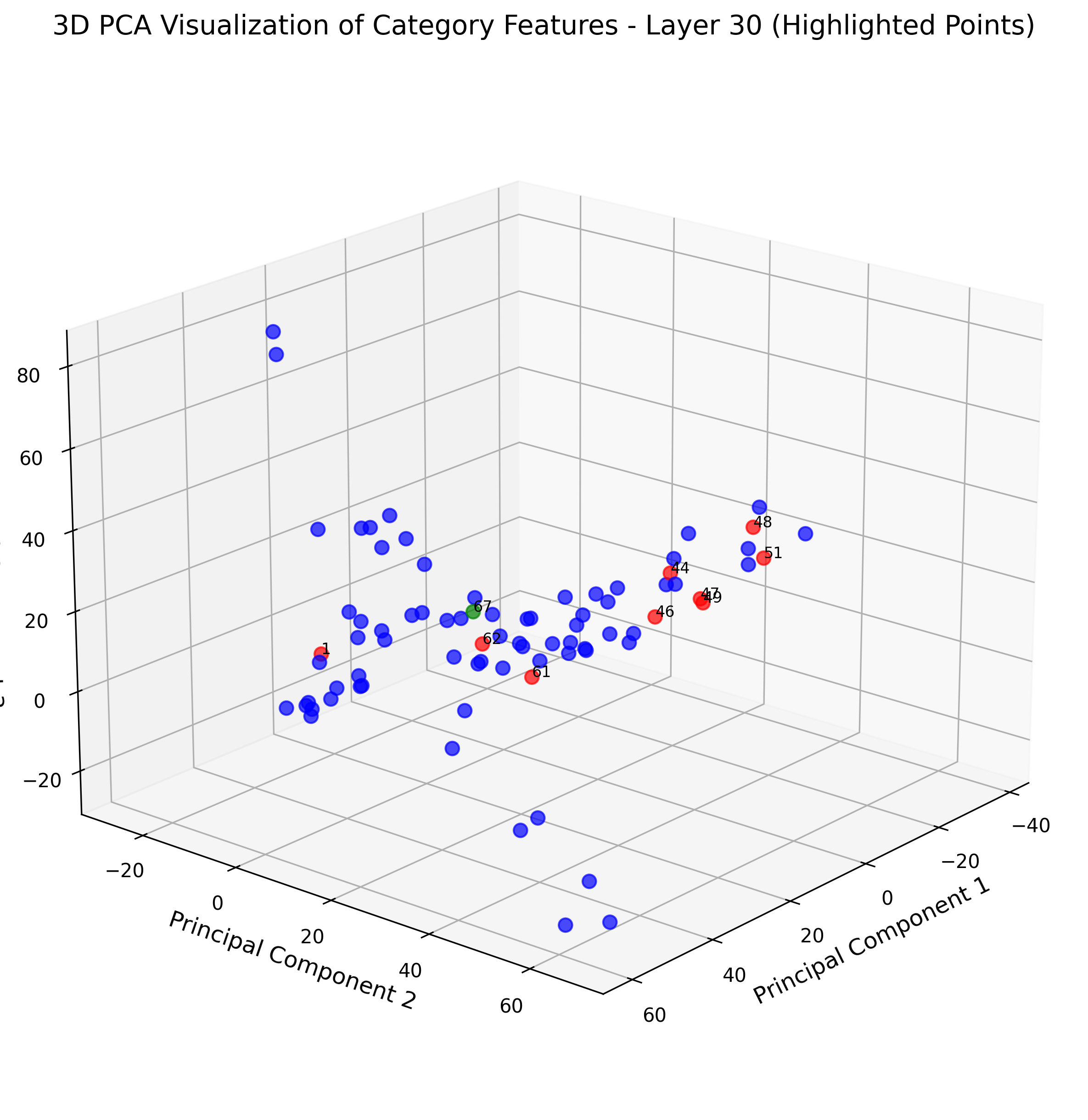}
  \hfill
  \includegraphics[width=0.32\linewidth]{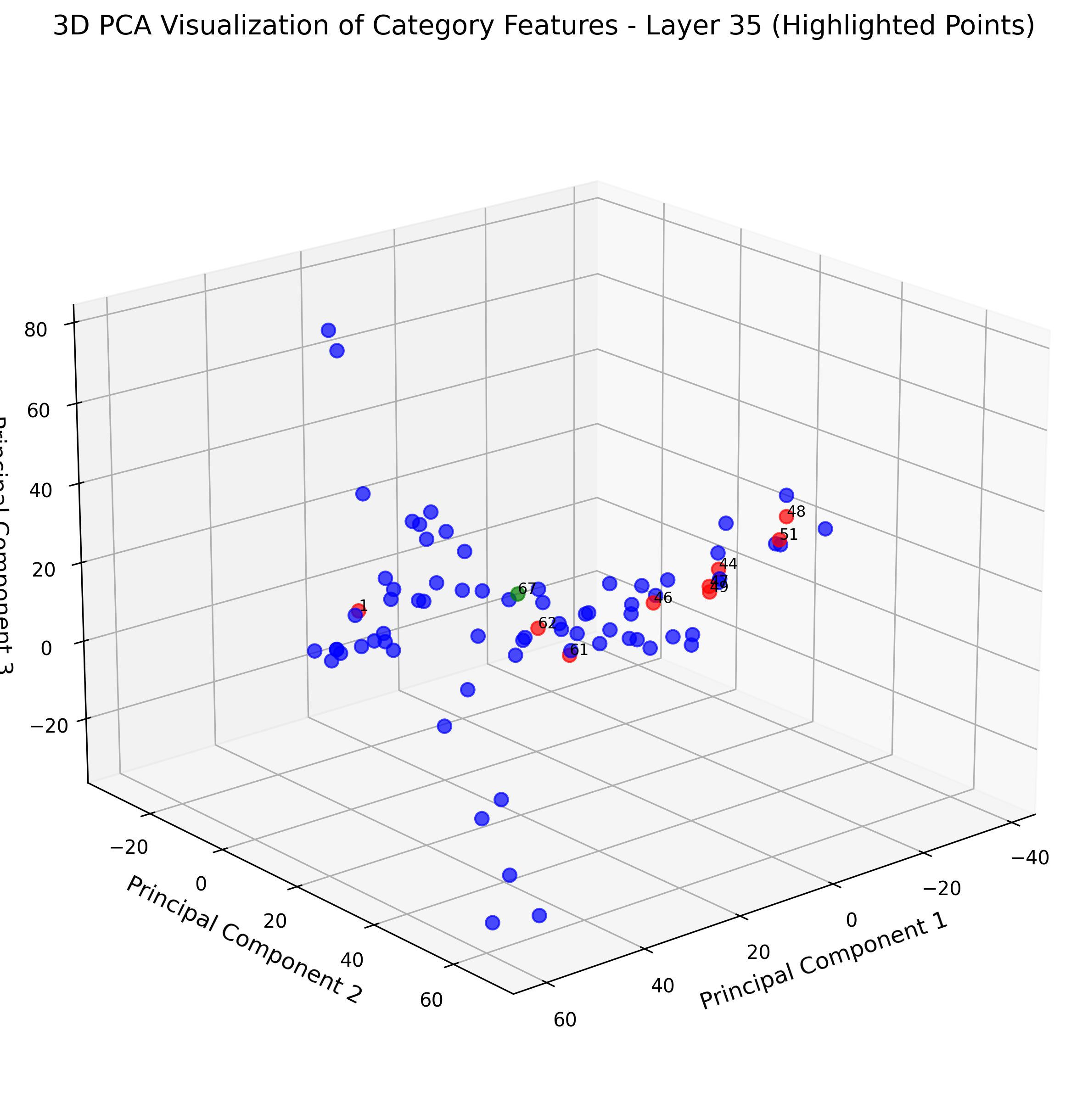}
  \hfill
  \includegraphics[width=0.32\linewidth]{./Paper_Draft_assets/disentangled/3d_PCA_layer40_elev20_azim70.png}
  \caption{(Disentangled) Final layers (30-40) achieving complete spatial separation.}
  \label{fig:disentangle_layer30_40}
\end{figure}

Key improvements over original representations:
\begin{itemize}
  \item \textbf{Early separation}: Disentanglement begins at layer 0 (\cref{fig:disentangle_layer0_10}), breaking projector-induced biases observed in original representations (\cref{fig:layer0_1,fig:layer3_5}).
  
  \item \textbf{Progressive differentiation}: By layer 15, dining tables separate from co-occurring objects (\cref{fig:disentangle_layer15_25}), outperforming persistent entanglement in original mid-layers (\cref{fig:layer10_15,fig:layer20_30}).
  
  \item \textbf{Final disentanglement}: Layer 40 achieves distinct semantic subspaces (\cref{fig:disentangle_layer30_40}), eliminating co-occurrence biases that original representations retain (\cref{fig:layer35_40}).
\end{itemize}

These results validate our framework's success in disrupting instruction bias propagation while preserving discriminative learning. Complete separation in final layers (\cref{fig:disentangle_layer30_40}) addresses MLLMs' tendency to confuse frequent co-occurring objects through representation disentanglement.

\end{document}